\documentclass{article}

\usepackage[letterpaper,margin=1in]{geometry}
\usepackage[authoryear,round]{natbib}
\usepackage{microtype}

\usepackage{bm} 
\usepackage{amsmath,amssymb,amsfonts,amsthm,bbm}
\usepackage{verbatim}
\usepackage{setspace}
\usepackage{graphicx}
\usepackage{subcaption}
\usepackage{enumitem}

\usepackage[dvipsnames]{xcolor}

\usepackage{caption}    % old file had this
\usepackage{wrapfig}    \usepackage{float}
\usepackage{array} 

\usepackage{microtype}
\usepackage{graphicx}
\usepackage{subcaption}
\usepackage{booktabs} % for professional tables
\usepackage{multirow}
\usepackage[normalem]{ulem}
\useunder{\uline}{\ul}{}

\usepackage{hyperref}

\usepackage{amsmath}
\usepackage{amssymb}
\usepackage{mathtools}
\usepackage{amsthm}

\usepackage[capitalize,noabbrev]{cleveref}

\theoremstyle{plain}
\newtheorem{theorem}{Theorem}[section]

\theoremstyle{definition}
\newtheorem{definition}[theorem]{Definition}

\theoremstyle{remark}

\newtheorem*{imp}{\textbf{Practical Implication}}

\def\eqref#1{equation~\ref{#1}}
\def\1{\bm{1}}

\DeclareMathAlphabet{\mathsfit}{\encodingdefault}{\sfdefault}{m}{sl}
\SetMathAlphabet{\mathsfit}{bold}{\encodingdefault}{\sfdefault}{bx}{n}

\def\gE{{\mathcal{E}}}
\def\gF{{\mathcal{F}}}

\def\gP{{\mathcal{P}}}

\def\gV{{\mathcal{V}}}

\newcommand{\KL}{D_{\mathrm{KL}}}

\DeclareMathOperator*{\argmax}{arg\,max}

\newcommand{\veps}{\varepsilon}
\newcommand{\op}{\mathrm{op}}
\renewcommand{\P}{\mathbb{P}}

\providecommand{\gV}{\mathcal{V}}
\providecommand{\gE}{\mathcal{E}}
\providecommand{\gF}{\mathcal{F}}
\providecommand{\gP}{\mathcal{P}}
\providecommand{\KL}{\mathrm{KL}}

\begin{document}

\title{How Does Unfaithful Reasoning Emerge from Autoregressive Training? A Study of Synthetic Experiments}
\author{Fuxin Wang\footnotemark[1], Amr Alazali, Yiqiao Zhong\footnotemark[1]\\~\\ University of Wisconsin-Madison, Madison, WI, USA} 
\footnotetext[1]{Correspondence to: fwang297@wisc.edu, yiqiao.zhong@wisc.edu}

\date{}

\maketitle
\begin{abstract}
\setcounter{footnote}{1}
Chain-of-thought (CoT) reasoning generated by large language models (LLMs) is often unfaithful: intermediate steps can be logically inconsistent or fail to reflect the causal relationship leading to the final answer. Despite extensive empirical observations, a fundamental understanding of CoT is lacking—what constitutes faithful CoT reasoning, and how unfaithfulness emerges from autoregressive training. We study these questions using well-controlled synthetic experiments, training small transformers on noisy data to solve modular arithmetic expressions step by step, a task we term \textit{Arithmetic Expression Reasoning}. We find that models can learn faithful reasoning that causally follows the underlying arithmetic rules, but only when the training noise is below a critical threshold, a phenomenon attributable to simplicity bias. At higher noise levels, training dynamics exhibit a transition from faithful stepwise reasoning to unfaithful skip-step reasoning via an intermediate mixed mode characterized by a transient increase in prediction entropy. Mechanistic analysis reveals that  models learn to encode internal uncertainty by resolving inconsistent reasoning steps, which suggests the emergence of implicit self-verification from autoregressive training.
%this non-monotonic entropy behavior reflects emergent internal uncertainty and implicit self-verification during autoregressive training.
Code is available\footnote{ \url{https://github.com/jwtr297/Arithmetic_Expression_Reasoning}}.
\end{abstract}

\section{Introduction}

Large language models (LLMs) have recently achieved remarkable performance on reasoning tasks. Much of this success is attributed to chain-of-thought (CoT) reasoning \citep{wei2022chain}, in which models generate intermediate steps before producing a final answer. Such behavior can be encouraged by training on data with explicit reasoning traces \citep{lightman2023let}, which promotes learning simpler, compositional reasoning steps. %Training on data with explicit reasoning traces \citep{lightman2023let} encourages models to learn simpler compositional steps. 

However, the mechanism of CoT reasoning remains largely opaque, and it is unclear whether and when LLMs truly acquire the rules governing the reasoning traces. Despite strong benchmark performance, LLMs are known to have stability issues: evaluation can be fragile under minor benchmark perturbation  or data contamination \citep{alzahrani2024benchmarks, gulati2024putnam, hosseini2not}. Moreover, CoT reasoning can be superfluous: LLMs sometimes generate reasoning steps as \textit{post-hoc explanations} for answers they already know, or even fabricate plausible-looking reasoning when  prompts steer them toward incorrect solutions \citep{turpin2023language, lanham2023measuring, barez2025chain}. The ambiguity has drawn criticisms, as \citet{shojaee2025illusion} argues that reasoning LLMs create ``illusion of thinking''. More broadly, the lack of a principled understanding of CoT reasoning has far-reaching implications for AI safety and alignment \citep{betley2025emergent}.

\paragraph{Can faithful CoT reasoning be clearly defined?} Unfaithful reasoning in LLMs has been widely documented. For instance, \citet{wei2022chain, bao2024llms, bao2025likely} show that models may produce incorrect reasoning alongside correct solutions, or correct reasoning with incorrect solutions. \citet{turpin2023language} find that generated reasoning can depend on spurious features such as prompt cues or the ordering of multiple-choice options. %While many have recognized the potential risks of unfaithful reasoning, there is a lack of consensus about what faithful reasoning entails.
However, there is still no consensus on what constitutes faithful reasoning.

Existing approaches for assessing CoT faithfulness fall into two categories. (i) Consistency checking, which examines whether the reasoning steps are logically consistent with the final answer \citep{lyu2023faithful}. (ii) Intervention on reasoning steps, which tests whether altering the reasoning steps, such as rephrasing or injecting errors, changes the final solution 
%we check whether modifying the reasoning steps, such as rephrasing and adding a mistake in reasoning, yields different final solutions
\citep{turpin2023language, lanham2023measuring}. These approaches target two key pitfalls of CoT reasoning: mimicking reasoning traces does not guarantee logical consistency, and the sequential structure of a reasoning trace does not imply a causal relationship. Motivated by these empirical studies, we investigate faithful reasoning from both angles.   

\paragraph{Model's internal uncertainty towards unfaithful reasoning.} One surprising aspect of LLMs is their meta-reasoning ability to reflect on their own reasoning traces, detect unfaithful reasoning, and thereby improve accuracy. Self-reflection and self-verification \citep{shinn2023reflexion, madaan2023self} externalize a model's internal uncertainty by inducing generated statements like ``I'm not sure''. A fundamental question is: how do models develop such reflective behavior through autoregressive training? Empirical evidence suggests that base models such as GPT-3 already exhibit primitive self-reflection abilities \citep{weng2023large}, but the mechanisms of emergence are poorly understood.

\paragraph{Modeling CoT reasoning via a synthetic task.}
Our synthetic task abstracts the mathematical reasoning benchmark \texttt{GSM8K} \citep{cobbe2021training}, for which
LLMs generate a sequence of reasoning tokens (the text preceding \#\#\#\#) by chaining multiple arithmetic expressions step by step, culminating in a final solution (the integer following \#\#\#\#).

\vspace{1mm}
{\small
\textbf{Prompt.} ``A candle melts by 2 centimeters every hour that it burns. How many centimeters shorter will a candle be after burning from 1:00 PM to 5:00 PM?''

\textbf{Answer.} ``The candle burns for \texttt{5 - 1 = <<5-1=4>>4} hours. Thus, the candle will be \texttt{2 * 4 = <<2*4=8>>8} centimeters shorter. \#\#\#\# 8.''}
\vspace{1mm}
%\AAA{I think we should remove the example GSM8K prompt and answer. I dont think it adds much value and may confuse the reader since you are not training on gsm8k.}

Our synthetic task, termed \textit{Arithmetic Expression Reasoning (AER)} task, preserves the arithmetic structure of each question while stripping away natural language. Formally, the training data consist of sequences formatted as arithmetic expression chains, where each chain follows the format:
\begin{equation}\label{eq:format}
\underbrace{a \times b - c}_{e_1 :\text{prompt}} \ \to \ \underbrace{d - c}_{e_2: \text{reasoning}} \ \to \ \underbrace{o}_{e_3: \text{solution}}.
\end{equation}

\begin{figure}[t]
\hspace{-4mm}
\centering
\includegraphics[width=0.55\columnwidth]{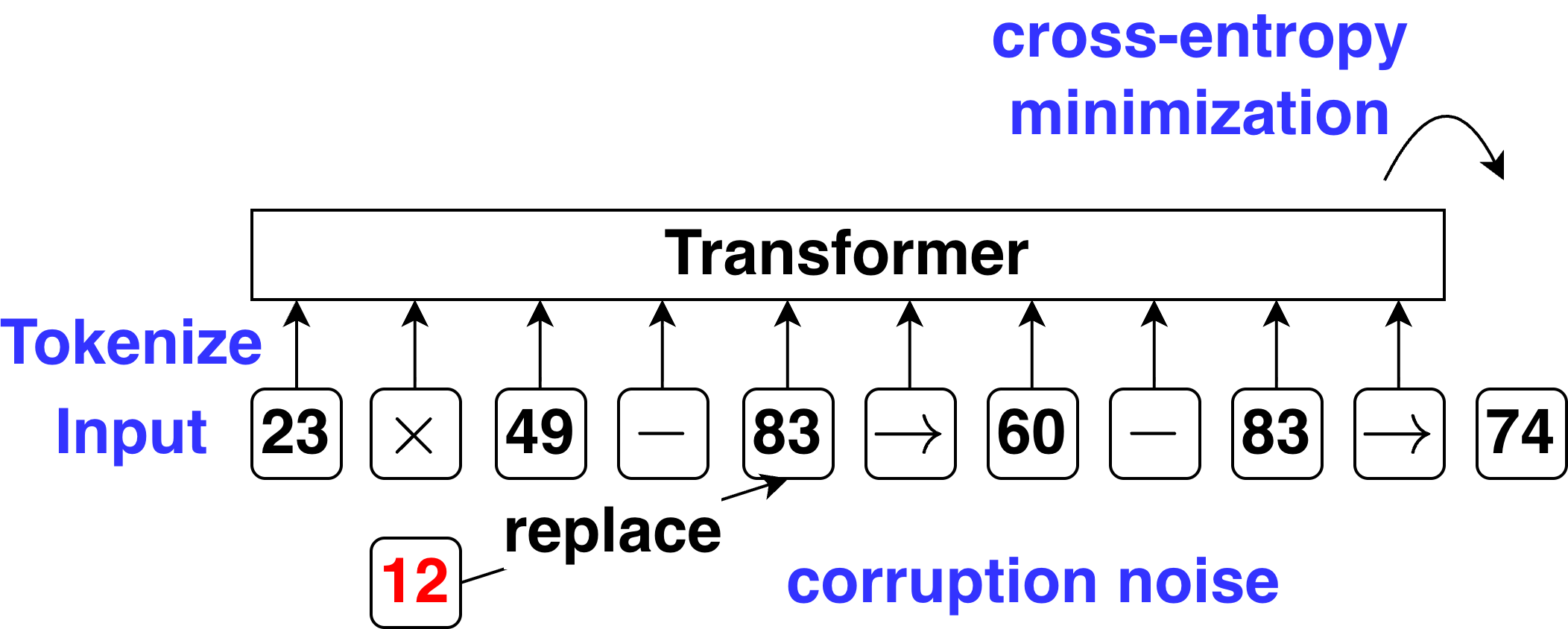}
    \caption{Illustration of our Arithmetic Expression Reasoning (AER) task for CoT reasoning. An input sequence of the format (\ref{eq:format}) is sampled and tokenized. Then a small transformer is trained from scratch on such data in the standard autoregressive fashion.}
    \label{fig:cot-example}

\end{figure}

Figure~\ref{fig:cot-example} shows one training example, which is tokenized as an input sequence to the model for next-token prediction training. In this example, we draw $a$, $b$, $c$ uniformly from $\gV_{N} := \{0,1,2,\ldots,N-1\}$ where $N=97$ is a fixed prime number, and then we determine $d$ and $o$ based on modulus $N$ arithmetic. Then, we introduce ``corruption noise'' by randomly selecting $a,b$ with probability $\veps_1$ and $d$ with probability $\veps_2$, and then replacing it with a uniform random number in $\gV_{N}$. Conceptually, the noise represents random corruption errors in large corpora and ambiguity of natural languages in describing mathematical objects. In this paper, we focus on autoregressive training (i.e., next-token prediction), which is a standard paradigm for pretraining and supervised finetuning.

Our AER task provides a natural and minimal (single reasoning step) compositional structure for CoT reasoning. Compared with prior literature on synthetic tasks, AER is akin to modular arithmetic in grokking \citep{power2022grokking}, but differs in its focus on the \textit{entire reasoning chain} (both $e_2$ and $e_3$) rather than solely on the final output. This design opens the door to analyzing reasoning composition.

\begin{figure*}[t]
    \centering
    \vspace{3mm}
    \begin{subfigure}[H]{0.62\textwidth}
        \centering
        \includegraphics[width=\linewidth]{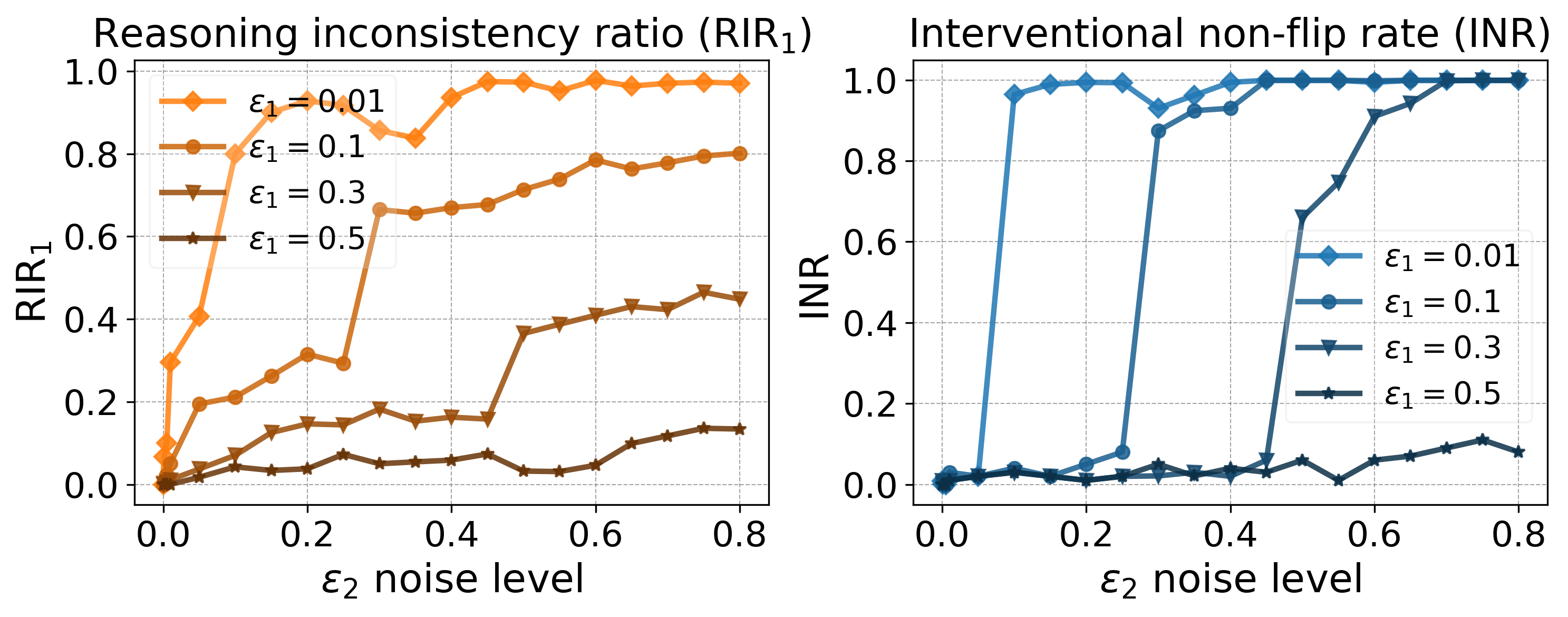}
    \end{subfigure}%
   \begin{subfigure}[H]{0.36\textwidth}
   
        \centering
        
\includegraphics[width=\linewidth]{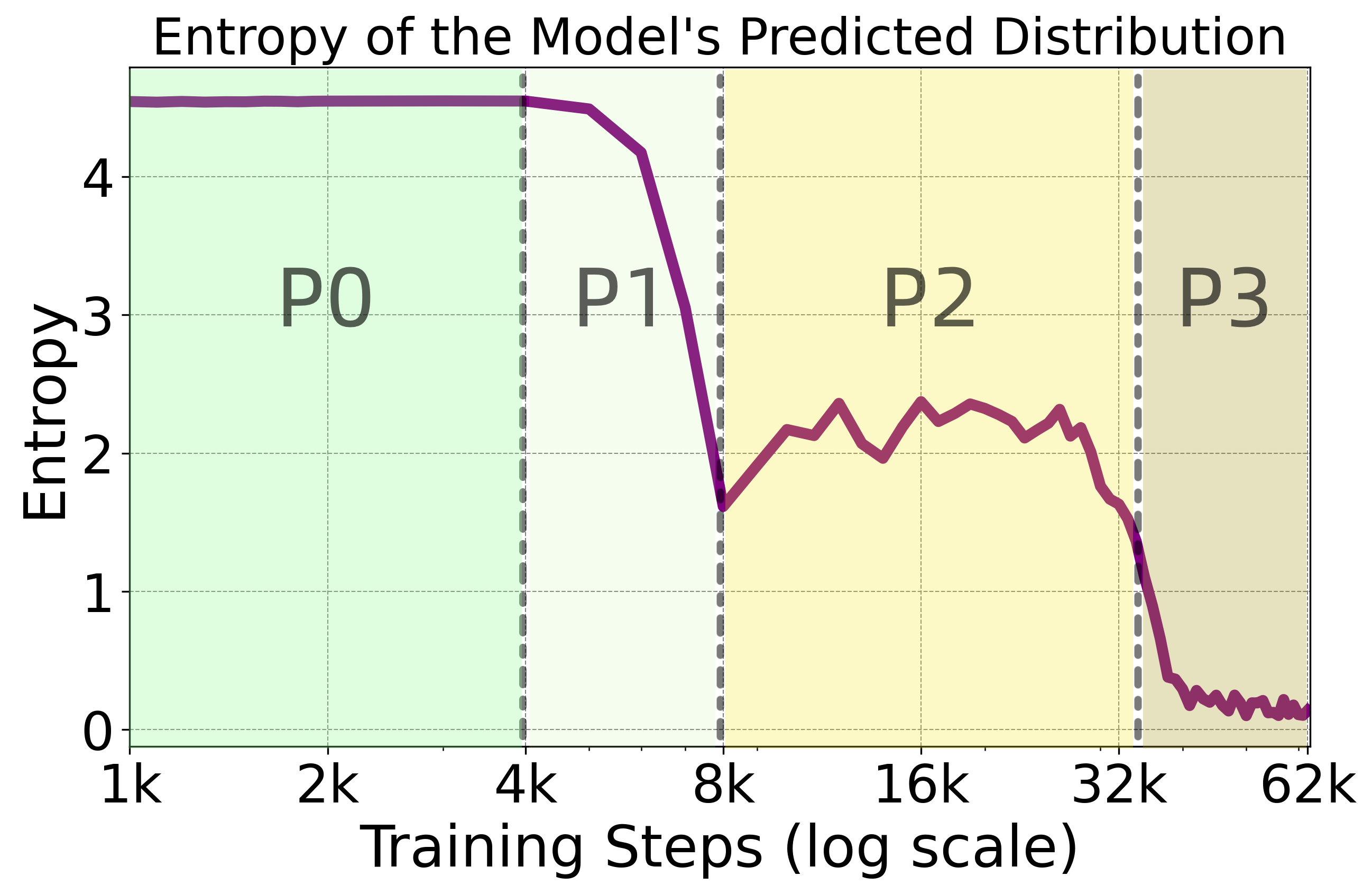}
    \end{subfigure}
    \vspace{-2mm}
    \caption{Training small transformers for the AER task. \textbf{Left/Middle: evaluating CoT reasoning under two faithfulness definitions}. We train multiple transformers separately on datasets of varying noise levels ($\veps_1$: prompt noise, $\veps_2$: reasoning noise). Both unfaithfulness metrics are small at low noise (faithful) until critical thresholds of $\veps_2$, beyond which unfaithfulness sharply increases---{\color{orange}orange curves} show generated reasoning chains become less consistent, and {\color{blue} blue curves} show interventions on reasoning barely change generated solutions.
    \textbf{Right: Prediction entropy across training}. The model experiences four phases: format following (P0), stepwise reasoning (P1), mixed reasoning (P2), and skip-step reasoning (P3). Its prediction uncertainty temporarily increases in P2 as it resolves conflicting information in the reasoning chain, which suggests implicit self-verification behavior. 
     }\label{fig:math-deduction}
   
\end{figure*}

\paragraph{Main contributions.} %We summarize our key findings below.
\begin{enumerate}
    \item \textit{A minimal testbed and formal definitions.} We introduce a synthetic task AER and formalize two faithfulness notions: mathematical consistency and counterfactual intervention on reasoning steps.
    %We present our AER task as a minimal testbed for CoT reasoning and provide two mathematical definitions of faithfulness---one is based on mathematical consistency, the other on intervention of reasoning steps.
    \item \textit{A faithfulness threshold from simplicity bias.} We find that reasoning faithfulness is possible only when training noise is below a critical threshold, a phenomenon explained by algorithmic simplicity bias.
    \item \textit{Three reasoning modes and emergent self-verification.} Under moderate noise, training dynamics exhibit three distinct reasoning modes, with a transition from faithful stepwise reasoning to unfaithful skip-step reasoning through an intermediate mixed mode marked by a transient rise in prediction entropy, indicating emergent implicit self-verification.
%We find three distinct reasoning modes under moderate noise levels: across training, the model transitions from faithful stepwise reasoning to eventual skip-step reasoning. Notably, an intermediate mixed reasoning mode emerges with prediction entropy momentarily increasing, an indication of the development of implicit self-verification.
\end{enumerate}

\section{Setup and Definitions}

We will train small transformers to solve the AER task. Following the task setup in the grokking paper \citep{power2022grokking}, we fix the modulus to be a prime number $N = 97$ for most experiments in the paper. 
Our input sequences are strings consisting of integers from $\gV_N = \{0,1,\ldots,N-1\}$, operators $\{+,-,\times\}$, and the deduction sign $\{\to\}$ of the format (\ref{eq:format}). The vocabulary is $\{0,1,\ldots,N-1, +, -, \times, \to\}$, with $N+4$ tokens in total. 

By default, we use standard transformer architecture with 3 layers, 2 heads, 128 embedding dimensions, and rotary positional embedding (RoPE) \citep{su2024roformer}. The models are trained with AdamW optimizer \citep{loshchilov2018decoupled} with learning rate 0.001, weight decay 0.01, batch size 512, using the standard autoregressive loss. See Section~\ref{sec:append-exp} for experiment details. All experiments are trained with an A30 GPU. We also consider different variants of experiments where we alter the modulus $N$, the model size, and the data format, and training sample size. The results are largely the same, which can be found in Section~\ref{sec:append-additional}. 

\subsection{Data Generation}

Let us describe the generation of one training example. First, we generate a noiseless chain by concatenating the components $e_1^0$, $\to$, $e_2^0$, $\to$, $e_3$ following the rules below: $e_1^0$ contains $a$, $\op_1$, $b$, $\op_2$, $c$ where $a,b,c$ are uniform integers from $\gV_N$, $\op_1$ is fixed as $\times$, and $\op_2$ is uniformly samples from $\{+, -\}$; $e_2^0$ and $e_3 \in \gV_N$ are an expression and an integer respectively, by evaluating $e_1^0$ under modulus $N$.
%\begin{itemize}
%    \item $e_1^0$ is an expression containing randomly drawn numbers and operators: $a$, $\op_1$, $b$, $\op_2$, $c$, where $a,b,c$ are uniformly and independently sampled from $\gV_N$, $\op_1$ is fixed as $\times$, and $\op_2$ is uniformly samples from $\{+, -\}$.
%    \item $e_2^0$ is an expression containing $d$, $\op_2$, $c$, where $d = a \times b \mod N$ (correct evaluation of modular multiplication) and $\op_2$, $c$ are the same operator and integer as in $E_1$.
%    \item $e_3 \in \gV_N$ is an integer (same as $o$) obtained by correctly evaluating $e_2$ modulo $N$ (modular addition/subtraction).
%%\end{itemize}

%Next, for given noise levels $\veps_1, \veps_2 \in [0, 1]$, we introduce corruption noise into the expressions: (i) with probability $\veps_1$ we replace either $a$ or $b$ in $e_1^0$ with equal chance with a uniform random integer in $\gV_N$, and we denote the new expression by $e_1$; (ii) with probability $\veps_2$ we replace $e_2^0$ by another independent uniform integer, and we denote the new expression by $e_2$. 
Next, for given noise levels $\veps_1, \veps_2 \in [0,1]$, we inject corruption noise into the
expressions as follows:
(i) in $e_1^0$ we select either $a$ or $b$ with equal probability and, with probability $\veps_1$,
replace the selected variable by a uniformly random integer in $\gV_N$, obtaining a new expression
denoted by $e_1$;
(ii) with probability $\veps_2$, we replace $d$ in $e_2^0$ by an independent uniformly random
integer in $\gV_N$, obtaining a new expression denoted by $e_2$.
As an input sequence to the transformer, the resulting chain $e_1 \to e_2 \to e_3$ consists of $11$ tokens.   Note that $e_1 = e_1^0$ with probability $1-\veps_1$ and $e_2 = e_2^0$ with probability $1-\veps_2$, so a fraction of training examples contains corruption noise. 
\vspace{-4pt}
\paragraph{Training distribution.} Let $\gE_1,\gE_2,\gE_3$ be the space of all possible expressions for $e_1$, $e_2$, $e_3$ respectively. Thus, the training distribution $\gP$ is the joint distribution of $(e_1,e_2,e_3)$ over $\gE_1 \times \gE_2 \times \gE_3$ as described in the data generation. The total number of training examples is 2,000,000.

\paragraph{Arithmetic rule as composition.} Let $f_1: \gE_1 \to \gE_2$ be the correct deduction and $f_2:\gE_2 \to \gV_N$ be the correct solution. For example, if $e_1 = 2 \times 9 - 19$, $e_2 = 18 - 19$, and $e_3 = 96$, then
$f_1(e_1) = e_2$ and $f_2(e_2) = e_3$.
CoT reasoning explicitly expresses the arithmetic rule as the composition $f_2 \circ f_1$, denoted as $f$. 
The goal is to learn the reasoning chain: given $e_1 \in \gE_1$, the model ideally generates
\begin{equation}
    e_1 \to f_{1}(e_1) \to \underbrace{f_2(f_{1}}_{f}(e_1)).
\end{equation}
We say that the chain $e_1 \to e_2 \to e_3$ is consistent if $e_2 = f_1(e_1), e_3 = f_2(f_1(e_1))$, i.e., satisfying the arithmetic rule as stated above. We write a reasoning chain equivalently $e_1 \to e_2 \to e_3$ as a tuple $(e_1,e_2,e_3)$.

\subsection{Two Definitions of Faithful Reasoning}

A transformer is a probabilistic model $p$ that gives a probability distribution given a context. Specifically, the conditional probabilities $p(e_2|e_1)$ and $p(e_3|e_1,e_2)$ characterize the model's behavior. Let $\gF_1$ be the function space that maps from $\gE_1$ to $\gE_2$, and $\gF_2$ be the function space that maps from $\gE_1 \times \gE_2$ to $\gV_N$. We define two random functions $\widehat f_1, \widehat f_2$, respectively, as $\gF_1$-valued and $\gF_2$-valued random elements satisfying, for all $e_1,e_2,e_3$,
\begin{align*}
     &\P\big( \widehat f_1( e_1) = e_2 \big) = p( e_2 | e_1), \\
     &\P\big( \widehat f_2(e_1, e_2) = e_3 \big) = p( e_3 | e_1, e_2).
\end{align*}
In reality, due to multiple factors (e.g., noise, underfitting, overfitting), we expect $\widehat f_1, \widehat f_2$ to deviate from $f_1, f_2$ with positive probabilities, i.e., not accurately learning the (groundtruth) arithmetic rules of $f_1$ and $f_2$.

\paragraph{Consistency-based faithfulness.} For a test prompt $e_1$, we generate a chain $(e_1, e_2', e_3')$ from a given  sampling scheme (e.g., greedy decoding or sampling under a certain temperature). A common way of characterizing the reasoning behavior is to compare $(e_1, e_2', e_3')$ with the consistent groundtruth chain $(e_1, e_2, e_3)$ (Table~\ref{tab:cat}). We say the CoT reasoning is faithful if $e_2 = e_2'$ and $e_3 = e_3'$ for every prompt $e_1$. 

This definition is an observational notion of faithfulness without causal dependence.
%It is simple for evaluation as long as we can draw prompts from a distribution and generate $e_2', e_3'$ according to a sampling scheme.
While conceptually simple, it suffers from two drawbacks: (i) The set of consistent chains has cardinality $|\gE_1|$, which is a small fraction of the set of all possible chains with cardinality $|\gE_1| \times |\gE_2| \times |\gE_3|$. Thus, we have no out-of-distribution robustness guarantees. (ii) Even when a model generates a consistent chain, we are unable to distinguish genuine step-by-step reasoning from post hoc explanation, where the latter lacks expected causal links. %In the latter case, generation of $e_3$ does not depend on $e_1$ and $e_2$ is effectively an explanation of $e_3$ rather than serving as a genuine step. 

\begin{table}[t]
\caption{\textbf{Consistency-based reasoning categorization.}
Given a test instance $e_1$, compare the model chain $(e_1,e'_2,e'_3)$
with the ground-truth chain $(e_1,e_2,e_3)$.
Here $e'_2$ is the reasoning step and $e'_3$ is the solution; R $=$ reasoning, S $=$ solution.}
\label{tab:cat}
\centering
\small
\setlength{\tabcolsep}{7pt}
\renewcommand{\arraystretch}{1.15}
\begin{tabular}{ccc}
\toprule
\textbf{Type} &  \textbf{Consistency check} & \textbf{Outcome} \\
\midrule
Faithful  &
$e'_2 = e_2,\; e'_3 = e_3$ &
R correct; S correct \\
\midrule
\multirow{3}{*}{Unfaithful }
 &
$e'_2 \neq e_2,\; e'_3 = e_3$ &
R incorrect; S correct \\
 &
$e'_2 = e_2,\; e'_3 \neq e_3$ &
R correct; S incorrect \\
 &
$e'_2 \neq e_2,\; e'_3 \neq e_3$ &
R incorrect; S incorrect \\
\bottomrule
\end{tabular}

\end{table}

% ==== end inlined floats/faithful-types.tex ====
\paragraph{Intervention-based faithfulness.} We provide a stronger faithfulness definition by considering counterfactuals of $e_2$
so that $e_1$ and $e_2$ are not necessarily consistent, especially in low-noise settings. This definition requires intervening on the reasoning $e_2$ and observing the resulting change in solution $e_3$---in particular, if $e_1$ and $e_2$ are not consistent, how much predicting $e_3$ depends on $e_1$ and $e_2$. The following definition characterizes two extreme modes.

\begin{definition}\label{def:reasoning}
    (i) \textbf{(Perfect) stepwise reasoning}: We say that a model's CoT reasoning follows (perfect) stepwise reasoning if the following holds with probability $1$: $\widehat f_2(e_1,e_2) = f_2(e_2)$ for all $e_1 \in \gE_1, e_2 \in \gE_2$. (ii) \textbf{(Perfect) skip-step reasoning}: We say that a model's CoT reasoning follows (perfect) skip-step reasoning if the following holds with probability $1$: $\widehat f_2(e_1, e_2) = f(e_1)$ for all $e_1 \in \gE_1, e_2 \in \gE_2$.
\end{definition}

Perfect step-by-step reasoning implies that $\widehat f_2$ depends solely on the reasoning expression $e_2$ regardless of $e_1$. Similarly, perfect skip-step reasoning implies that $\widehat f_2$ only depends $e_1$ and ignores the reasoning expression $e_2$. Under this definition, a model's CoT reasoning is perfectly faithful if it follows perfect stepwise reasoning and $\widehat f_1(e_1) = f_1(e_1)$ for all $e_1 \in \gE_1$ with probability 1. Clearly, this intervention-based faithfulness implies consistency-based faithfulness. %Evaluation of faithfulness under this definition requires intervention on reasoning expression $e_2$.

\subsection{Metrics}

%By default, as we train the models from scratch, we evaluate the models using a suite of metrics every 1000 training steps on a separately held test examples. Our evaluation metrics in this subsection are model-agnostic and only assume black-box access to the models. 

During training, we evaluate the models using a suite of metrics every 1,000 training steps on 1,000 held-out test examples. We denote the generated reasoning chains by $(e_2', e_3')$ given $e_1$ under a specified decoding scheme.
%using either greedy decoding or under a sampling temperature.

%\paragraph{Standard metrics.} The training loss is the autoregressive loss calculated based on a sample of $1,000$ training examples, and the test loss is calculated on a held-out sample of $1,000$ test examples from the same distribution. 

%We also calculate the training/test errors using a sample of $1,000$ train/test examples. Specifically, given the prompts from the these examples, we use greedy decoding to generate the reasoning chains $(e_2', e_3')$ and count as error if $e_2' \neq e_2$ or $e_3' \neq e_3.$
\paragraph{Consistency-based metrics.} We calculate the proportions of the four categories from Table~\ref{tab:cat} by sampling $n = 1,000$ test prompts $(e_{i,1})_{i\le n}$ following the training distribution and use the model to generate $(e_{i,2}', e_{i,3}')$ under a fixed the temperature $1$. 
%For example, the consistency-based faithfulness is 
%\begin{equation*}
%\text{Consistency rate}:= \frac{1}{n} \sum_{i=1}^{n} \mathbf{1}\{e_{i,2}'=e_{i,2}, e_{i,3}' = e_{i,3} \},
%\end{equation*}
%where $(e_{i,1}, e_{i,2}, e_{i,3})$ forms a consistent chain. Ideally, this score equals 1 if the model's CoT reasoning is perfectly faithful. 
We mainly consider two metrics of \textit{reasoning inconsistency ratio (RIR)}:
\begin{align}
\mathrm{RIR_1} &:= \frac{\sum_{i \le n} \mathbf{1}\{e_{i,2}' \neq e_{i,2}, e_{i,3} = e_{i,3}'\}}{\sum_{i \le n}\mathbf{1}\{e_{i,2}' \neq e_{i,2}\}}   \label{def:RIR} \\
\mathrm{RIR_2} &:= \frac{\sum_{i \le n} \mathbf{1}\{e_{i,2}' \neq e_{i,2}, e_{i,3} = e_{i,3}'\}}{\sum_{i \le n}\mathbf{1}\{e_{i,3}' = e_{i,3}\}}.
\end{align}
The two metrics differ in the denominators and are complementary in quantifying unfaithfulness (higher means more unfaithful). Intuitively, $\text{RIR}_1$ measures the proportion of recovering a correct solution $e_3$ when it generates incorrect reasoning $e_2$; and $\text{RIR}_2$ measures the proportion of incorrect reasoning among solution-correct test samples.  

\paragraph{Intervention-based metrics.} We will apply interventions to reasoning $e_2$ and measure how well a model fits into the two categorization in Definition~\ref{def:reasoning}. We will focus on \textit{uniformly random intervention}, where we replace $d$ in $e_2$ by a uniform random variable in $\gV_N$, which yields $\tilde e_2$. For $n$ independent test prompts $(e_{i,1})_{i \le n}$, model's generated reasoning $(e_{i,2}')_{i \le n}$, and the replaced reasoning $(\tilde e_{i,2})_{i \le n}$, we calculate the two metrics: Interventional distribution sensitivity (IDS) and Interventional non-flip rate (INR):
\begin{align*}
\mathrm{IDS}   &:= \frac{1}{n} \sum_{i=1}^{n}  \KL \big(\widehat f_2(e_{i,1},e_{i,2}'), \widehat f_2(e_{i,1}, \tilde e_{i,2}) \big); %\label{def:IDS} 
\\
\mathrm{INR} &:= \begin{small} \frac{1}{n} \sum_{i=1}^{n} \mathbf{1}\big\{\argmax_{e \in \gV_N} p(e|e_{i,1}, e_{i,2}')=\argmax_{e \in \gV_N} p(e|e_{i,1}, \tilde e_{i,2})  \big\}. %\label{def:INR}
\end{small}
\end{align*}%
where $\KL$ means KL divergence. A small IDS and a large INR mean weak causal effects of $e_2$ on $e_3$, which are indications of skip-step reasoning. Conversely, a large IDS and a small INR are evidence of stepwise reasoning.

%In addition, we will consider similar KL divergence metrics under two additional interventions. (i) Solution-consistent intervention: we place $e_2$ by a uniform $\tilde e_2$ conditioning on $f_2(e_2) = e_3$. This intervention examines spurious reasoning that is consistent with $e_3$ but not with $e_1$. (ii) Adversarial intervention: we search for $e_2$ by maximizing $\KL \big(\widehat f_2(e_{i,1},e_{i,2}), \widehat f_2(e_{i,1}, \tilde e_{i,2}) \big)$, which quantifies worst-case counterfactuals. 

\paragraph{Prediction uncertainty metrics.} We measure a model's uncertainty with prediction entropy: for test prompts $(e_{i,1})_{i \le n}$ and the model's generated chains $(e_{i,2}', e_{i,3}')_{i \le n}$, we calculate prediction entropy (PE): 
\begin{align}\label{def:entropy}
    &\mathrm{PE} := \frac{1}{n}\sum_{i=1}^{n} \sum_{e \in \gV_N} 
     - p(e|e_{i,1},e_{i,2}') \log \,p(e|e_{i,1},e_{i,2}').
\end{align}

\section{Main Results}

\subsection{How Does Noise Induce Unfaithfulness?}

We construct training data of varying noise levels $\veps_1$ and $\veps_2$. 
In particular, the training data is noiseless when $\veps_1 = \veps_2 = 0$. For each combination of $(\veps_1, \veps_2)$, we train a transformer for 62,500 steps and evaluate the model at the final checkpoint. 

Our aim is to investigate factors, particularly noise levels, that impact faithfulness of reasoning. While we anticipate that lower noise levels improve consistency-based faithfulness, it is a priori unclear whether intervention-based faithfulness is possible at all. Consider training a model with noiseless training data ($\veps_1=\veps_2=0$). Observing only consistent chains, the model could theoretically learn the solution $\widehat f_2(e_1,e_2)$ as $f_1(e_1)$, $f_2(e_2)$, or a mix of the two---all of which would achieve zero loss and perfect consistency-based faithfulness. Yet, only learning $f_2(e_2)$ satisfies intervention-based faithfulness criteria. Thus, we ask the question---\textit{Is intervention-based faithfulness possible at all? If so, what are contributing factors?}

\paragraph{Intervention-based faithfulness is possible below a critical noise level.} We find that faithful reasoning is achievable, but only when the noise level is sufficiently small. In Figure~\ref{fig:math-deduction} (left/middle), we consider prompt noise level $\veps_1 \in \{0.01, 0.1, 0.3, 0.5\}$ and vary the reasoning noise level $\veps_2 \in \{0.001, 0.005, 0.01, 0.05, 0.1, \\
0.3, 0.5, 0.9\}$. 
%The model trained for each noise-level combination are evaluated through two metrics measuring unfaithfulness of reasoning: 
We measure unfaithfulness with two metrics, namely consistency-based $\mathrm{RIR}_1$ and intervention-based INR, and find that both metrics are non-decreasing as the noise level increases.

Moreover, Figure~\ref{fig:math-deduction} reveals the existence of a critical threshold $\tau_c(\veps_1)$ exists in $\veps_2$ for each $\veps_1$. Indeed, both unfaithfulness metrics are approximately flat and close to 0 below the threshold $\tau_c(\veps_1)$, and increase dramatically beyond the thresholds. This suggests that competition between $e_1$ and $e_2$ as predictive signals for lowering loss largely determines reasoning behavior at the final checkpoint.

%A natural explanation for the sharp transitions is that, $e_1$ and $e_2$ offer compet by minimizing the loss, the model learns to use either expressions $e_1$ or $e_2$, whichever has less noise. 

Interestingly, both unfaithfulness metrics are close to 0 under the noiseless setting $(\veps_1=\veps_2=0)$, indicating that faithful reasoning is achievable when training only on consistent chains. More generally $\veps_1 < \tau_c(\veps_1)$, which means that at low noise, models exhibit inertia: it continues to reply on stepwise reasoning rather than immediately switching to skip-step reasoning for a slightly lower loss. This motivates a study of algorithmic effects  beyond noise levels alone.

%namely that the critical threshold of $\veps_2$ is larger than $\veps_1$. This phenomenon can not be solely explained by loss minimization, as skip-step reasoning would yield perfect accuracy as well. At low noise, there is a tendency towards faithful reasoning. 

%achieves high intervention-based faithfulness when trained only on consistent chains (noiseless data)

%an accuracy-comptradeoff likely determines the reasoning behavior. Figure~\ref{fig:math-deduction} shows that $\veps_1 < \tau_c(\veps_1)$, namely that the critical threshold of $\veps_2$ is larger than $\veps_1$---a phenomenon that can not be fully explained by the statistical (loss minimization) narrative. This is because the prompt $e_1$ contains less noise than reasoning $e_2$ when $\veps_2 \in (\veps_1, \tau_c(\veps_1))$, yet the trained model preserves highly faithful reasoning. The model's inertia of using stepwise reasoning, instead of immediately switching to skip-step reasoning with a lower loss, is an evidence of algorithmic effects. This motivates us to examine the computational aspect via the lens of simplicity bias.

%Overall, Figure~\ref{fig:math-deduction} shows that transformers with autoregressive training can be faithful under the stronger intervention-based definition as long as the noise level is low. In particular, the model achieves high intervention-based faithfulness when trained only on consistent chains (noiseless data) without interventional data.

\paragraph{Faithfulness benefits from simplicity bias (low-complexity preference).}
%Why is CoT reasoning highly faithful at low-noise levels? This puzzle is pronounced in the noiseless regime, where perfect stepwise reasoning and perfect skip-step reasoning would both yield $100\%$ prediction accuracy; yet the model favors stepwise reasoning. 
We hypothesize that the training algorithm favors learning a simpler rule (stepwise reasoning involving a single operator) over  a more complex rule (skip-step reasoning involving two operators) if both mechanisms yield comparable accuracy. This hypothesis is based on Occam's razor and recent theoretical study of inductive bias in overparametrized neural networks \citep{soudry2018implicit, montanari2022interpolation, abbe2024generalization}, where models prefer smooth, low-degree, and low-complexity functions. 

\begin{figure}[t]
    \centering
    \begin{minipage}{0.4\columnwidth}
        \centering
        \includegraphics[width=\linewidth]{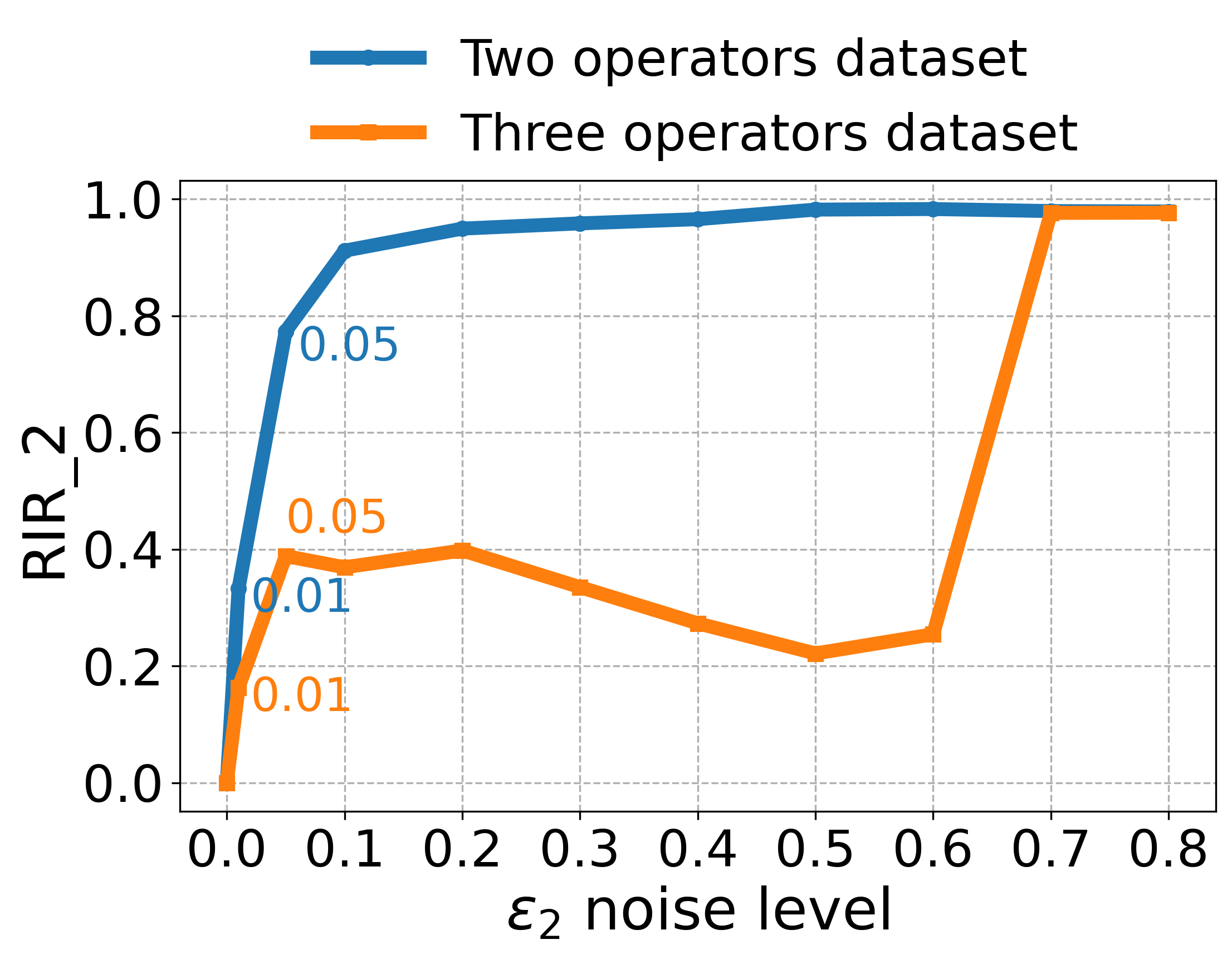}
    \end{minipage}
    \begin{minipage}{0.4\columnwidth}
        \centering
        \includegraphics[width=\linewidth]{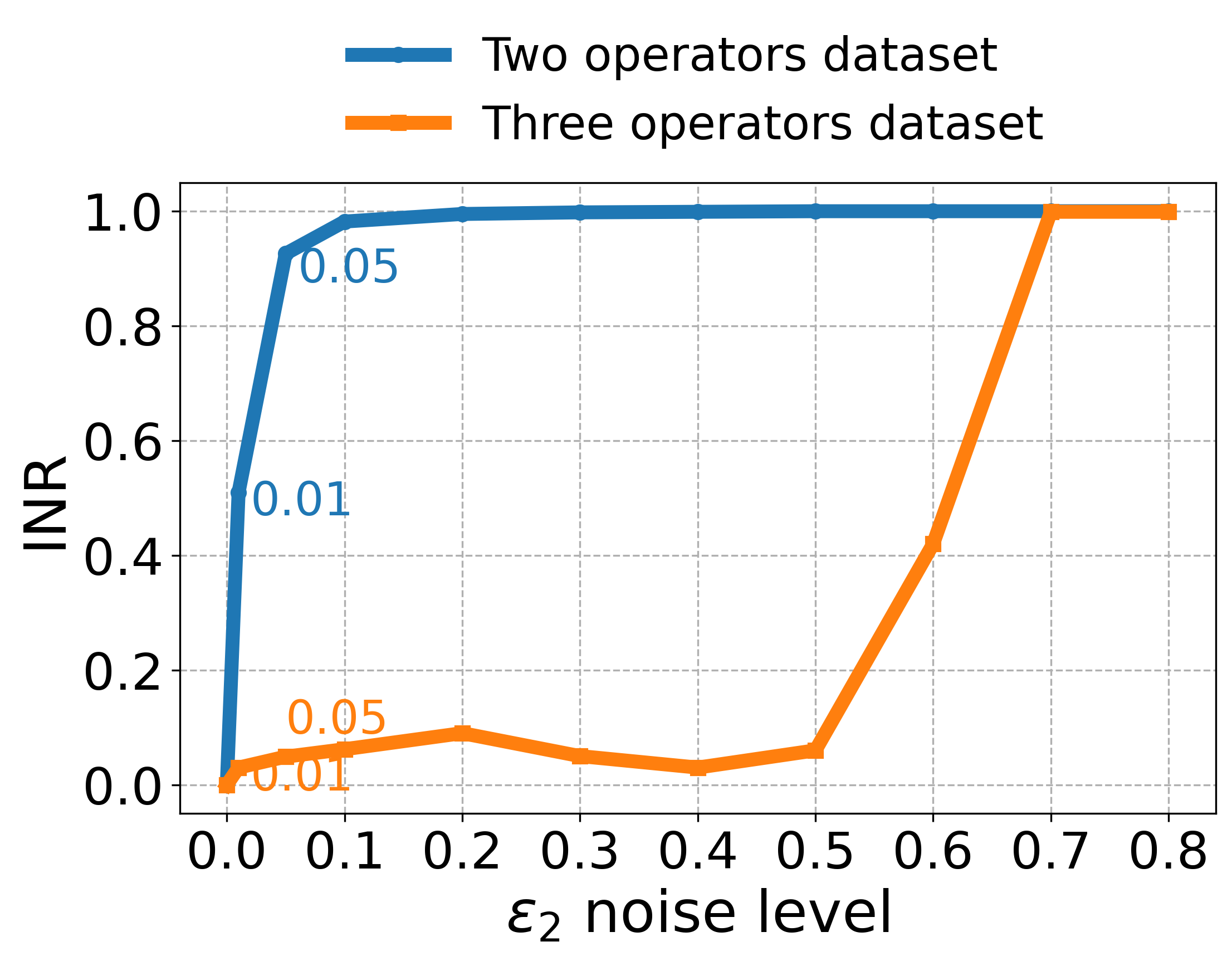}
    \end{minipage}
    \caption{\textbf{Larger complexity gap between $e_1$ and $e_2$ reduces unfaithfulness.} Fixing $\varepsilon_1=0.01$, we sweep $\varepsilon_2$ and compare training transformers on two-operator prompts $e_1$ and three-operator prompts. 
    Both metrics indicate that larger difficulty gaps encourage stepwise reasoning and reduce unfaithfulness. 
    }
    \label{fig:complexity_gap}
   
\end{figure}

To validate this hypothesis, we modify the experiment to increase the complexity gap between the prompt and reasoning: in data generation, we replace $e_1^{0}=a \times b-c$ with $(a + b) \times c - d$, and inject noise in $e_1$ by randomly replacing $a$, $b$, or $c$ at noise level $\varepsilon_1$ while the noise injection scheme for $e_2$ remains unchanged; we also changed the modulus to $N=29$. This modification increases the reasoning complexity of $e_1\to e_2$ which evaluates 2 operators in one step. Figure~\ref{fig:complexity_gap} shows that a larger complexity gap between $e_1$ and $e_2$ reduces unfaithfulness, and that models tolerate higher reasoning noise before the sharp transition to skip-step reasoning.
%Figure~\ref{fig:complexity_gap} shows that under training data with $\varepsilon_1 = 0.01$, the model transitions to skip-step reasoning (increasing reliance on $e_1$) at around $\tau_{c}(\varepsilon_2) \approx 0.6$ (with a prediction accuracy gain of about $0.6$). This stands in contrast to training on the original data, where RIR and INR sharply increase around $\tau_{c}(\varepsilon_2) \approx 0.05$ (where the prediction accuracy gain is approximately $0.04$).
The results demonstrate that a larger complexity gap between successive steps in the reasoning chain induces a stronger simplicity bias, which creates an algorithmic barrier for the model to develop unfaithful skip-step reasoning.

Finally, we show that the complexity gap between reasoning steps---not the relative positions---forms the algorithmic inductive bias, thereby excluding a confounding factor. Specifically, we swap the positions of the prompt and the reasoning, so that the modified chain becomes $e_2\to e_1 \to e_3$. Section~\ref{sec:append-swap} shows that reasoning faithfulness is determined by complexity, and not by relative positions.
%Table~\ref{tab:complexity_gap} in Section~\ref{sec:append-additional} reports four unfaithfulness measures on clean (noise-free) training data before and after the swap. The intervention-based metrics (IDS and INR) remain almost unchanged, indicating that the effect is not driven by sequence order: the model intrinsically prefers the simpler $e_2$ for prediction. The two RIR scores drop from $1$ in the original format to $0$ after swapping since relying on $e_2$ becomes ``skip-step'' reasoning under the swapped order.

\begin{imp}
\textit{Low noise in training data and large complexity gaps in reasoning steps are critical to inducing high faithfulness in CoT reasoning.}
\end{imp}

\subsection{Phases of Reasoning Modes Across Training}\label{sec:phases}

%Now we investigate the model's behavior throughout the training process, aiming to understand the emergence of unfaithfulness of CoT reasoning. We focus on the model's prediction on solution $e_3$ given both $e_1$ and $e_2$.

\begin{figure*}[t]
    \centering
    \hspace*{7mm}
\includegraphics[width=0.95\textwidth]{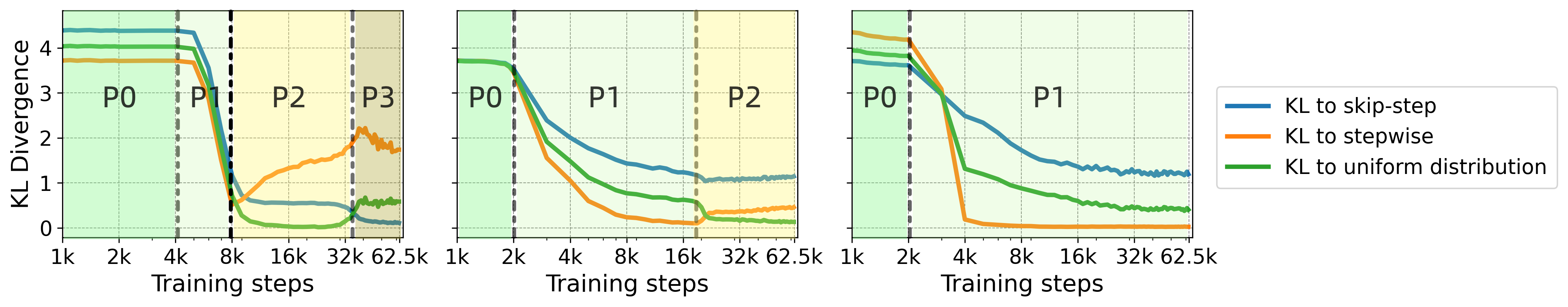}
    \caption{\textbf{Training dynamics exhibit distinct phases of reasoning modes.} We conduct three experiments under different noise configurations and evaluate models with three metrics (logarithmic x-axis).
\textbf{Left}: Noise levels $(\varepsilon_1, \veps_2) = (0.01, 0.1)$. The model exhibits four distinct reasoning phases, validating our proposed phase characterization.
\textbf{Middle}: $(\varepsilon_1, \veps_2) = (0.1, 0.1)$. The model exhibits only the first three phases, as equal noise levels incentivizes the model to stay in the mixed reasoning mode P2.
\textbf{Right}: $(\varepsilon_1, \veps_2) = (0.1, 0.01)$. The model exhibits only the first three phases, as lower prompt noise incentivizes stepwise reasoning P1.}

\label{fig:3phase_transition_pattern}
\end{figure*}

\paragraph{Phase characterization of reasoning modes.} We find that a model's reasoning mode evolves across training in a single experiment.
For illustration, we fix the noise level to be $\varepsilon_1=0.01,\varepsilon_2=0.1$ 
to generate the training data and train the model for 62,500 steps. We are mostly interested in the model's probability distribution of $\widehat f_2(e_1, e_2)$ across training. We hypothesize that the model is characterized by the following four phases with respect to training step $t$. Let $(e_1,e_2)$ follows the same distribution as the training data.
\vspace{-5pt}
\begin{itemize}
\setlength\itemsep{-0.2em}

    \item \textit{Phase 0 (format following)}: predicting operator token ``$+$'' or ``$-$'' and deduction ``$\to$'' in correct positions.
    \item \textit{Phase 1 (stepwise reasoning)}: prediction depends mostly on $e_2$ in the sense of $\widehat f_2(e_1, e_2) \approx f_2(e_2)$.
    \item \textit{Phase 2 (mixed reasoning)}: Denote the indicator $c(e_1,e_2) = \mathbf{1}\{f(e_1) = f_2(e_2)\}$. Then
    $\widehat f_2(e_1, e_2) \approx f_2(e_2) c(e_1,e_2) + U (1-c(e_1,e_2))$
    where $U$ is a uniform random variable in $\gV_N$. That is, prediction is most uncertain (uniform distribution) if $e_1$ and $e_2$ are inconsistent, and consistent with both $e_1, e_2$ otherwise.
    %\AAA{I think adding an additional sentence to explain the intuition behind this formulation will be helpful to the reader."}
    \item \textit{Phase 3 (skip-step reasoning)}: prediction depends mostly on $e_1$ in the sense of $\widehat f_2(e_1, e_2) \approx f(e_1)$.
\end{itemize}
\vspace{-5pt}
We validate the hypothesized characterization using several evaluation metrics (Figures~\ref{fig:3phase_transition_pattern} and~\ref{fig:Format_score}). (i) \textbf{Format score}: we generate the full reasoning traces and solutions given test prompts, and calculate the rate of generating operators and deduction tokens at correct positions. (ii) \textbf{KL to stepwise}: we calculate the KL divergence between the model's prediction probability and perfect stepwise reasoning $n^{-1}\sum_{i\le n}\KL(\widehat f_2(e_{i,1}, e_{i,2}'), f_2(e_{i,2}'))$ where $(e_{i,1},e_{i,2}')_{i \le n}$ is generated based on $n=1,000$ test prompts $e_{i,1}$. (iii) \textbf{KL to skip-step:} we calculate the KL divergence between the model's prediction probability and perfect skip-step reasoning $n^{-1}\sum_{i\le n}\KL(\widehat f_2(e_{i,1}, e_{i,2}'), f(e_{i,1}))$ where $(e_{i,1},e_{i,2}')_{i \le n}$ is similarly sampled and generated. (iv) \textbf{KL to uniform distribution:} we calculate the KL divergence between the model's prediction distribution and a uniform distribution $n^{-1}\sum_{i\le n}\KL(\widehat f_2(e_{i,1}, e_{i,2}'), U)$ %where $U$ is a uniform random variable, 
evaluated only on inconsistent chains where $f(e_1) \neq f_2(e_2)$. We note reasoning modes in Phase~1--3 have causal interpretations; see Section~\ref{sec:append-phase} and Figure~\ref{fig:causal} for further explanations.

Our results support our phase characterization. Figure~\ref{fig:Format_score} in Section~\ref{sec:append-phase} shows that the model quickly learns format following in Phase 0. In Figure~\ref{fig:3phase_transition_pattern} (left) P1, the orange curve quickly descends to the minimum and is the smallest among the three KL divergence metrics, which indicates the model's preference for stepwise reasoning in the early phase of training. In P2, the orange curve rises while the blue curve drops to approximately 0. This indicates that the model moves away from stepwise reasoning, and learns to predict a uniform distribution (least informative) for inconsistent chains $(e_1, e_2)$. In P3, the blue curve further descends to its minimum and is the smallest among the three metrics, which indicates that the model eventually learns skip-step reasoning driven by loss minimization. 

We also find that under different noise configurations, models may only reach Phase 1 or Phase 2. Figure~\ref{fig:3phase_transition_pattern} (middle) shows that models trained with equal noise levels eventually stay in the mixed reasoning mode, since combining both the prompt and the reasoning expressions yields the smallest loss. Figure~\ref{fig:3phase_transition_pattern} (right) shows another experiment with a smaller reasoning noise, which discourages the model to rely on the prompt expression $e_1$.

\paragraph{Factors that determine reasoning modes.} Based on the above experiments, we hypothesize that two deciding factors of a model's reasoning mode are data quality (namely noise levels) and compute (namely training steps). We fix $\veps_1 = 0.1$ and let the reasoning noise %\AAA{Should this be reasoning noise instead since we are referring to } 
level $\veps_2$ range from $0.001$ to $0.9$. For each noise level and evaluation checkpoint during training, we classify the model's reasoning mode into P1, P2, or P3, based on whichever phase has the smallest KL divergence among the three KL metrics. Figure~\ref{fig:phase_diagram} in Section~\ref{sec:append-phase} shows that more compute and higher prompt noise generally push the model towards skip-step reasoning. As a side result, we also find that the phase boundaries are irregular and vary across different random seeds, suggesting the complexities of the optimization landscape , which echos the unpredictability of emergent phenomena in LLMs \citep{weiemergent}.

\begin{imp}
\textit{Prolonged autoregressive training for CoT reasoning (particularly supervised finetuning) may reduce faithfulness by encouraging skip-step reasoning.}
\end{imp}

\subsection{Implicit Self-Verification}

Meta-reasoning abilities such as self-reflection and self-verification are intriguing emergent skills from reasoning scaling \citep{guo2025deepseek}. We aim to closely examine mixed reasoning mode and address the following questions---
    \textit{Does the model learn to encode reasoning uncertainty? If so, what is its internal mechanism?}

\begin{figure}[t]
    \centering
\includegraphics[width=0.9\columnwidth]{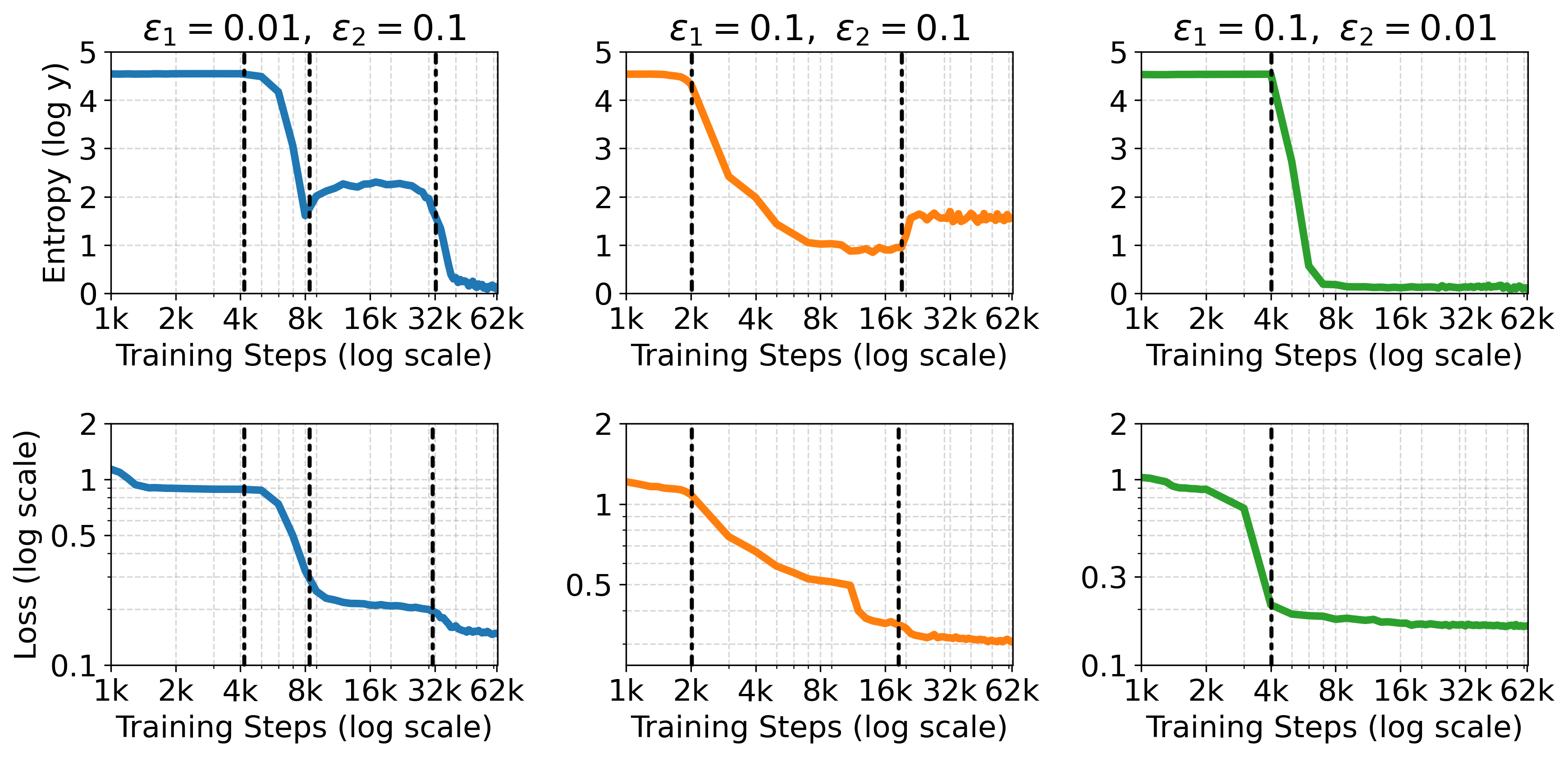}
\caption{\textbf{Prediction entropy shows a temporary ascent despite continuously decreasing training loss.} Models encode reasoning uncertainty in the mixed reasoning mode, which matches most sharp descents of the loss.}
   
\label{fig:entropy_phases}
\end{figure}

\paragraph{Prediction entropy momentarily increases during mixed reasoning mode.} We use the prediction entropy (\ref{def:entropy}) to quantify the model's uncertainty, an approach used in recent empirical papers \citep{fu2025deep, zhao2025learning}. In the presence of inconsistent chains $(e_1,e_2)$, our Phase 2 characterization of the mixed reasoning mode suggests that model outputs approach a near-uniform distribution, corresponding to maximal uncertainty. Figure~\ref{fig:entropy_phases} (top)  confirms this behavior: under the three noise configurations as in Figure~\ref{fig:3phase_transition_pattern}, prediction entropy exhibits a clear non-monotonic trend, rising as the model enters the mixed reasoning mode.

Figure~\ref{fig:entropy_phases} (bottom) shows the evolution of the training loss. Unlike prediction entropy, the loss decreases monotonically, as expected. Yet, most sharp drops in loss align closely with transitions between reasoning modes. This correspondence suggests that changes in reasoning behavior and uncertainty are reflected in the geometry of the loss landscape. 
%already encodes reasoning modes and uncertainty, an investigation of which we leave to future work.

\begin{figure}[t]
    \centering
\includegraphics[width=0.9\columnwidth]{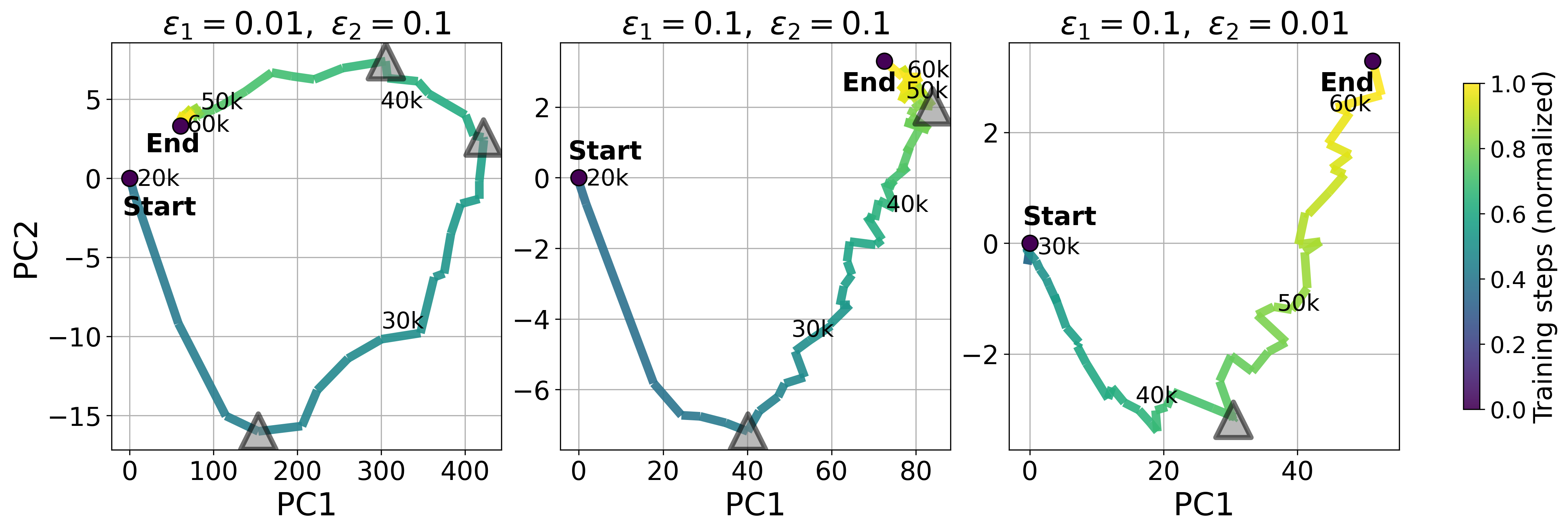}
    \caption{\textbf{PCA trajectories of hidden states visualize reasoning phases.} Under three experiment settings, we extract the top-layer hidden states, and project them to the principal subspace.
\textbf{Left}: Noise levels $(\varepsilon_1, \veps_2) = (0.01, 0.1)$. Triangles mark the changes in movement directions, which correspond to the three phase.
\textbf{Middle}: Noise levels $(\varepsilon_1, \veps_2) = (0.1, 0.1)$. Phase changes are marked by two triangles.
\textbf{Right}: Noise levels $(\varepsilon_1, \veps_2) = (0.1, 0.01)$. One triangle marks the only one directional change.}
\vspace{-5mm}
    \label{fig:pca}
\end{figure}

\paragraph{A mechanistic analysis reveals uncertainty-encoding features.} To probe how the model internally represents uncertainty, we examine the evolution of top-layer hidden states throughout training. The PCA visualization in Figure~\ref{fig:pca} shows that hidden states encode the distinct reasoning phases, with clear changes in trajectory direction aligned with phase transitions.

%Moreover, we find that certain coordinates of the hidden states encode the uncertainty features. To be specific, we calculate two metrics at each evaluation checkpoint:
Moreover, we examine whether certain coordinates of hidden states encode uncertainty features and the effects of attentions. To this end, we introduce:
{\small
\begin{align*}
    &\text{Hidden State Contrast (HSC)}:= \\
    &\quad \frac{1}{n}\sum_{i=1}^{n}\Bigl|\bigl\|h_{L,e_3}(e_{i,1}^{+},e_{i,2}^{+})\bigr\|_{\infty}
-
\bigl\|h_{L,e_3}(e_{i,1}^{-},e_{i,2}^{-})\bigr\|_{\infty}\Bigr|,
\\
&\text{Attention Contrast (AC)}:= \\
&\quad \frac{1}{n}\sum_{i=1}^{n}\bigl\|h_{L-1,i}^{+}W^{(L)}_{QK}(h_{L-1,i}^{+})^{\top}  -h_{L-1,i}^{-}W^{(L)}_{QK}(h_{L-1,i}^{-})^{\top}\bigl\|_{\mathrm{op}},
\end{align*}
}%
where $\| \cdot \|_{\mathrm{op}}$ denotes the matrix operator norm (Section~\ref{par:operator_norm}). Here, we sample consistent (noiseless) chains $(e_1^{+},e_2^{+})$ as \textbf{positive} samples, %from the noiseless dataset are treated as \textbf{positive} samples, 
and inconsistent (noisy) chains $(e^{-}_1,e^{-}_2)$ as \textbf{negative} samples with noise level $\varepsilon_1=0,\ \varepsilon_2=1$ (i.e., $e_1$ and $e_2$ always yield different results). %are treated as \textbf{negative} samples. 
We denote by $h_{L,e_3}(e^{+}_1,e^{+}_2)$ the post-FFN output vector at the solution $e_3$ position from the last layer $L$, and by $h^{+}_{L-1,i}$ the hidden-state matrix at the penultimate layer $L-1$ for the $i$-th positive sample. $h_{L,e_3}(e^{-}_1,e^{-}_2)$ and $h^{-}_{L-1,i}$ are defined analogously. These two quantities measure the degree of differentiation between positive and negative samples, %the model differentially processes samples where $e_1$ and $e_2$ are consistent versus inconsistent. 
with larger values indicating stronger internal differentiation that functions like consistency checks.
%Large values suggest  internal consistency checks within the model.  %that the model performs internal self-verification.

\begin{figure}[t]
\vspace{3mm}
    \centering
\includegraphics[width=0.9\columnwidth]{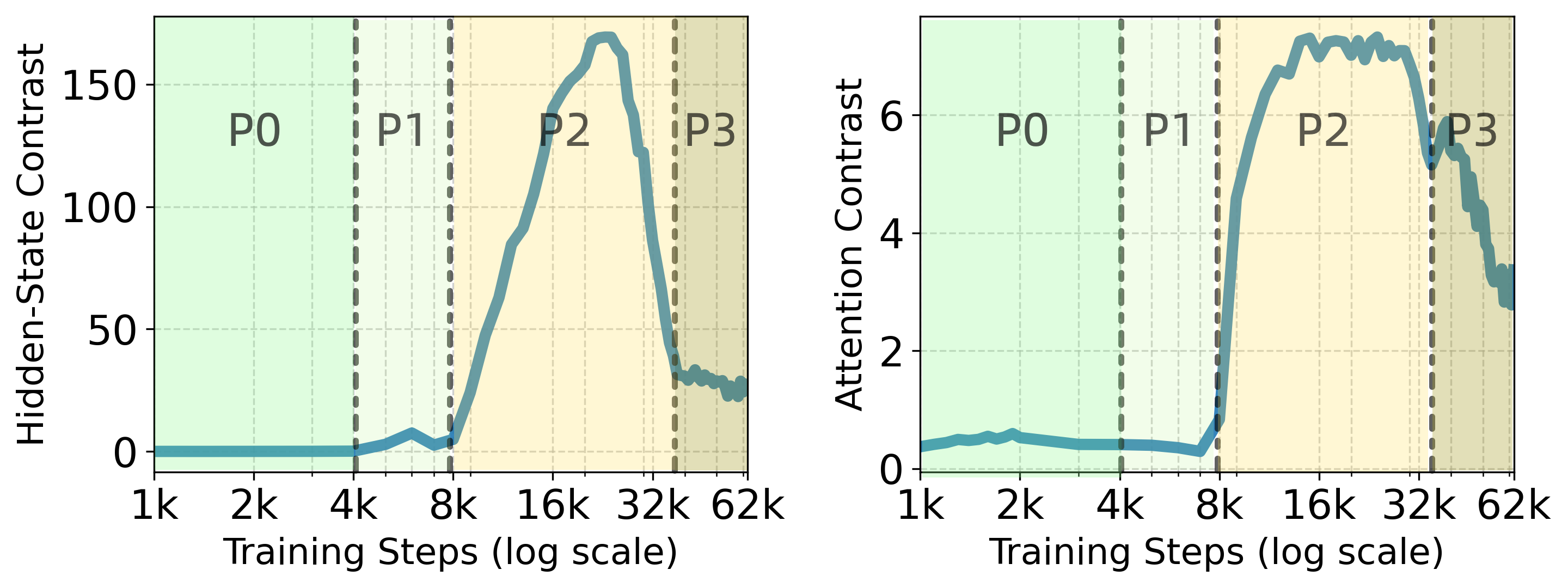}
    \caption{\textbf{Emergence of internal uncertainty via mechanistic analysis:} Evolution of HSC and AC at noise levels $\varepsilon_1=0.01$ and $\varepsilon_2=0.1$. Both metrics experience sharp ascents and subsequent declines, matching phase changes of prediction entropy. 
    %The sharp rise to large values and the subsequent partial drop in both metrics align exactly with Phase 2 as identified by entropy-based phase segmentation.
    }
    \label{fig:self_verification}

\end{figure}

Figure~\ref{fig:self_verification} shows both metrics exhibit sharp changes at the same phase thresholds of reasoning modes. %This suggests that the model internally learns to encode consistency-checks akin to self-verification between $e_1$ and $e_2$, which is externalized as prediction uncertainty. 
This indicating that model internally encodes consistency checks between $e_1$ and $e_2$, resembling a self-verification mechanism that is expressed externally as prediction uncertainty.
%externalizes the uncertainty via prediction entropy
%This suggests that during P2 the model internalizes an indicator of $e_2$–$e_3$ consistency that facilitates self-verification, consistent with our hypothesis.

\begin{imp}
\textit{Reasoning uncertainty is encoded by the model through autoregressive training (including pretraining and supervised finetuning), serving the cornerstone for meta-reasoning abilities such as self-verification.}
\end{imp}

\section{How Shortcut Features Amplify Unfaithfulness}\label{sec:shortcut}

Empirical studies have shown that LLMs are prone to shortcut features that spuriously correlate with solutions \citep{turpin2023language}, which incentivizes models to bypass step-by-step reasoning. For example, adding the cue ``A famous professor thinks the solution is option C'' to the prompt motivates the model to follow the suggested answer regardless of its correctness, together with unfaithful post-hoc rationalization of this answer. 
%We view instruction cues as shortcut features that models learn from training to potentially bypass reasoning.
We view instruction cues as shortcut features that can be used to bypass reasoning.

We investigate the effects of shortcut features by modifying our experiment setting. We replace the modulus with a composite number $N=38$ and sample prompts and reasoning traces in a different format, for example:
%\begin{equation*}
%\underbrace{(a - b) \times c}_{e_1 :\text{prompt}} \ \to \ \underbrace{d \times c}_{e_2: \text{reasoning}} \ \to \ \underbrace{o}_{e_3: \text{solution}}.
%\end{equation*}
\begin{equation*}
(a - b) \times c \ \to \ d \times c \ \to \ o \;,
\end{equation*}
where the first operator in $e_1$ is uniformly sampled from $\{+, -\}$, and the second operator is $\times$. 
%As before, for the noiseless data, all arithmetic expressions are determined under modulus $N$. 
As before, all expressions are determined under modulus $N$. 

\paragraph{Shortcut features.} We view $\mathbf{1}\{c = 0~ \text{or} ~ c=2 ~ \text{or} ~ c = N/2 \}$ as the shortcut feature, which has a direct correlation with the solution. Indeed, when $c=0$ or $c=2$ or $c=N/2$, there is a high probability that $o=0$ without deriving the intermediate reasoning step $e_2$. To evaluate the impact of shortcut features on reasoning faithfulness, we use two test datasets: (i) full test dataset: $n=1,000$ test prompts following the training distribution; (ii) shortcut test subset: a subset of 1000 samples selected from an additional test set, where $c \in \{0, 2, N/2\}$ in the prompt expression $e_1$.

\begin{figure}[t]
    \centering
    \includegraphics[width=0.6\columnwidth]{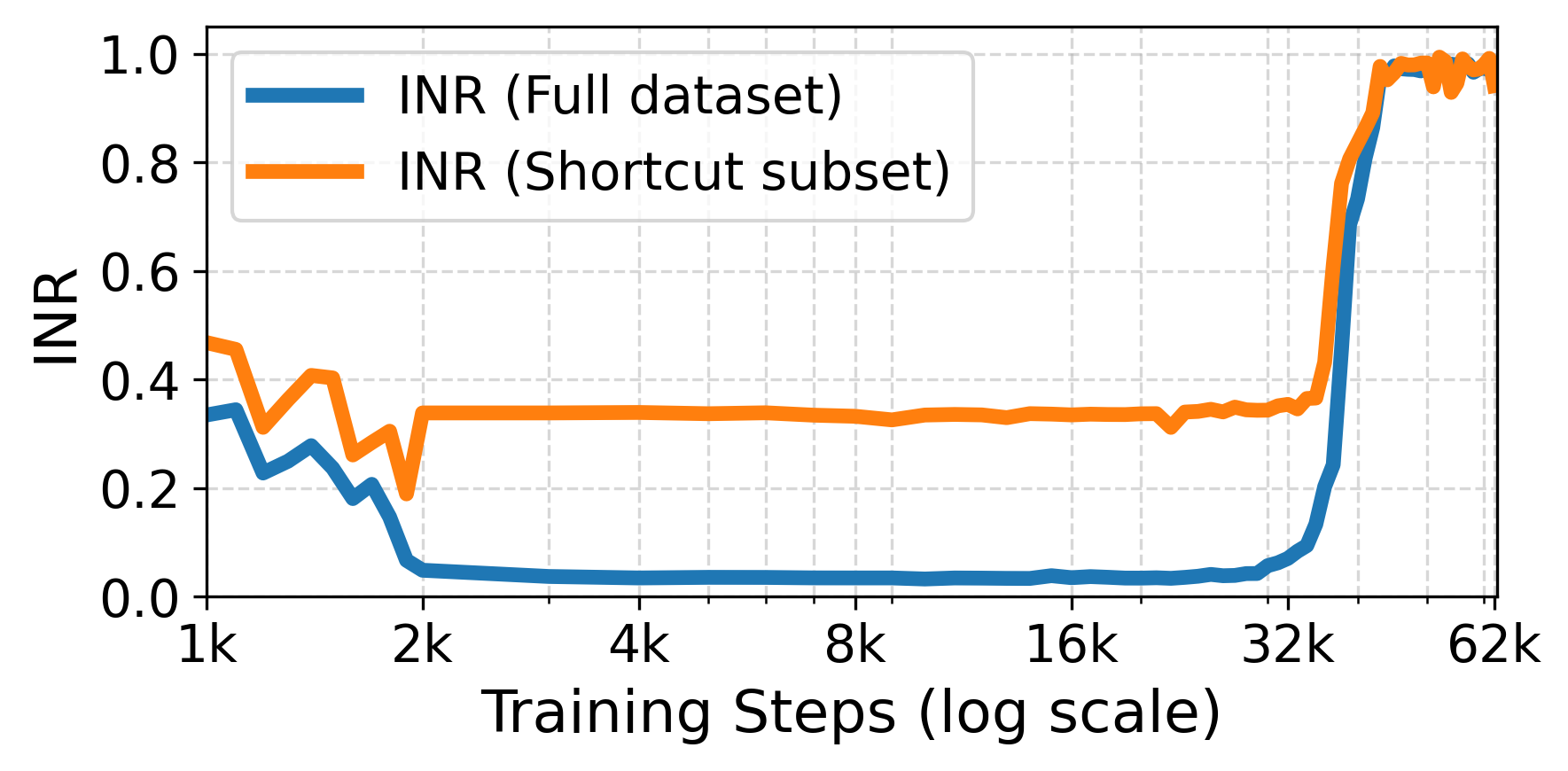}
    \caption{\textbf{Training dynamics of intervention-based metrics on the shortcut and full datasets.} %Both consistency- and intervention-based metrics rise to large values earlier on the shortcut dataset than on the full dataset. 
    INR, as a measure of unfaithfulness, becomes large early, which
    indicates exploitation of shortcut features undermines reasoning faithfulness.
    %to boost accuracy, and the presence of shortcuts increases unfaithfulness.
    }\label{fig:shortcut}
    
\end{figure}

In Figure~\ref{fig:shortcut}, we report intervention-based unfaithfulness metrics over training
under the noise setting $(\veps_1, \veps_2) = (0.01, 0.1)$. Throughout training, INR is
consistently lower on the shortcut test set than on the full test set, indicating a stronger
reliance on skip-step reasoning whenever shortcuts are available. We also compute
consistency-based metrics such as $\mathrm{RIR}_1$ (Section~\ref{sec:append-shortcut}), which corroborate this
behavior from a complementary perspective.

\section{Related Work}

\paragraph{Training dynamics on synthetic tasks.} To understand the learning dynamics of non-reasoning LLMs, well-controlled synthetic tasks have been proposed as proxies for extensive experiments. For example, solving modular arithmetic (e.g., addition, multiplication) requires learning algebraic rules, which exhibits the intriguing grokking phenomenon \citep{power2022grokking}, where models memorize training data long before abruptly generalizing. For copying task, \citet{elhage2021mathematical, olsson2022context, doi:10.1073/pnas.2417182122} identified induction heads as key components for copying arbitrary sequences from context. Through in-context linear regression \citep{garg2022can, von2023transformers}, in-context learning has been shown to behave as a meta-learning algorithm akin to gradient descent. However, none of these works focus on CoT reasoning or test-time composition. In contrast, our AER task aims to fill this gap.

\paragraph{Faithfulness of CoT reasoning.} Prior work offers an empirical account of reasoning faithfulness of pretrained LLMs. It is shown that spurious features and prompt cues shape generated reasoning, suggesting that CoT may function as post-hoc rationalization of answers rather than a causal reasoning process \citep{turpin2023language, arcuschin2025chain, chen2025reasoning, chua2025deepseek, guan2025monitoring}. To address this issue, \citet{turpin2023language, lanham2023measuring, bao2024llms, bao2025likely} proposed intervention-based metrics to assess the causal relationships within reasoning traces. However, without clean experiment setups targeted at CoT reasoning,  existing works lack a principled framework for faithfulness. Moreover, they focus on static models instead of the emergence of unfaithfulness, yet the latter is critical to understanding the causes of unintended reasoning behaviors, with implications for misalignment and AI safety. In contrast, our work studies the learning dynamics using a clean experimental setup.

\paragraph{Self-reflection and uncertainty metrics.} LLMs exhibit striking meta-reasoning skills since GPT-3 \citep{brown2020language, weng2023large}. The abilities of self-reflection and self-verification are further boosted through iterative refinement \citep{madaan2023self} and post-training \citep{guo2025deepseek}, which significantly reduce inconsistency and unfaithfulness. Recent work has demonstrated the promise of leveraging such metrics to improve LLM reasoning~\citep{fu2025deep, zhao2025learning}.  However, existing empirical advances offer little insight into the underlying mechanism: how these meta-reasoning skills are developed through training, and how the models internally represent such skills. Our work addresses this gap by offering a mechanistic account: models learn to encode consistency checks and internal uncertainty as they resolve inconsistencies in reasoning traces over the course of training.

\section{Limitations and Future Work}

Our work focuses on autoregressive training which is the standard training paradigm for pretraining and supervised finetuning (SFT). In future work, it would be interesting to investigate reinforcement learning with verifier rewards (RLVR) \citep{guo2025deepseek} which becomes a standard recipe for enhancing LLM's reasoning capabilities. We would also be interested in applying our  metrics to LLMs beyond our synthetic setting.

%Our work analyzes training dynamics with autoregressive training, which is the standard training paradigm for pretraining and supervised finetuning (SFT). However, recent two years have seen increasing interests in post-training strategies such as reinforcement learning with verifier rewards (RLVR) \citep{guo2025deepseek}, which significantly enhances LLM's reasoning capabilities. An interesting question is to understand how RLVR externalizes implicit self-verification---a skill that we identify in this work---into explicit meta-reasoning behavior, and why RLVR is better at out-of-distribution compared with SFT \citep{chu2025sft}.

%In addition, while our work focus on the fundamental mechanism of CoT reasoning and the emergence of implicit self-verification, we have not proposed actionable methods to improve faithfulness evaluations. It would be interesting to explore in future work whether our insights lead to new intervention metrics (besides random perturbation), and whether hidden-states-based uncertainty metrics can better quantify a model's internal uncertainty. 

\section*{Impact Statement}

Our work analyzes the fundamental mechanism of CoT reasoning using controlled synthetic experiments. We do not train new models or study practical enhancements that may lead to harmful applications. The main positive impact is to improve scientific understanding of CoT reasoning, particularly the mechanism of unfaithfulness in autoregressive training. Through clear definitions, evaluation metrics, and principled analysis, our results may offer insights to improve the reliability of reasoning capabilities of LLMs. In particular, our findings about models' internal uncertainty and implicit self-verification offer insights into meta-reasoning skills and may lead to safer finetuning techniques. A potential negative impact is over-generalization of our findings beyond the synthetic task. Our conclusions would require further extensive evaluations across LLMs and tasks before being used in policy making or deployment decisions.

\section*{Acknowledgments}

Y.Z.~is partially supported by NSF-DMS grant 2412052 and by a Coefficient Giving (formerly Open Philanthropy) grant. We also thank Bin Yu for helpful discussions.

\bibliography{refs}
\bibliographystyle{plainnat}

%%%%%%%%%%%%%%%%%%%%%%%%%%%%%%%%%%%%%%%%%%%%%%%%%%%%%%%%%%%%%%%%%%%%%%%%%%%%%%%
% APPENDIX
%%%%%%%%%%%%%%%%%%%%%%%%%%%%%%%%%%%%%%%%%%%%%%%%%%%%%%%%%%%%%%%%%%%%%%%%%%%%%%%
\newpage
\appendix
\onecolumn

%\section{Experiment Details}\label{sec:append-exp}

% ==== begin inlined latex_files/append_exp.tex ====
\section{Experimental details}\label{sec:append-exp}

\paragraph{Models.} Our model follows the standard transformer architecture \cite{vaswani2017attention} and the standard practice where layer normalization is placed before self-attention and feedforward neural network (FFN) \cite{Karpathy2022}. We also used RoPE \cite{su2024roformer} which is the default positional embedding for many LLMs. During training, we did not use any scheduler to adjust the learning rate.
\begin{itemize}
\item Number of layers ($L$): 3
\item Number of heads ($H$): 2
\item Model dimension: 128 
\item FFN dimension: 512 
\item Dropout: 0.01 
\item Theta parameter is RoPE: 10000 
\item Weight decay: 0.01 
\end{itemize}

\paragraph{Training.} Each training example follows the format in \eqref{eq:format}, consisting of 11 tokens plus an EOS token. We compute the autoregressive loss starting from the first $\rightarrow$ token (the 6th token) through the final solution token. The two unfaithfulness metrics, along with auxiliary quantities such as prediction entropy, are computed online during training to avoid additional overhead. Training a single model under a given noise configuration—including all metric evaluations—takes approximately 50 minutes on a single NVIDIA A30 GPU.

% ==== end inlined latex_files/append_exp.tex ====
%\section{Additional Results}\label{sec:append-additional}

% ==== begin inlined latex_files/append_additional.tex ====
\section{Additional experiment results}\label{sec:append-additional}

\subsection{Validating simplicity bias: swapping prompt and reasoning}\label{sec:append-swap}
We investigate the effect of positional order on reasoning faithfulness, focusing on the noiseless setting for clarity. Specifically, we compare two configurations:
\begin{itemize}
\item No swap: $e_1 \to e_2 \to e_3$
\item Swap: $e_2 \to e_1 \to e_3$
\end{itemize}
where $(e_1, e_2, e_3)$ denotes a consistent reasoning chain. In addition to the original (no-swap) setting, we train a transformer on the swapped data using identical hyperparameters. After training, we compare the four faithfulness metrics, summarized in Table~\ref{tab:complexity_gap}.
For completeness, we restate the unfaithfulness metrics defined in the main text:
\begin{align}
\nonumber
&\mathrm{RIR}_1 := \frac{\sum_{i \le n} \mathbf{1}\{e_{i,2}' \neq e_{i,2},\, e_{i,3} = e_{i,3}'\}}{\sum_{i \le n}\mathbf{1}\{e_{i,2}' \neq e_{i,2}\}}   \\
\nonumber
&\mathrm{RIR}_2 := \frac{\sum_{i \le n} \mathbf{1}\{e_{i,2}' \neq e_{i,2},\, e_{i,3} = e_{i,3}'\}}{\sum_{i \le n}\mathbf{1}\{e_{i,3}' = e_{i,3}\}} \\
\nonumber
&\mathrm{IDS}   := \frac{1}{n} \sum_{i=1}^{n}  \KL \big(\widehat f_2(e_{i,1}, e_{i,2}'),\, \widehat f_2(e_{i,1}, \tilde e_{i,2}) \big) \\
\nonumber
&\mathrm{INR}   := \frac{1}{n} \sum_{i=1}^{n} \mathbf{1}\big\{\argmax_{e \in \gV_N} p(e \mid e_{i,1}, e_{i,2}')
  = \argmax_{e \in \gV_N} p(e \mid e_{i,1}, \tilde e_{i,2}) \big\}
\nonumber
\end{align}

For the swapped CoT format, we correspondingly redefine INR and IDS. 
Conceptually, both metrics are still defined via perturbations applied to \(e_2\); 
however, after swapping, \(e_2\) now appears in the first step of the reasoning chain. 
Similarly, in the definitions of \(\mathrm{RIR}_1\) and \(\mathrm{RIR}_2\), 
the functional roles of \(e_2\) and \(e_3\) remain unchanged, 
except that \(e_2\) is now positioned at the beginning of the chain.

 \begin{table}[H]
    \caption{Unfaithfulness metrics before and after swapping $e_1$ and $e_2$ on noise-free data.}
    \label{tab:complexity_gap}
    \begin{center}
    \begin{small}
    \begin{sc}
    \begin{tabular}{lcccc}
        \toprule
        Data layout & $\text{RIR}_1$ & $\text{RIR}_2$ & IDS & INR \\
        \midrule
        $e^{0}_1 \to e^{0}_2 \to e_3$ & 0 & 0 & 24.7 & 0 \\
        $e^{0}_2 \to e^{0}_1 \to e_3$ & 1.00 & 0.99 & 26.0 & 0 \\
        \bottomrule
    \end{tabular}
    \end{sc}
    \end{small}
    \end{center}
\end{table}

The intervention-based metrics (IDS and INR) remain nearly unchanged, 
indicating that the observed effect is not driven by positional order: 
the model inherently prefers to rely on the simpler \(e_2\) for prediction. 
In contrast, both RIR scores drop from \(1\) in the original format to \(0\) after swapping, 
since relying on \(e_2\) for prediction becomes a form of ``skip-step'' reasoning under the swapped ordering.

%\subsection{KL to uniform distribution}
%\subsection{Explanation for mixed reasoning mode}
%\label{sec:KL_uniform}

\subsection{Reasoning phases}\label{sec:append-phase}

\paragraph{Further explanation for mixed reasoning mode.} We hypothesize that, during training, the model enters a \emph{mixed reasoning phase} in which its
prediction behavior can be approximated as follows. Define the indicator
\[
c(e_1, e_2) = \mathbf{1}\{ f(e_1) = f_2(e_2) \},
\]
and conjecture that the model effectively predicts according to
\[
\widehat f_2(e_1, e_2)
\;\approx\;
f_2(e_2)\, c(e_1,e_2)
\;+\;
U \, (1 - c(e_1,e_2)),
\]
where \(U\) denotes the uniform distribution on \(\gV_N\).
In other words, the model develops a form of self-verification: it compares the results computed
from \(e_1\) and \(e_2\), and whenever these results disagree, it outputs an approximately uniform
predictive distribution.

To quantify this phenomenon, we consider the metric
\[
\frac{1}{|\mathcal{I}|}
\sum_{i \in \mathcal{I}}
D_{\mathrm{KL}}\!\bigl(
\widehat f_2(e_{i,1},\, e'_{i,2}),\,
U
\bigr),
\]
where \(U\) is the uniform distribution on \(\gV_N\), and
\[
\mathcal{I}
=
\{\, i : f(e_{i,1}) \neq f_2(e_{i,2}) \,\}
\]
is the set of inconsistent chains. A smaller value of this metric indicates that, on inconsistent
examples, the model's predictive distribution is closer to uniform, providing stronger evidence that it
has transitioned into the mixed reasoning (self-verification) phase.

\paragraph{Qualitative characteristics of phase diagrams.} 
For noise levels \(\varepsilon_1 = 0.01\) and \(\varepsilon_1 = 0.1\), we construct phase diagrams with
training compute on the x-axis and reasoning noise \(\varepsilon_2\) on the y-axis. For each pair
\((\text{compute}, \varepsilon_2)\), we assign a phase according to which of three reference
distributions—the stepwise solution, the skip-step solution, or a uniform distribution (evaluated
only on inconsistent chains)—has the smallest KL divergence to the model’s predictive
distribution.

\begin{figure}[H]
    \centering
    \begin{subfigure}{0.32\textwidth}
        \centering
        \includegraphics[width=\linewidth]{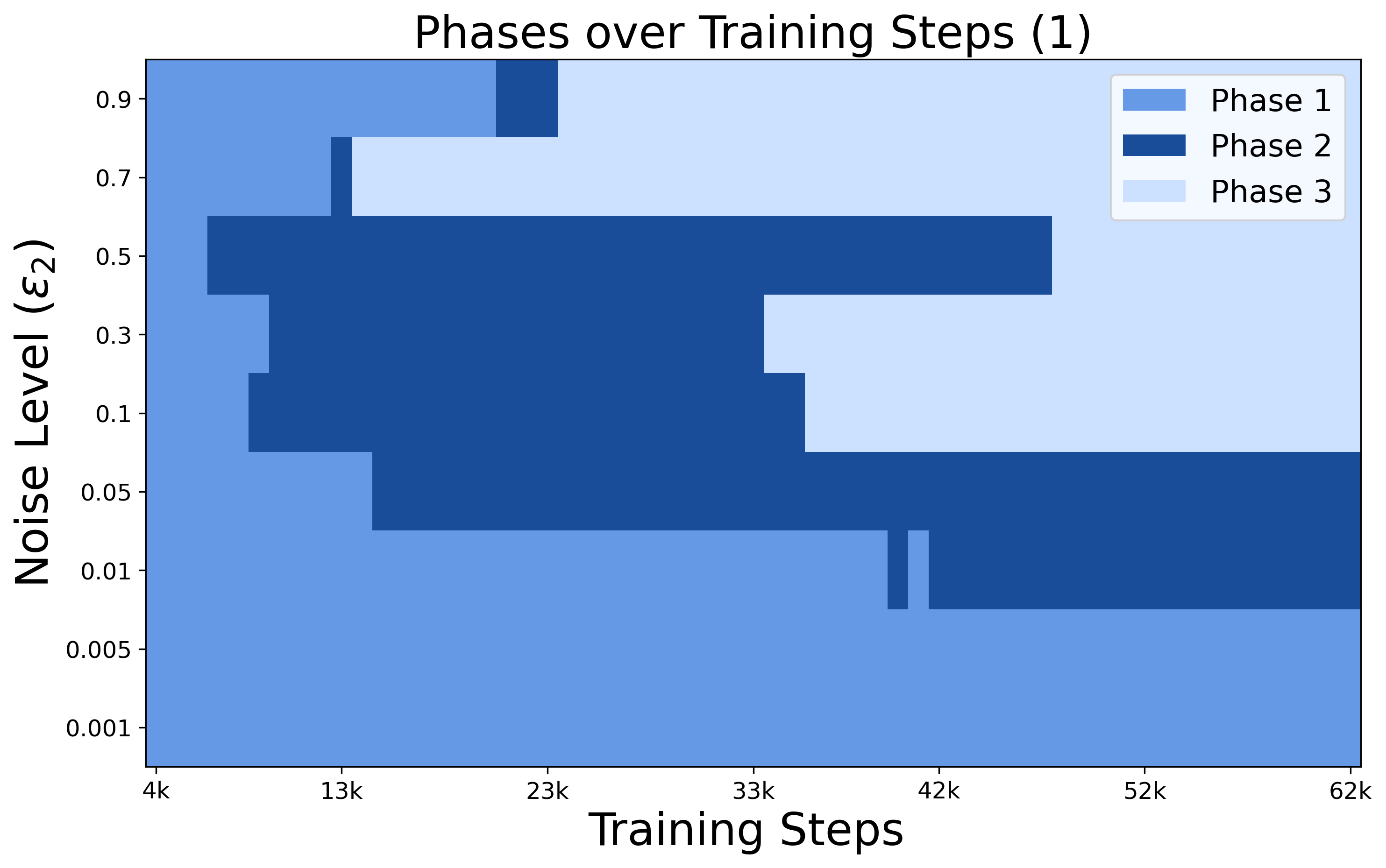}
    \end{subfigure}
    \begin{subfigure}{0.32\textwidth}
        \centering
        \includegraphics[width=\linewidth]{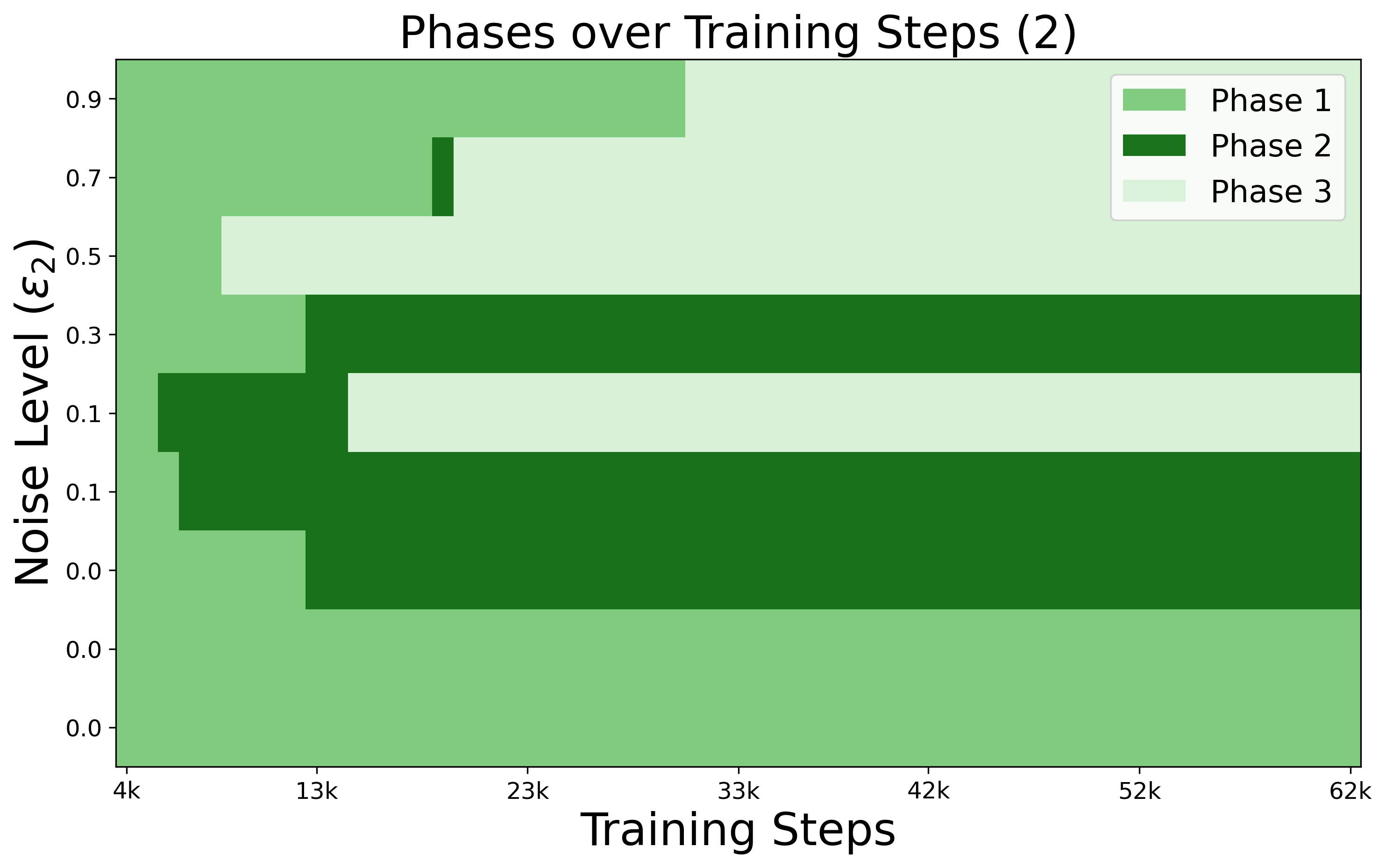}

    \end{subfigure}
    \begin{subfigure}{0.32\textwidth}
        \centering
        \includegraphics[width=\linewidth]{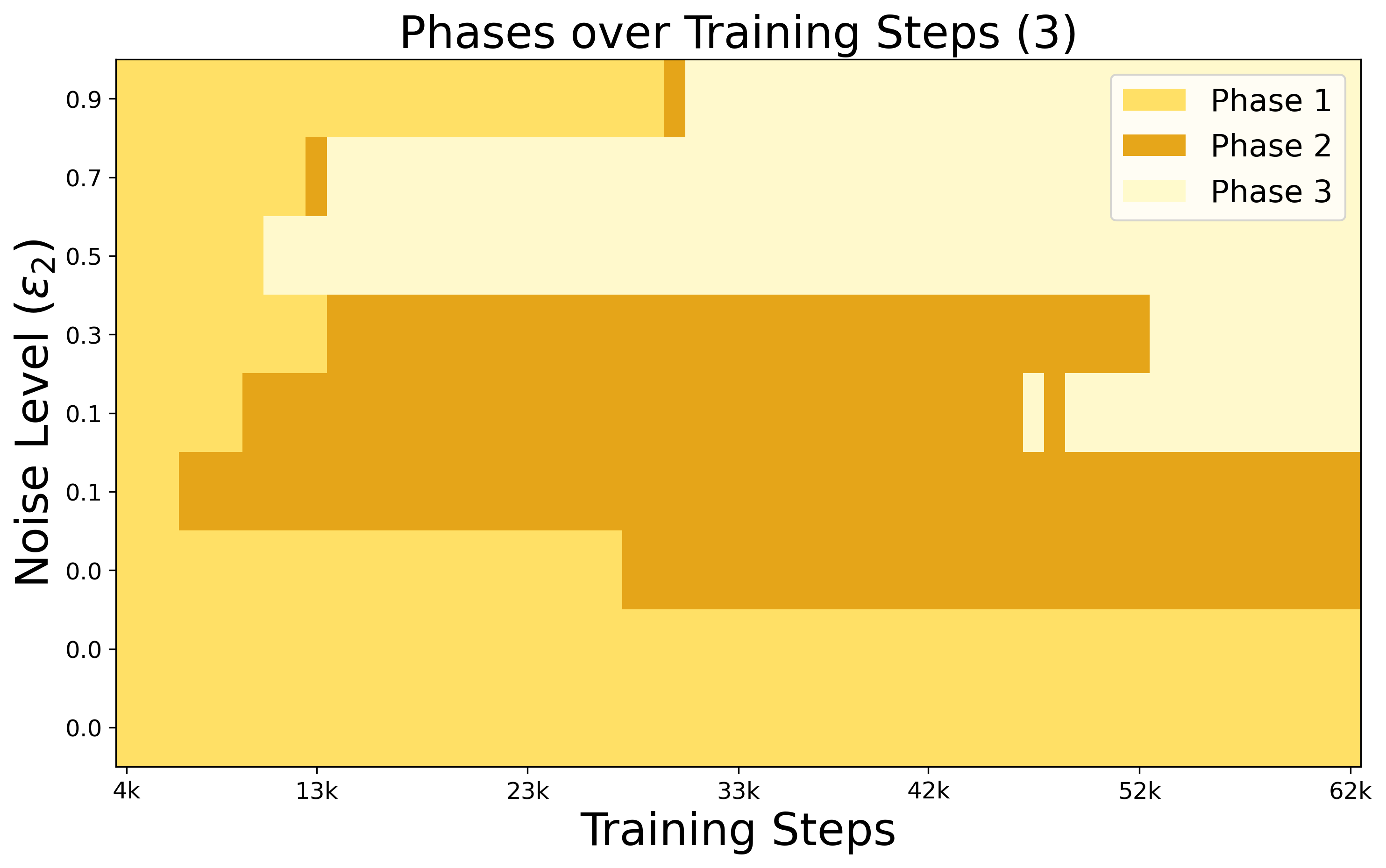}
    
    \end{subfigure}
    \caption{\textbf{Data quality and compute as deciding factors of reasoning modes.} We train models under fixed $\veps_1=0.1$ and varying $\veps_2$. More training steps (compute) and higher noise (data quality) tend to induce skip-step reasoning. The three panels show phase diagrams under three random seeds from the same data distribution, illustrating that the training dynamics are sensitive to randomness in the data and optimization.}
    \label{fig:phase_diagram}
\end{figure}

As shown in Figure~\ref{fig:phase_diagram}, we observe a clear qualitative pattern in phase
development: as \(\varepsilon_2\) increases, the model transitions from exhibiting two phases, to
three, and eventually to four distinct phases. For larger values of \(\varepsilon_2\), increasing
compute further drives a gradual shift from stepwise reasoning toward skip-step reasoning.
Intuitively, a larger \(\varepsilon_2\) (relative to \(\varepsilon_1\)) induces a greater gap in
achievable prediction accuracy between different reasoning modes, and the model’s drive to
maximize accuracy consequently pushes it toward adopting the skip-step reasoning strategy.

We also note that the model’s phase transitions and training dynamics are sensitive to both the training data and the optimization process. Figure~\ref{fig:phase_diagram} shows phase diagrams obtained by training the model on three different training sets generated from the same underlying distribution but with different random seeds. Despite sharing the same data distribution, the resulting trajectories differ substantially across seeds, including the timing of each phase transition and even the final phase to which the model achieve.

\begin{figure}[H]
\centering
\includegraphics[width=0.2\columnwidth]{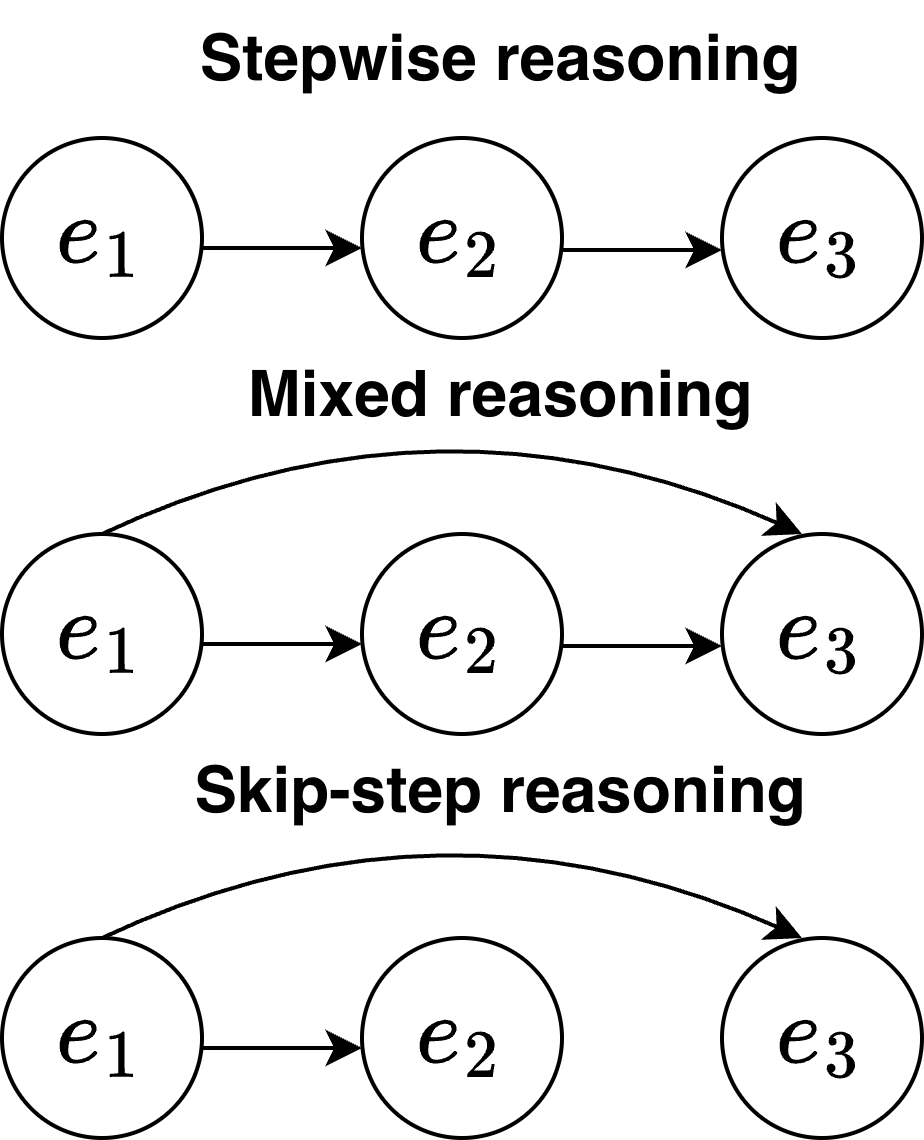}
    \caption{\textbf{Causal graphs representing three reasoning modes.} See \citet{bao2024llms} for a similar causal perspective.}
    \label{fig:causal}
\end{figure}

\paragraph{Causal interpretations of three reasoning modes.} Figure~\ref{fig:causal} illustrates three qualitatively distinct reasoning patterns. The first row depicts fully
faithful stepwise reasoning, which is the ideal behavior we aim to elicit from the model during
training. The second row shows a mixed reasoning mode: the model relies on \(e_1\) when
generating \(e_2\), and then uses both \(e_1\) and \(e_2\) when generating \(e_3\). This constitutes
an unfaithful reasoning process, in which the model effectively couples \(e_2\) and \(e_3\) during
inference. The third row presents another form of unfaithful reasoning, \emph{skip-step
deduction}, where the generation of \(e_3\) depends solely on \(e_1\) and is entirely independent
of \(e_2\).

\subsection{Additional results}\label{sec:metrics_results}
\paragraph{Operator norm in Attention Contrast (AC).}\label{par:operator_norm}
We briefly recall the notion of the matrix operator norm used in the definition of Attention Contrast (AC).
Let \(\|\cdot\|\) be a norm on \(\mathbb{R}^d\). For a matrix \(A \in \mathbb{R}^{d \times d}\), the (induced) operator norm of \(A\) with respect to \(\|\cdot\|\) is defined as
\[
\|A\|_{\mathrm{op}}
\;:=\;
\sup_{x \neq 0} \frac{\|Ax\|}{\|x\|}
\;=\;
\sup_{\|x\| = 1} \|Ax\|.
\]
Intuitively, \(\|A\|_{\mathrm{op}}\) measures the largest factor by which \(A\) can stretch a vector under the chosen norm.
In the Euclidean case \(\|\cdot\| = \|\cdot\|_2\), this coincides with the largest singular value of \(A\), i.e., \(\|A\|_{\mathrm{op}} = \sigma_{\max}(A)\).

In our AC, the matrix inside the operator norm is
\[
h_{L-1,i}^{+}W^{(L)}_{QK}(h_{L-1,i}^{+})^{\top}
\;-\;
h_{L-1,i}^{-}W^{(L)}_{QK}(h_{L-1,i}^{-})^{\top},
\]
which represents the difference between the last-layer attention score matrices for a positive (consistent) chain \((e_1^{+}, e_2^{+})\) and a negative (inconsistent) chain \((e_1^{-}, e_2^{-})\) at sample index \(i\). The operator norm
\(\bigl\|h_{L-1,i}^{+}W^{(L)}_{QK}(h_{L-1,i}^{+})^{\top}
- h_{L-1,i}^{-}W^{(L)}_{QK}(h_{L-1,i}^{-})^{\top}\bigr\|_{\mathrm{op}}\)
thus captures the maximal change in attention scores along any direction in the representation space when we move from consistent to inconsistent chains. Averaging this quantity over \(i\) in the definition of AC yields a global measure of how strongly the model’s attention mechanism separates positive and negative samples, providing evidence for internal uncertainty encoding or self-verification behavior.

\begin{figure}[H]
\centering
\begin{subfigure}{0.40\textwidth}
        \centering
        \includegraphics[width=\linewidth]{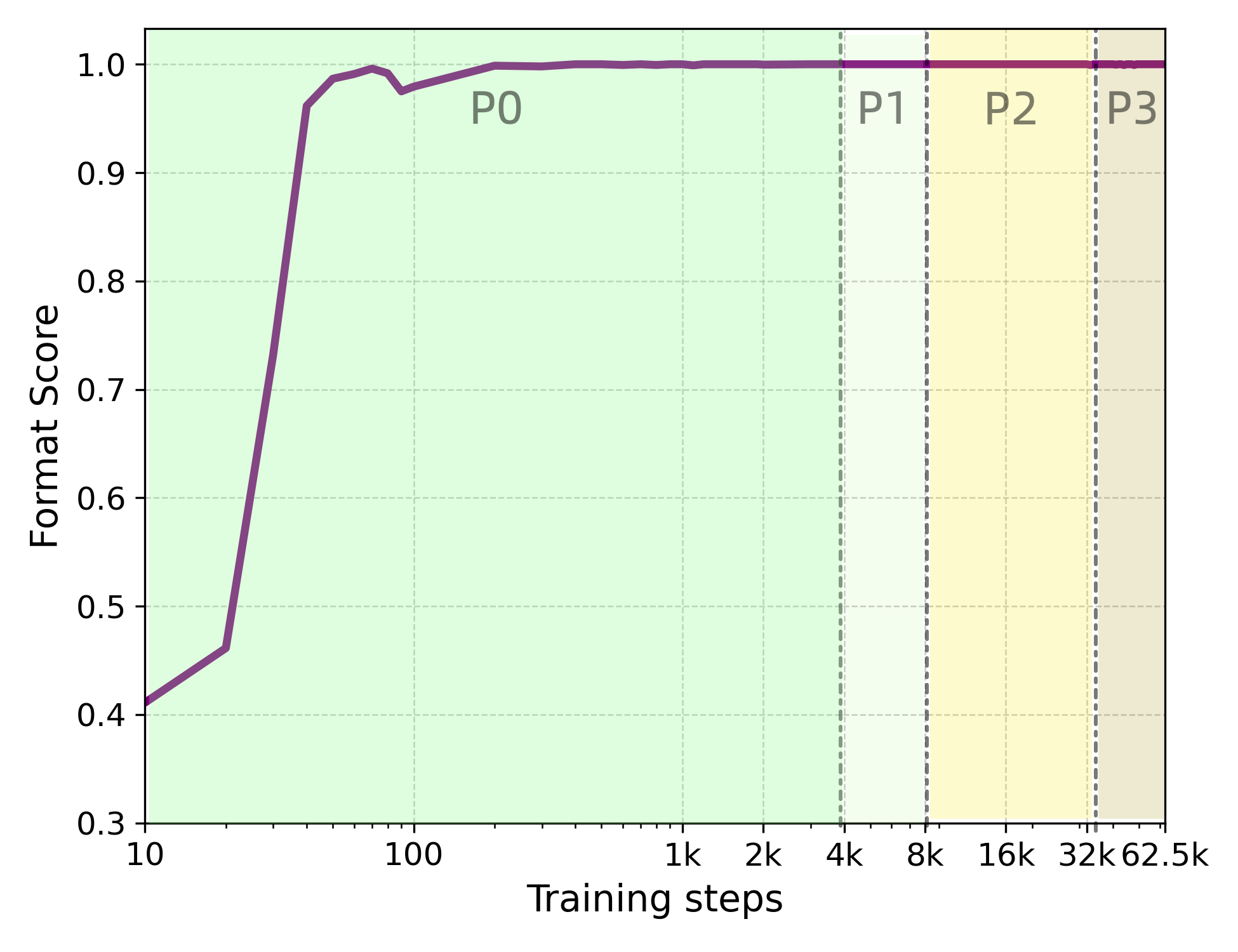}
    \end{subfigure}
\begin{subfigure}{0.40\textwidth}
        \centering
        \includegraphics[width=\linewidth]{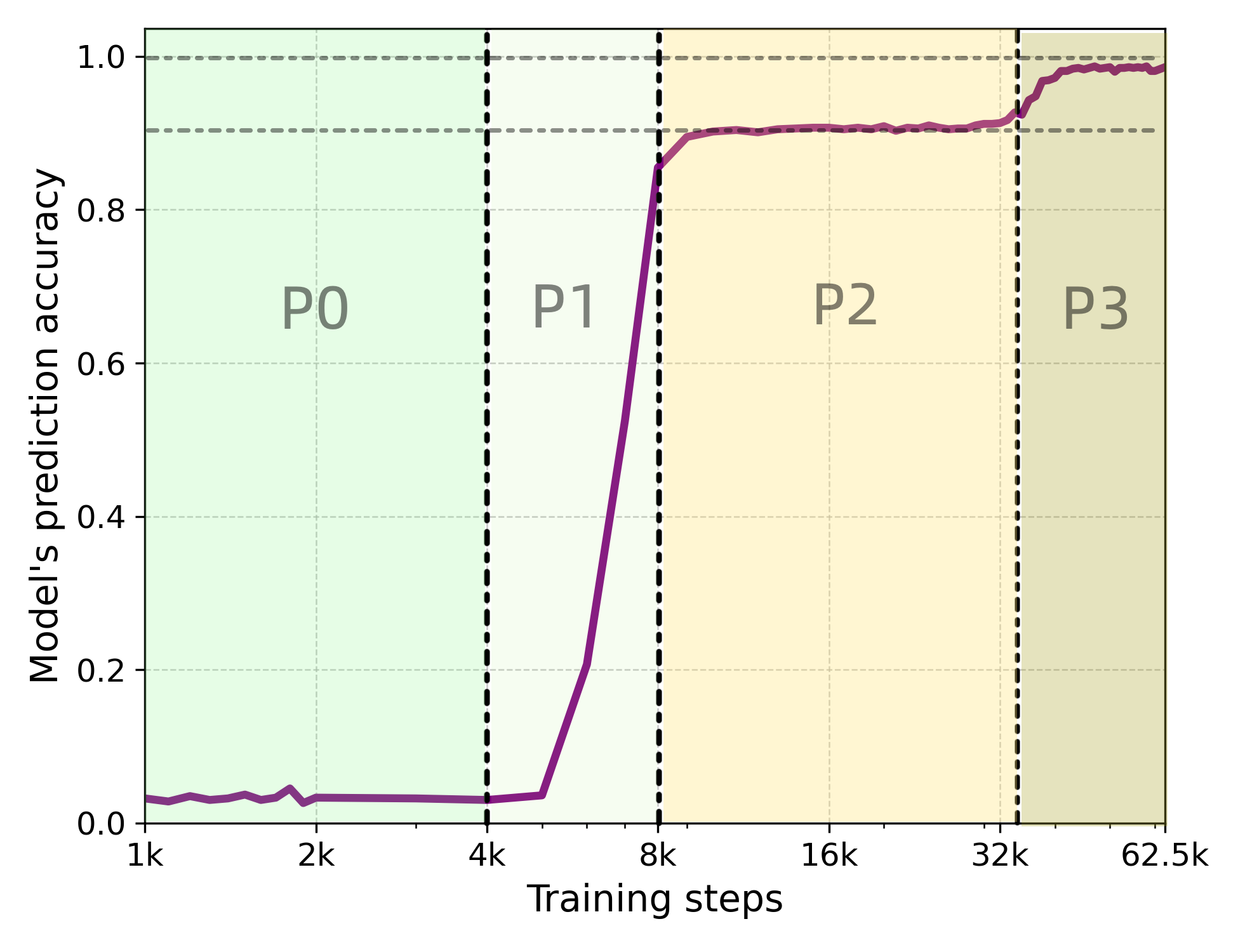}
    \end{subfigure}
    \caption{\textbf{Training trajectories of the model’s format score and predictive accuracy.} \textbf{Left:} \textbf{The format score over training steps.} During Phase 0, the model rapidly improves its accuracy in predicting the structural positions (operators and deduction tokens). \textbf{Right:} \textbf{The model's prediction accuracy (on $e_3$) over training steps.} The model first learns to rely on \(e_2\) and reaches an accuracy of about \(90\%\) in P1. In P2, the model develops a self-verification mechanism, which does not further improve prediction accuracy. Eventually, the model transitions to skip-step reasoning, and the accuracy increases to approximately \(99\%\). }
    \label{fig:Format_score}
\end{figure}

\paragraph{Phase 0 and format score.} In Section ~\ref{sec:phases}, we hypothesize that there is an initial Phase~0 during training in which the model primarily learns the data format. Figure~\ref{fig:Format_score} shows how the format score evolves during training on datasets with noise levels \(\varepsilon_1 = 0.01\) and \(\varepsilon_2 = 0.1\). We observe that the format score rapidly increases from 0 to 1 within the first \(1\mathrm{k}\) steps, indicating that, in this early stage, model indeed first focuses on matching the data format. In Figure~\ref{fig:Format_score}, we also plot how the model’s prediction accuracy evolves over training. In Phase~1, due to simplicity bias, the model first learns to perform stepwise prediction of \(e_3\) from \(e_2\), achieving an accuracy of approximately \(1 - p_2 = 90\%\). In P3, the model switches to skip-step reasoning, and the accuracy increases to about \(1 - p_1 = 99\%\). In contrast, P2 is dominated by self-verification: according to our hypothesis, for samples with \(f(e_1) = f_2(e_2)\) the model outputs an approximately uniform distribution and thus cannot predict the correct result. Consistent with this picture, the accuracy curve in Figure~9 does not increase during P2 but instead remains flat, providing indirect evidence for our proposed mechanism in Phase~2.

\begin{figure}[H]
    \centering
    \begin{subfigure}{0.32\textwidth}
        \centering
        \includegraphics[width=\linewidth]{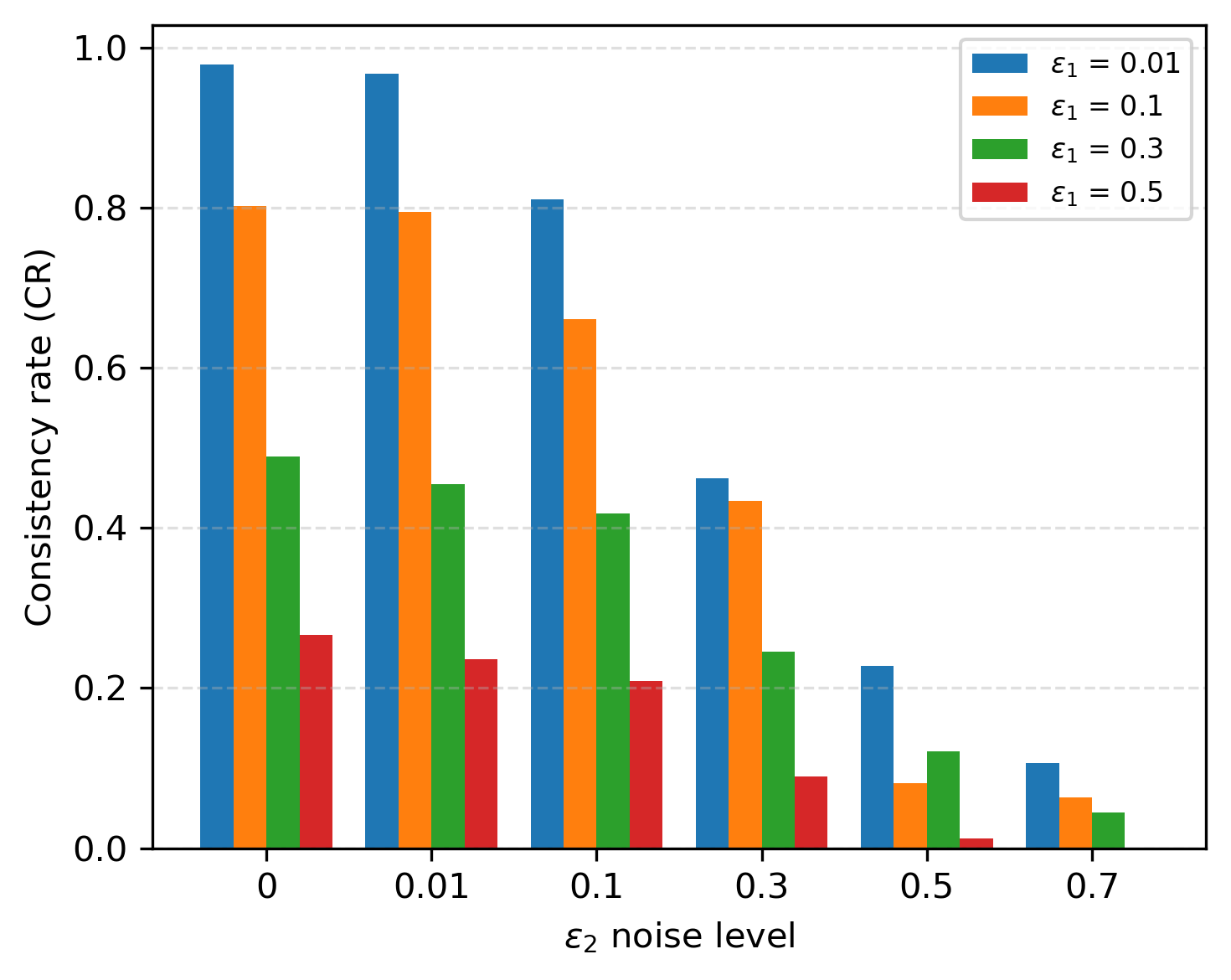}
    \end{subfigure}
    \begin{subfigure}{0.32\textwidth}
        \centering
        \includegraphics[width=\linewidth]{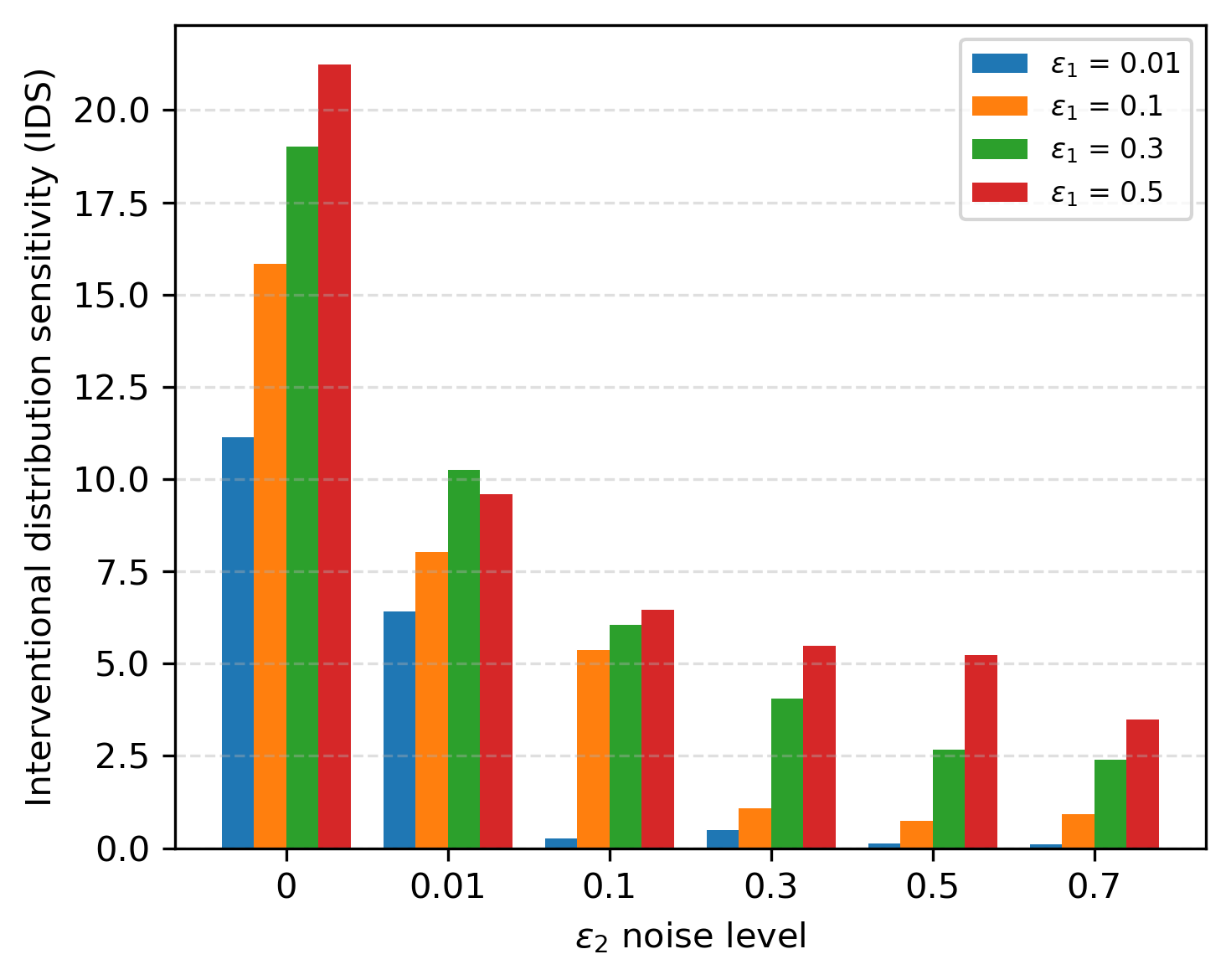}

    \end{subfigure}
    \begin{subfigure}{0.32\textwidth}
        \centering
        \includegraphics[width=\linewidth]{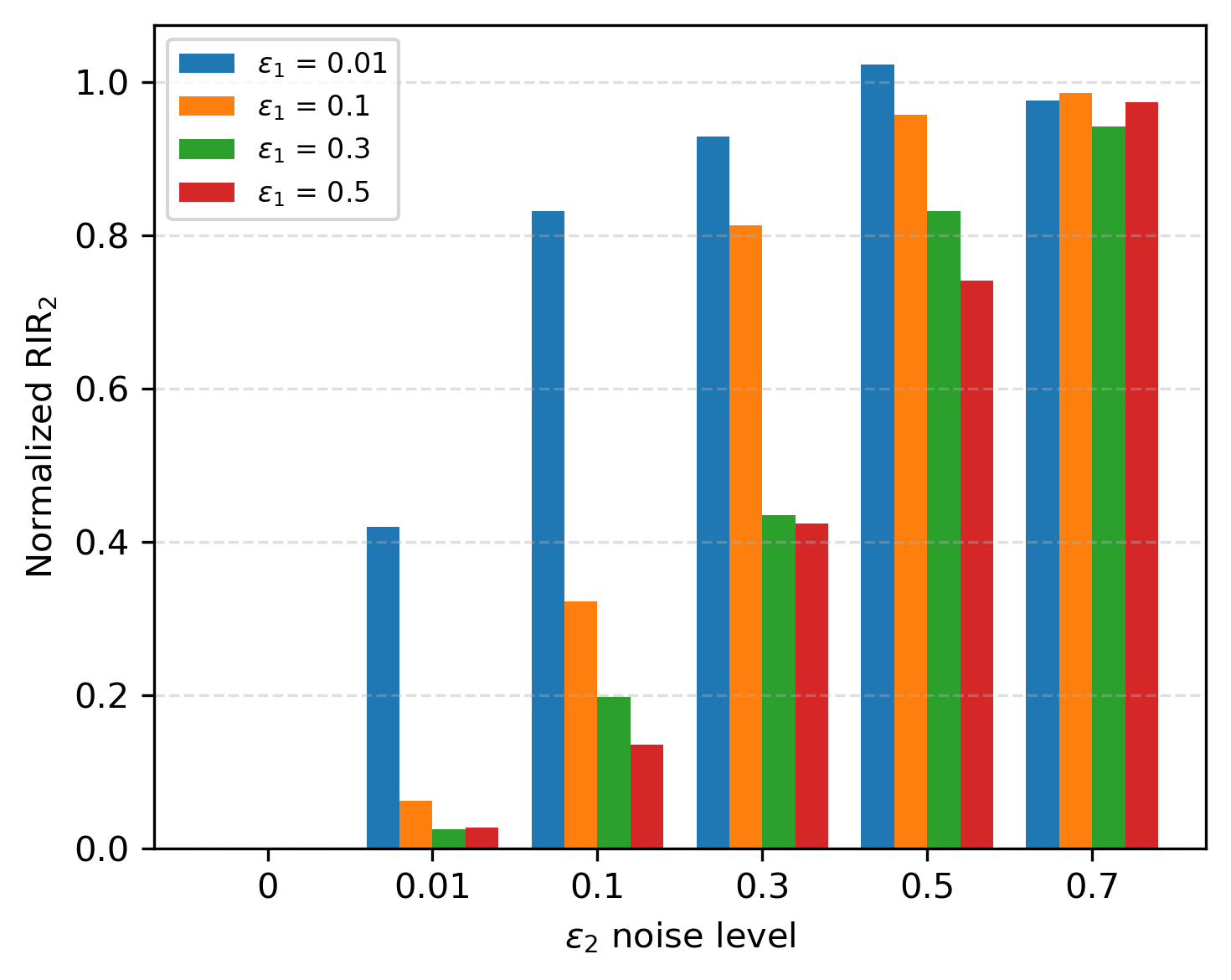}
    
    \end{subfigure}
    \caption{\textbf{Results of additional unfaithfulness metrics for models trained across different noise regimes.} \textbf{Left:} As $\varepsilon_2$ increases, the consistency ratio gradually decreases, indicating that the model becomes increasingly less capable of performing successful CoT reasoning. \textbf{Middle/Right:} IDS decreases with increasing $\varepsilon_2$, whereas $\mathrm{RIR}_2$  increases. Together, these trends corroborate that the model’s degree of unfaithfulness grows as the noise level becomes larger.}
    \label{fig:additional_metrics}
\end{figure}

\paragraph{Additional unfaithfulness metrics.} We present additional metrics for quantifying the model’s unfaithfulness. Figure~\ref{fig:additional_metrics} shows, over a grid of noise levels 
\(\varepsilon_1 \in \{0.01, 0.1, 0.3, 0.5\}\) and 
\(\varepsilon_2 \in \{0, 0.01, 0.1, 0.3, 0.5, 0.7\}\), 
the model’s consistency ratio (CR), interventional distribution sensitivity (IDS), and reasoning inconsistency ratio (RIR) after training. 
Since we compare \(\mathrm{RIR}_2\) across different values of \(\varepsilon_2\), we report a version of \(\mathrm{RIR}_2\) that is normalized by the effective noise level in \(\varepsilon_2\): $\mathrm{Normalized\ RIR_2}=\mathrm{RIR_2}/(n^{-1}\sum_{i \le n}\mathbf{1}\{e_{i,2}' \neq e_{i,2}\})$. 
Figure~\ref{fig:additional_metrics} further shows that, for any fixed \(\varepsilon_1\), the model’s unfaithfulness increases as \(\varepsilon_2\) grows. 
Moreover, as \(\varepsilon_1\) becomes larger, the fraction of successful reasoning chains decreases, and the model increasingly favors stepwise reasoning over skip-step reasoning, as the latter no longer provides a meaningful improvement in prediction accuracy in this regime.

\begin{figure}[H]
    \centering
    \hspace*{-0.5cm}
\includegraphics[width=0.9\textwidth]{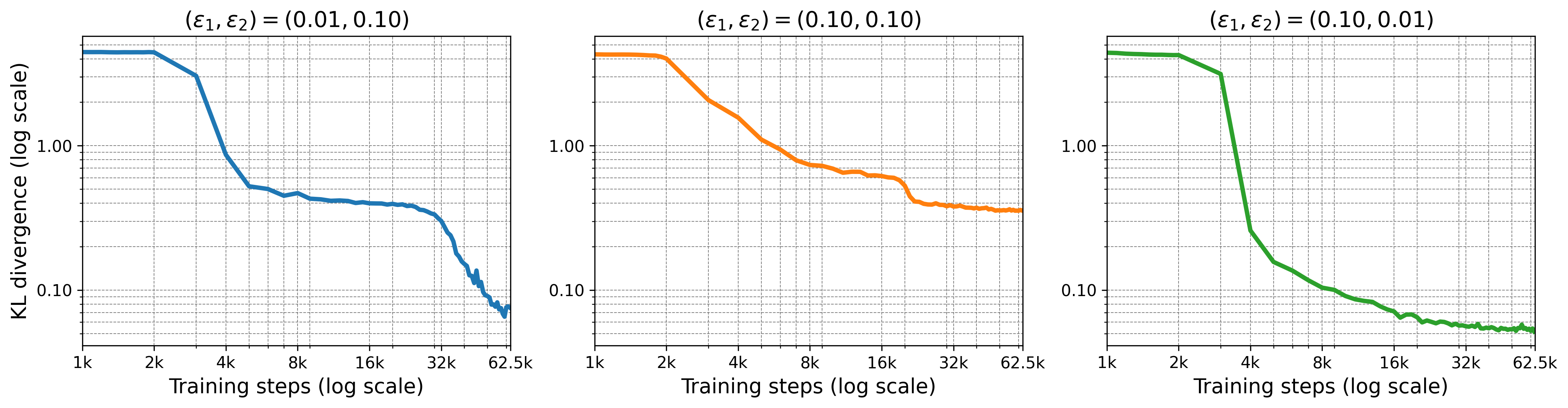}
    \caption{\textbf{KL divergence between the model’s learned distribution and the optimal distribution \(p(e_3 \mid e_1, e_2)\) during training for three noise configurations.} For $(\varepsilon_1,\varepsilon_2)=(0.01,0.1)$, the model converges to skip-step reasoning, and the KL divergence to the theoretical optimum goes to $0$. Similarly, for $(\varepsilon_1,\varepsilon_2)=(0.1,0.01)$, it converges to stepwise reasoning with vanishing KL. In contrast, for $(\varepsilon_1,\varepsilon_2)=(0.1,0.1)$, the model gets stuck in the self-verification regime and does not reach the optimal predictive distribution.}
    
    \label{fig:KL_mixed_distribution}
\end{figure}

\paragraph{Further analysis of the optimization trajectory.} During training, we also track the KL divergence between the model’s learned distribution and the corresponding theoretically optimal distribution: $p(e_3 \mid e_1, e_2)$, in order to better understand the optimization trajectory. Figure~\ref{fig:KL_mixed_distribution} shows the evolution of this KL divergence for the three noise configurations considered in the main text. For noise levels \((\varepsilon_1, \varepsilon_2) = (0.01, 0.1)\) and \((0.1, 0.01)\), the KL divergence eventually approaches zero, reflecting that the model converges to the respective optimal target distributions under these noise settings. In contrast, for \((\varepsilon_1, \varepsilon_2) = (0.1, 0.1)\), Figure~12 shows that within 62{,}500 training steps the model remains in Phase~2 (self-verification). From an optimization perspective, we hypothesize that this phase corresponds to a local minimum or saddle point in the loss landscape. When \(\varepsilon_1\) and \(\varepsilon_2\) are close, such critical structures appear more frequently, making it harder for the model to escape and thus requiring substantially more training steps.

\subsection{Shortcut features}\label{sec:append-shortcut}
\begin{figure}[H]
    \centering
    \begin{subfigure}{0.4\textwidth}
        \centering
        \includegraphics[width=\linewidth]{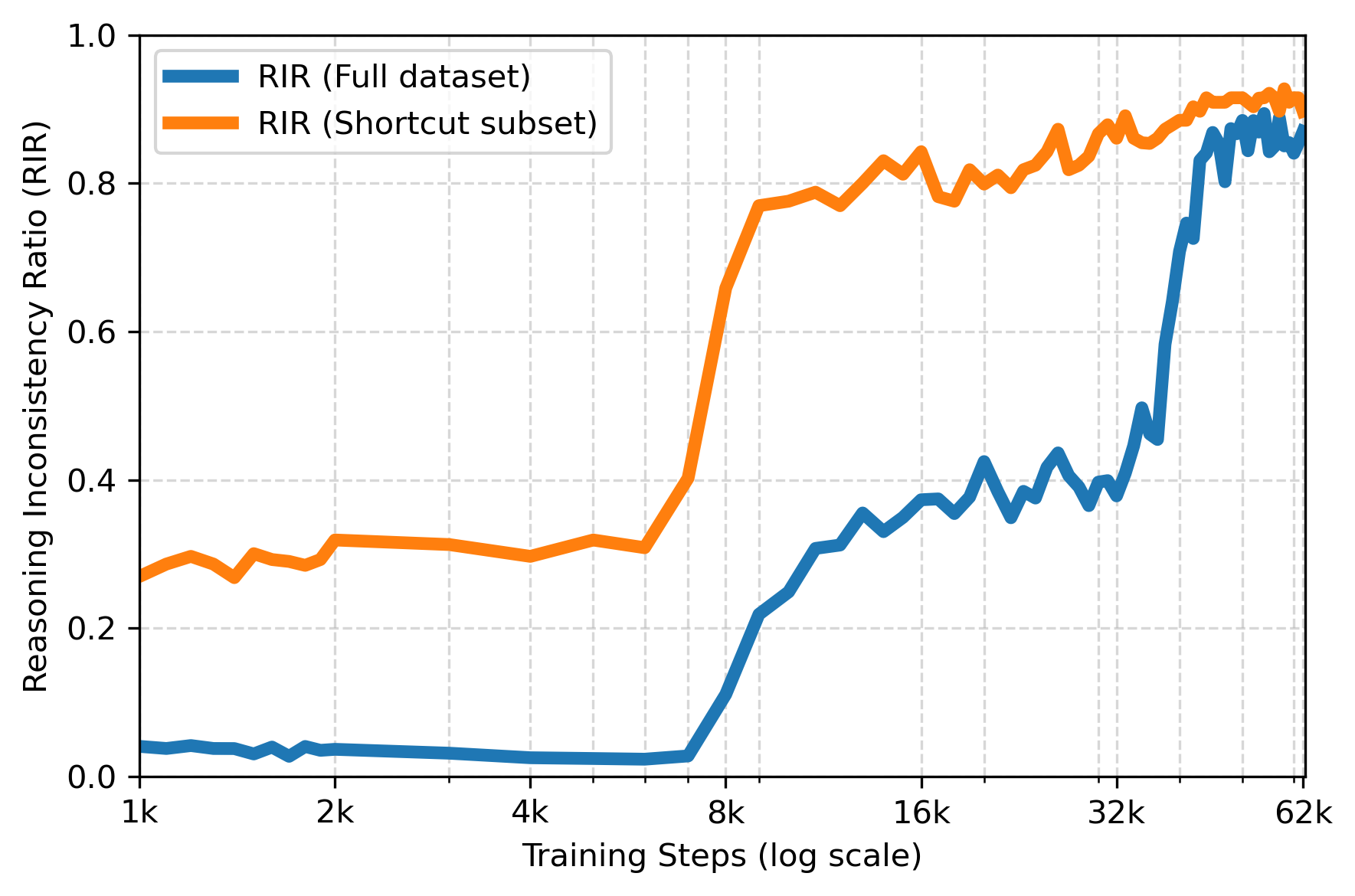}
    \end{subfigure}
    \begin{subfigure}{0.4\textwidth}
        \centering
        \includegraphics[width=\linewidth]{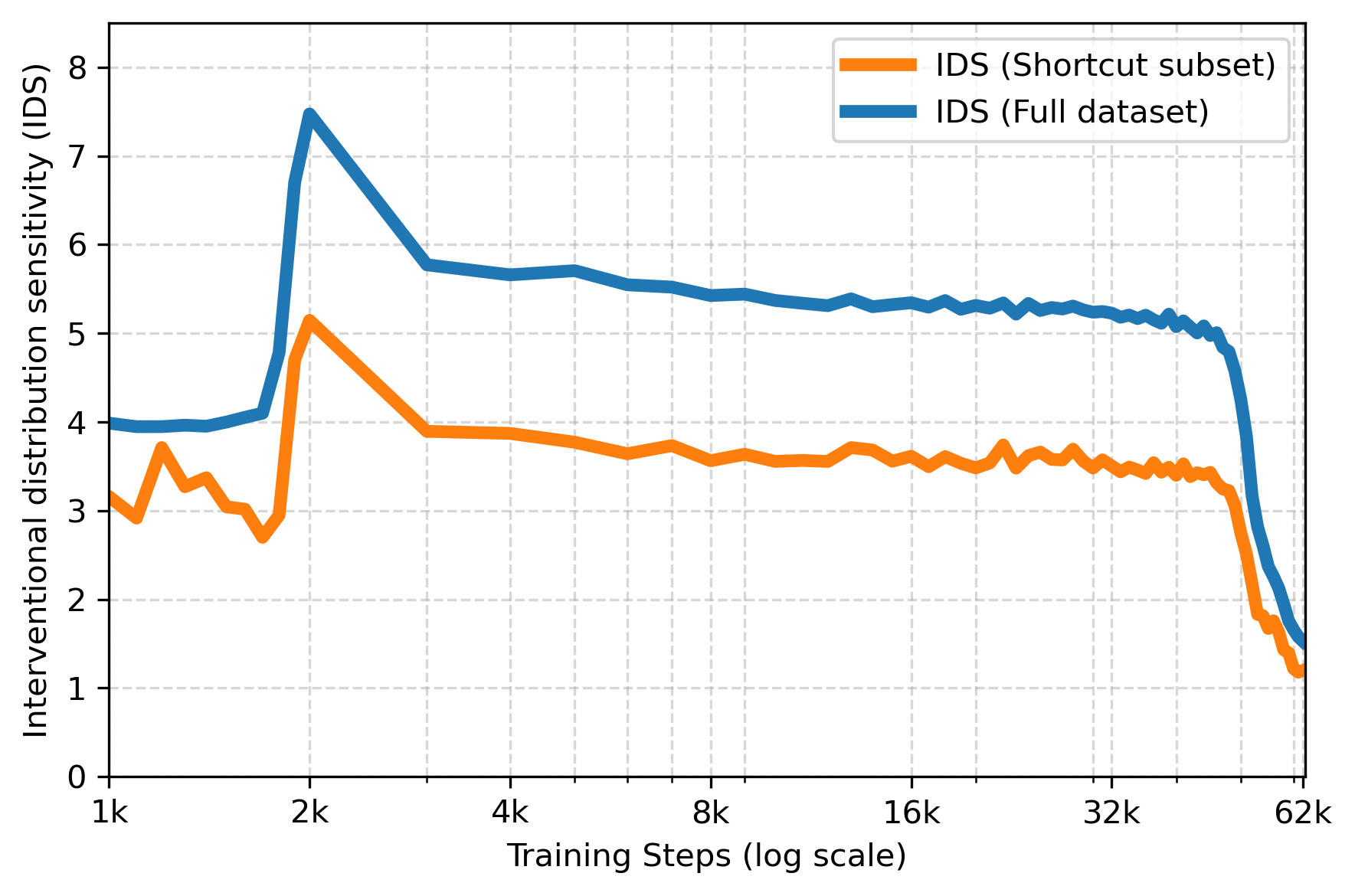}

    \end{subfigure}

    \caption{\textbf{Additional dynamics of unfaithfulness metrics on the shortcut and full datasets.} \textit{\textbf{Left:}}  Throughout training, the $\mathrm{RIR}_1$  measured on the shortcut dataset remains consistently higher than that on the full dataset, indicating that the model is more unfaithful on shortcut data and likely exploits the shortcut. \textit{\textbf{Right:}} Throughout training, the IDS on the shortcut dataset stays consistently lower than on the full dataset, suggesting reduced reliance on $e_2$ and a stronger tendency to adopt skip-step reasoning in the presence of shortcuts.}\label{fig:shortcut2}
\end{figure}

The left panel of Figure~\ref{fig:shortcut2} shows the training-time trajectories of the consistency-based $\mathrm{RIR}_1$ metric on the shortcut dataset and on the full dataset. A larger $\mathrm{RIR}_1$ indicates more pronounced unfaithfulness on the corresponding data. We observe that throughout training, $\mathrm{RIR}_1$ on the shortcut data remains consistently higher than the value computed on the full dataset, consistent with Figure~\ref{fig:shortcut}. Notably, $\mathrm{RIR}_1$ exhibits a step-like pattern over the course of training, whereas $\mathrm{INR}$ increases only once near the end of training. This difference arises because the definition of $\mathrm{RIR}_1$ explicitly accounts for the model's learning accuracy on $e_2$; as the model's accuracy on $e_2$ improves steadily during training, $\mathrm{RIR}_1$ increases in a corresponding stepwise manner.
We plot the training trajectories of the metrics on the shortcut subset and the full dataset under different noise levels, as shown below.

\begin{figure}[H]
    \centering
    \begin{subfigure}{0.31\textwidth}
        \centering
        \includegraphics[width=\linewidth]{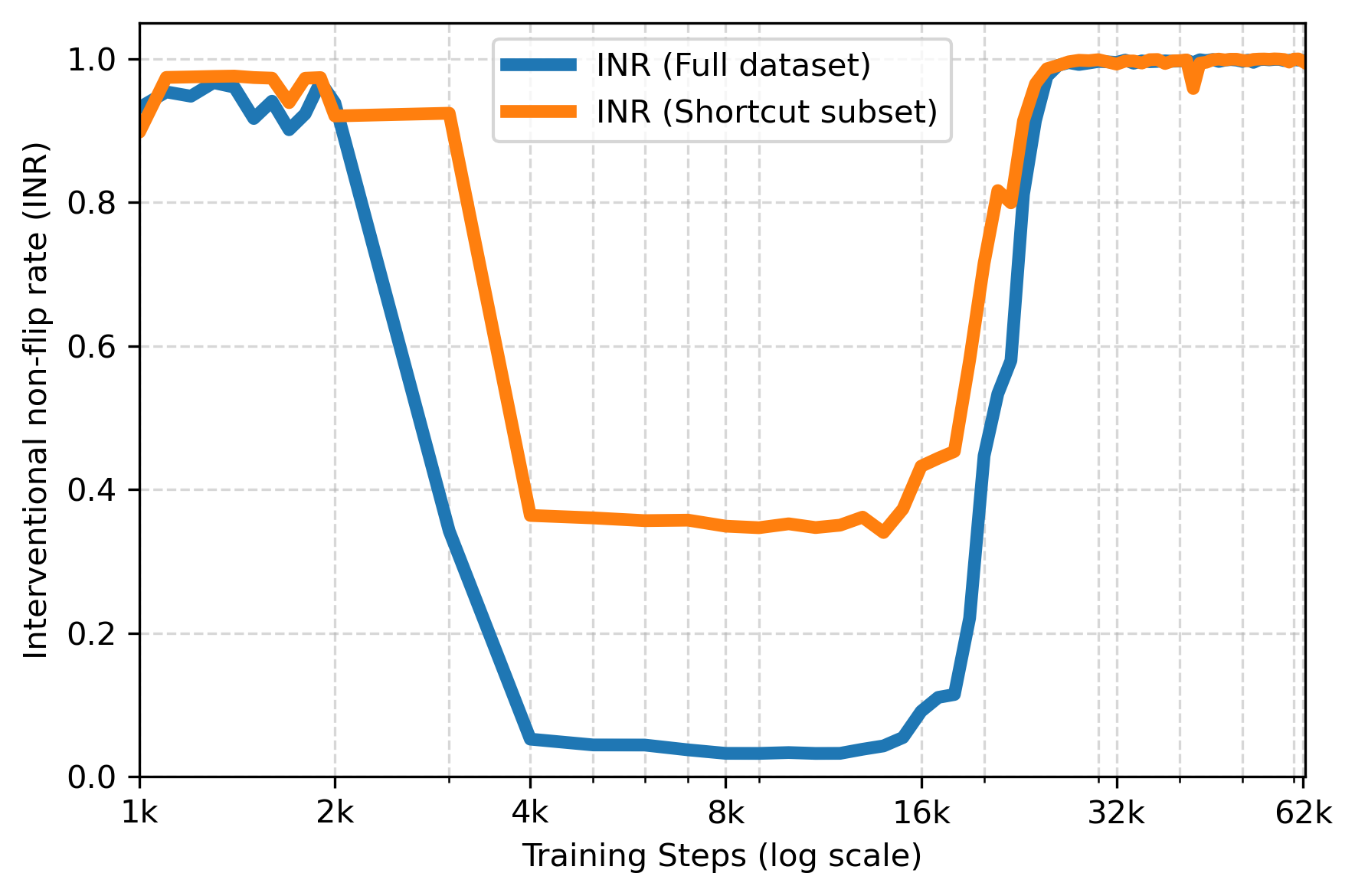}
    \end{subfigure}
    \begin{subfigure}{0.31\textwidth}
        \centering
        \includegraphics[width=\linewidth]{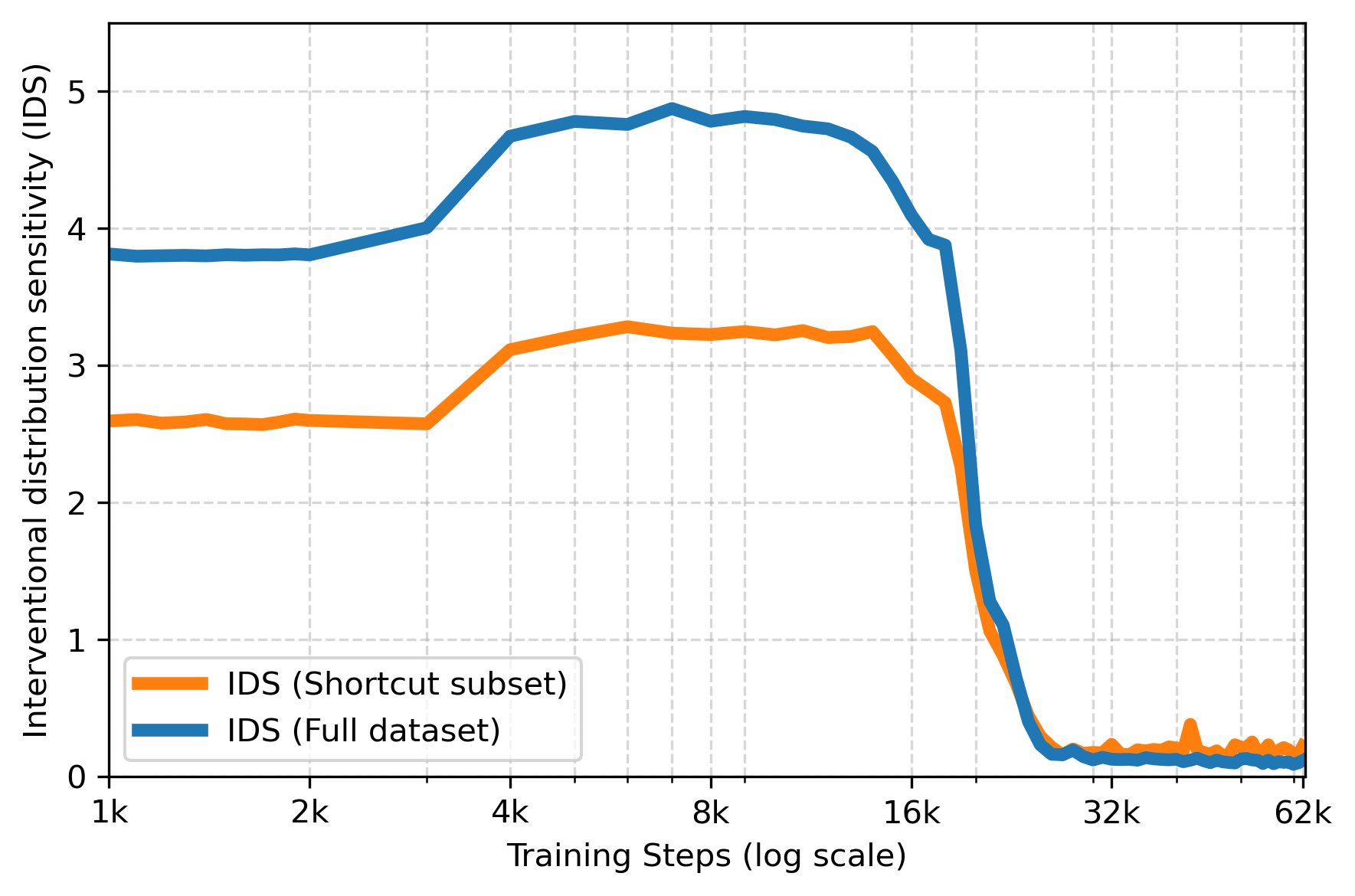}

    \end{subfigure}
    \begin{subfigure}{0.31\textwidth}
        \centering
        \includegraphics[width=\linewidth]{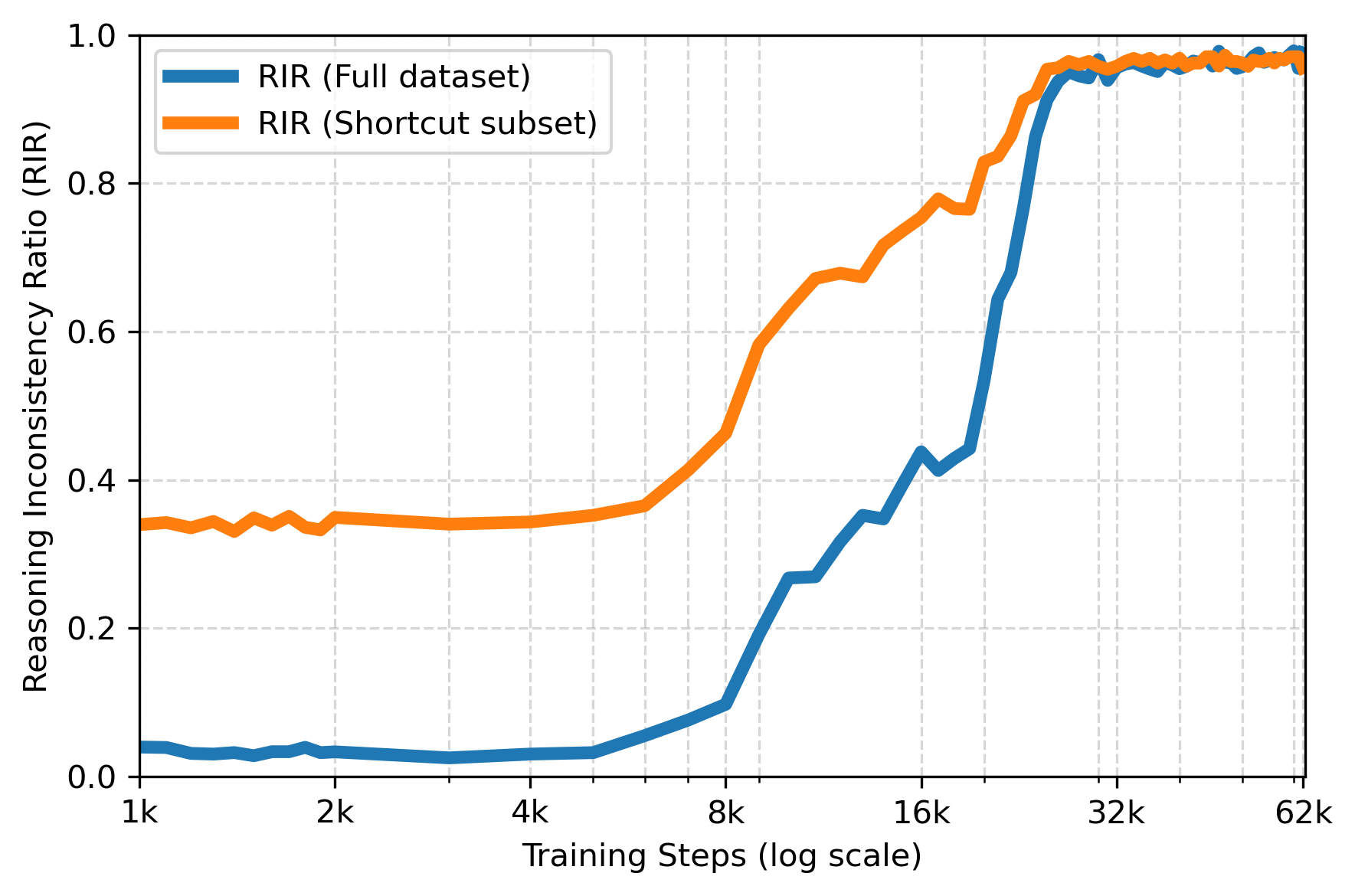}
    
    \end{subfigure}
    \caption{Training dynamics of INR, IDS, and $\mathrm{RIR}_1$ under the setting $N = 94$ and noise configuration $(\varepsilon_1, \varepsilon_2) = (0.01, 0.3)$.}
    \label{fig:mod94_shortcut_0.3}
\end{figure}

\begin{figure}[H]
    \centering
    \begin{subfigure}{0.31\textwidth}
        \centering
        \includegraphics[width=\linewidth]{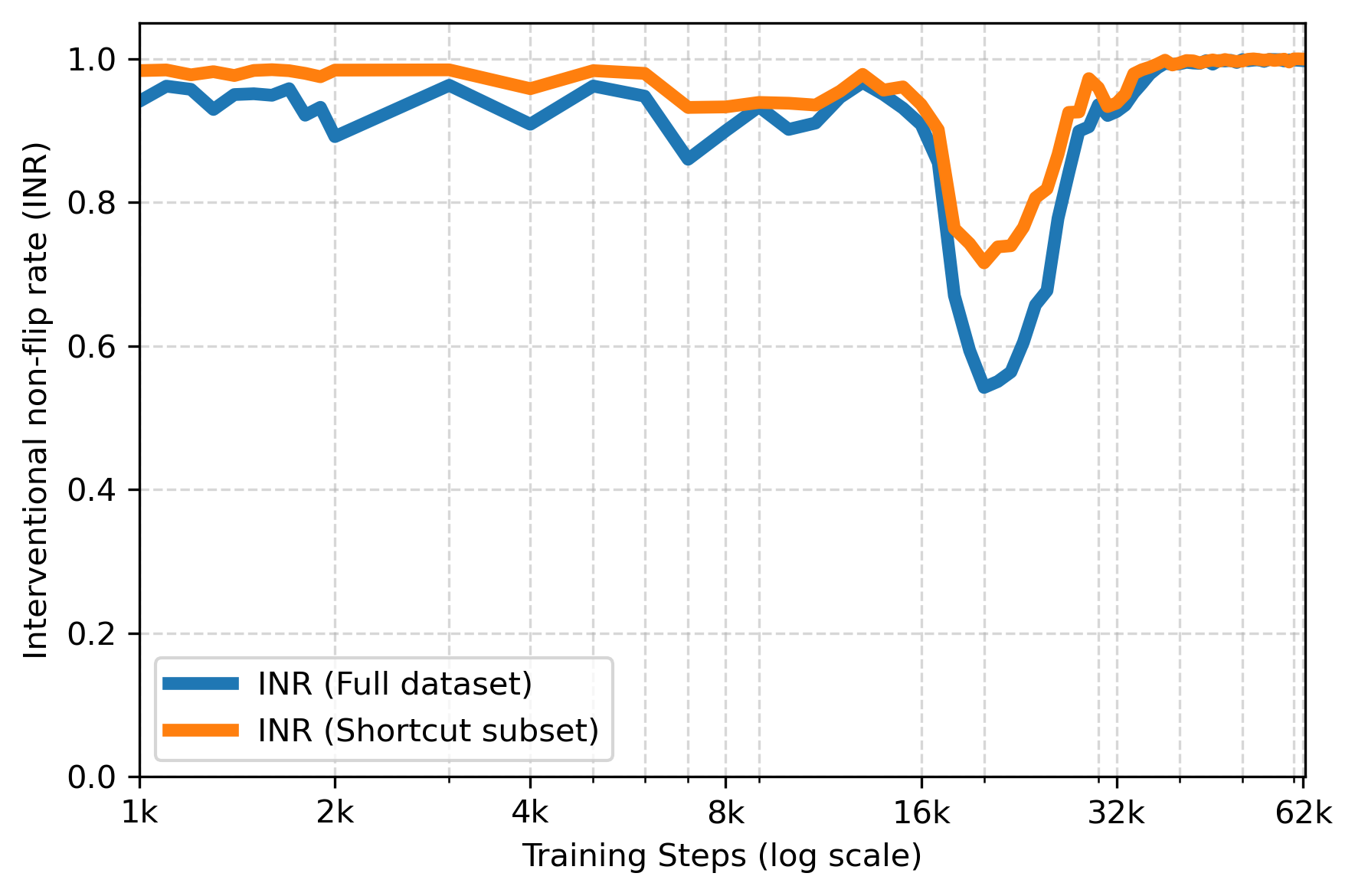}
    \end{subfigure}
    \begin{subfigure}{0.31\textwidth}
        \centering
        \includegraphics[width=\linewidth]{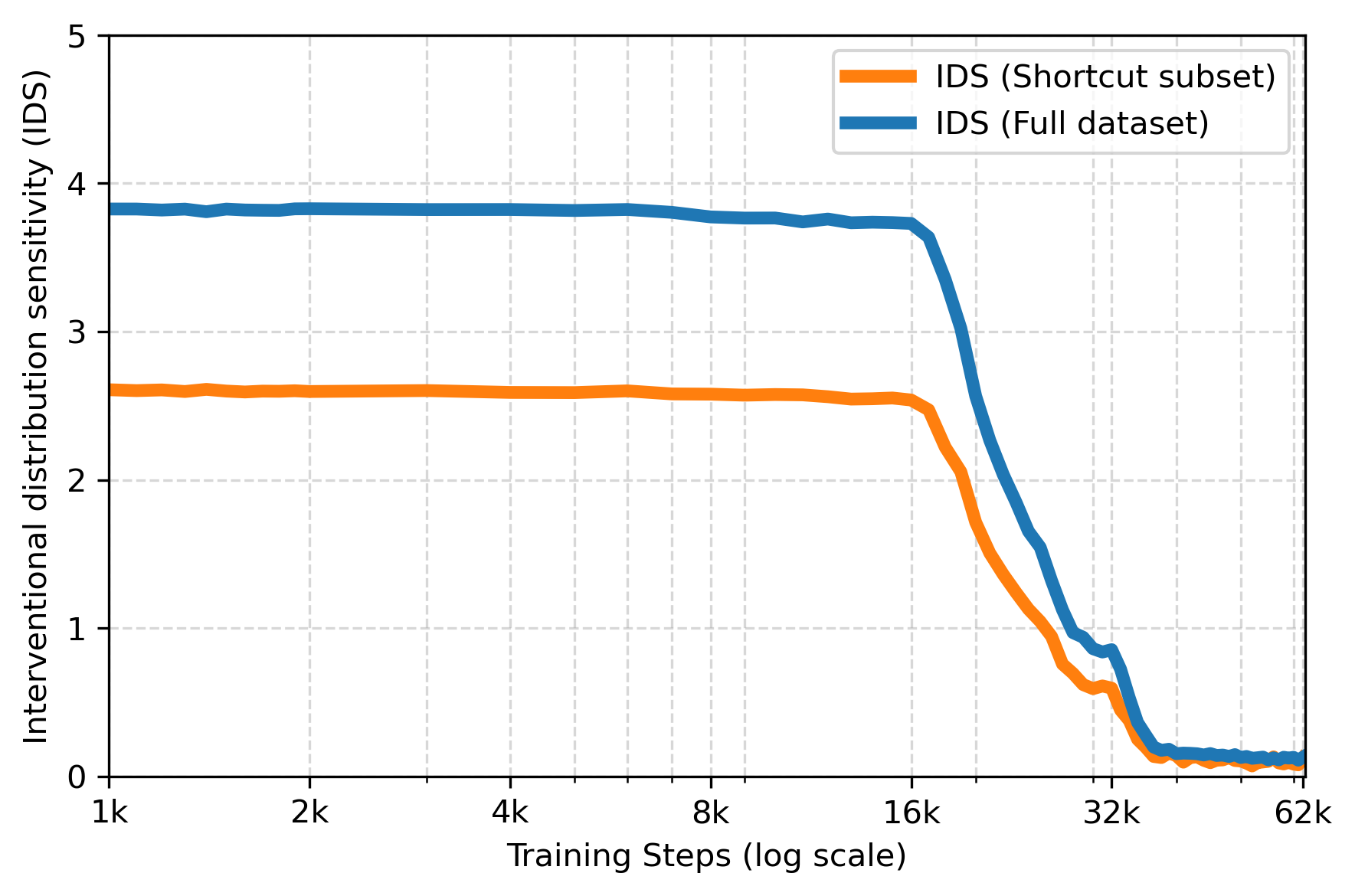}

    \end{subfigure}
    \begin{subfigure}{0.31\textwidth}
        \centering
        \includegraphics[width=\linewidth]{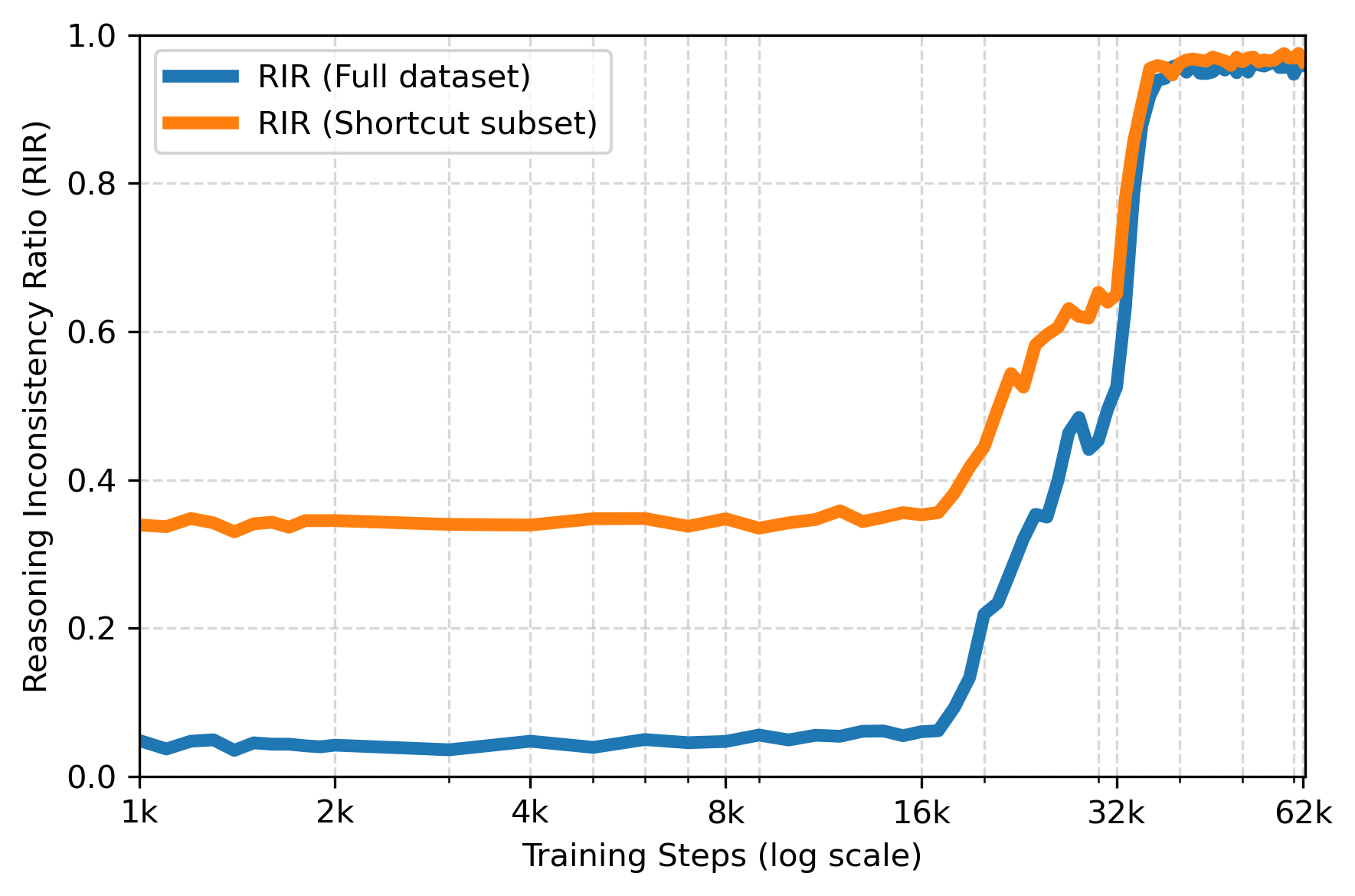}
    
    \end{subfigure}
    \caption{Training dynamics of INR, IDS, and $\mathrm{RIR}_1$ under the setting $N = 94$ and noise configuration $(\varepsilon_1, \varepsilon_2) = (0.01, 0.5)$.}
    \label{fig:mod94_shortcut_0.5}
\end{figure}

\begin{figure}[H]
    \centering
    \begin{subfigure}{0.31\textwidth}
        \centering
        \includegraphics[width=\linewidth]{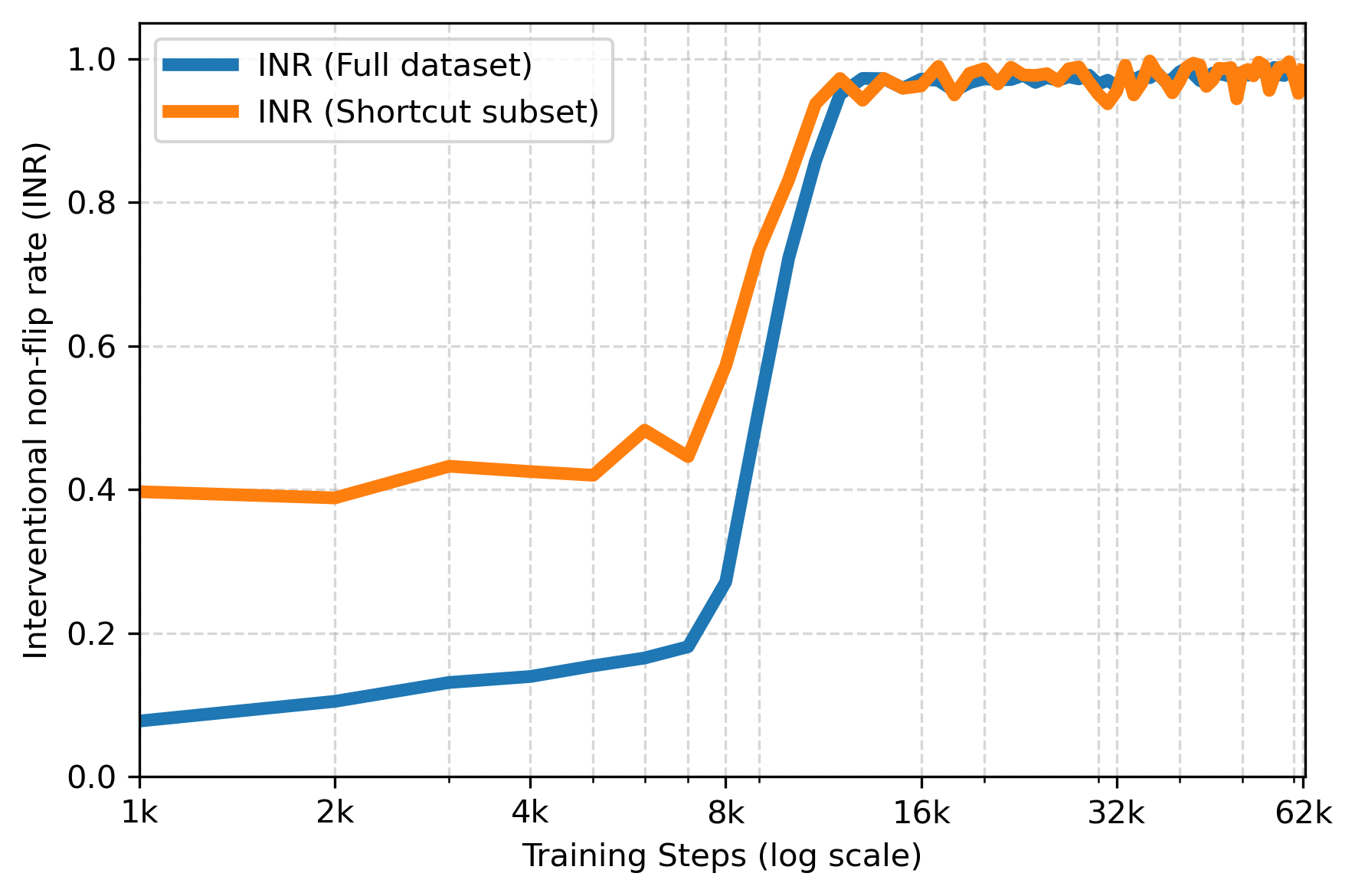}
    \end{subfigure}
    \begin{subfigure}{0.31\textwidth}
        \centering
        \includegraphics[width=\linewidth]{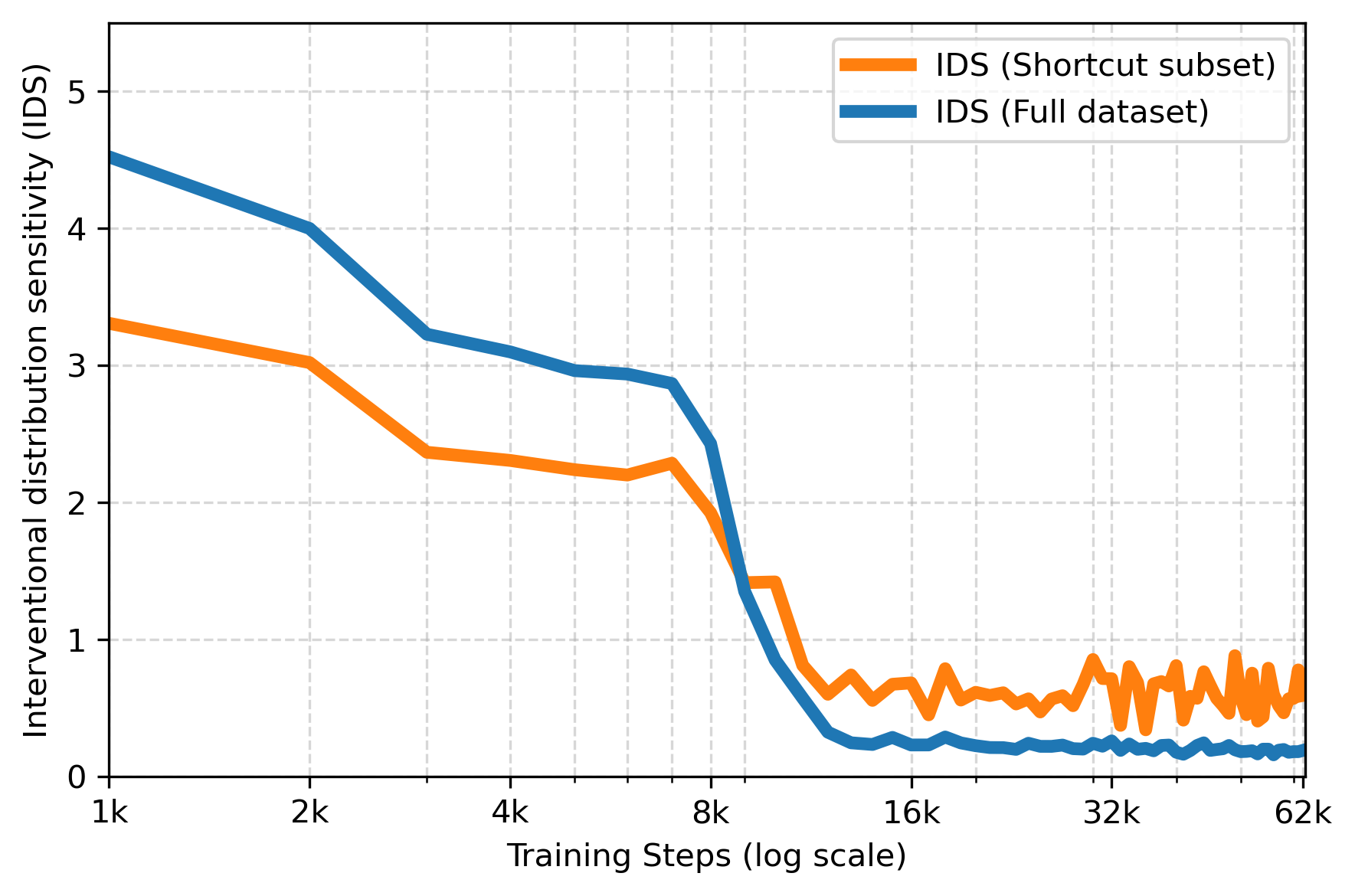}

    \end{subfigure}
    \begin{subfigure}{0.31\textwidth}
        \centering
        \includegraphics[width=\linewidth]{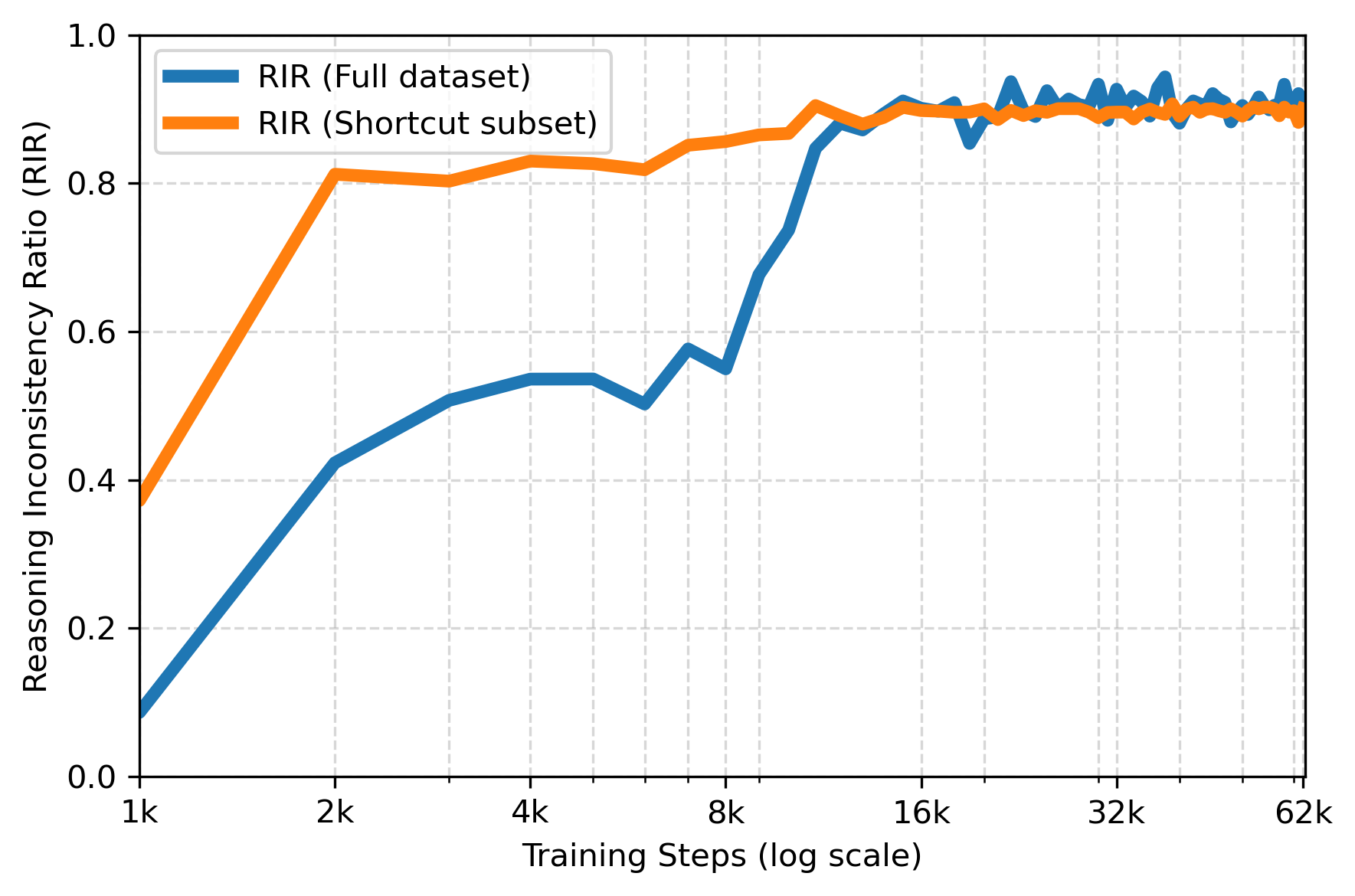}
    
    \end{subfigure}
    \caption{Training dynamics of INR, IDS, and $\mathrm{RIR}_1$ under the setting $N = 38$ and noise configuration $(\varepsilon_1, \varepsilon_2) = (0.01, 0.1)$.}
    \label{fig:mod38_shortcut}
\end{figure}

We repeat the experiments with the modulus reduced from 94 to 38 while fixing the noise levels to $(\varepsilon_1,\varepsilon_2)=(0.01,0.1)$. The corresponding training metrics are shown in Figure \ref{fig:mod38_shortcut}.

\section{Ablations and variants of experiment setups}

We present further results under variants of experimental setups.

\begin{figure}[H]
    \centering
    \begin{tabular}{cc}

        \begin{subfigure}{0.33\textwidth}
            \centering
            \includegraphics[width=\linewidth]{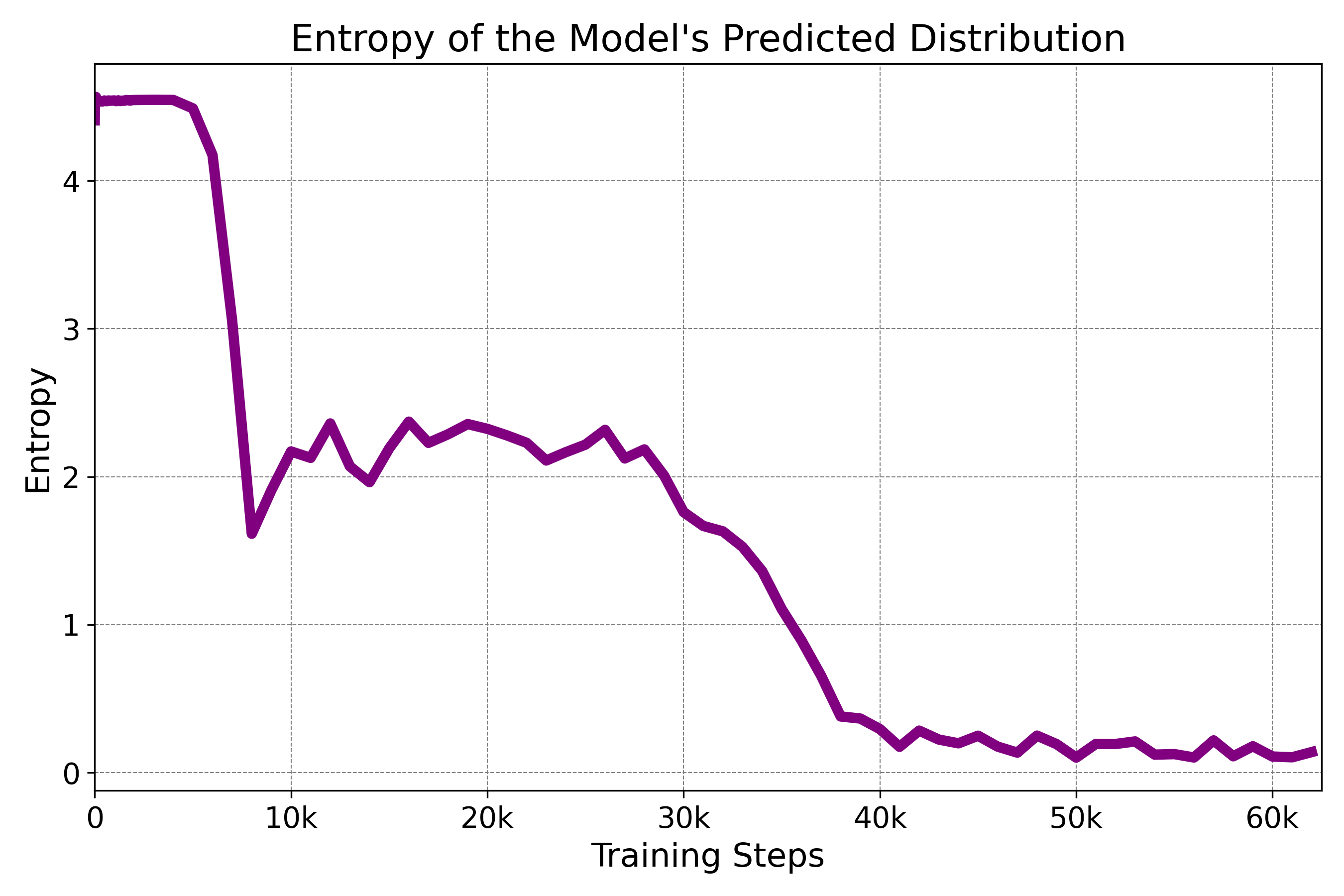}
            \label{fig:sub1}
        \end{subfigure}
        &

        \begin{subfigure}{0.33\textwidth}
            \centering
            \includegraphics[width=\linewidth]{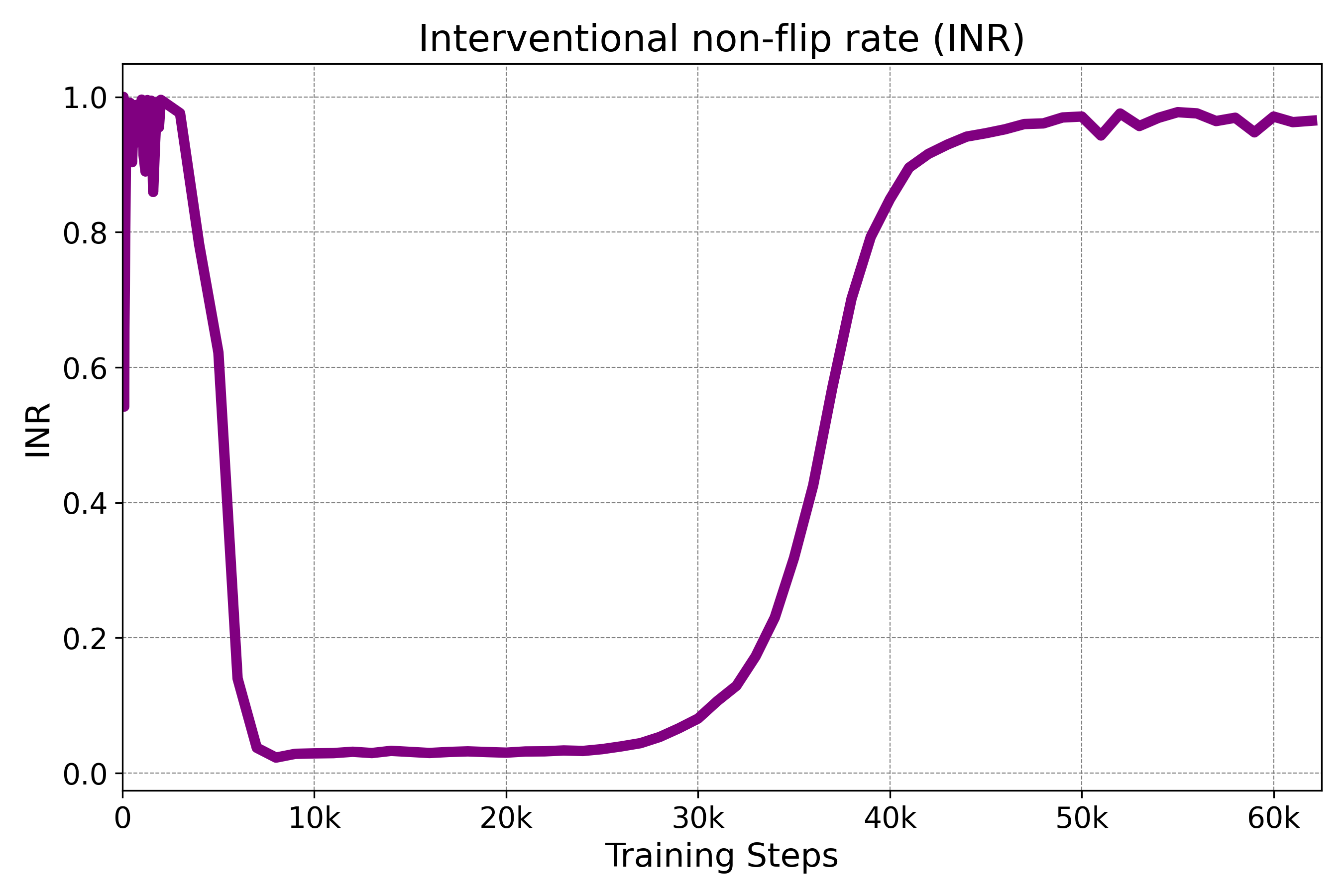}
            \label{fig:sub2}
        \end{subfigure}
        \\
    
        \begin{subfigure}{0.33\textwidth}
            \centering
            \includegraphics[width=\linewidth]{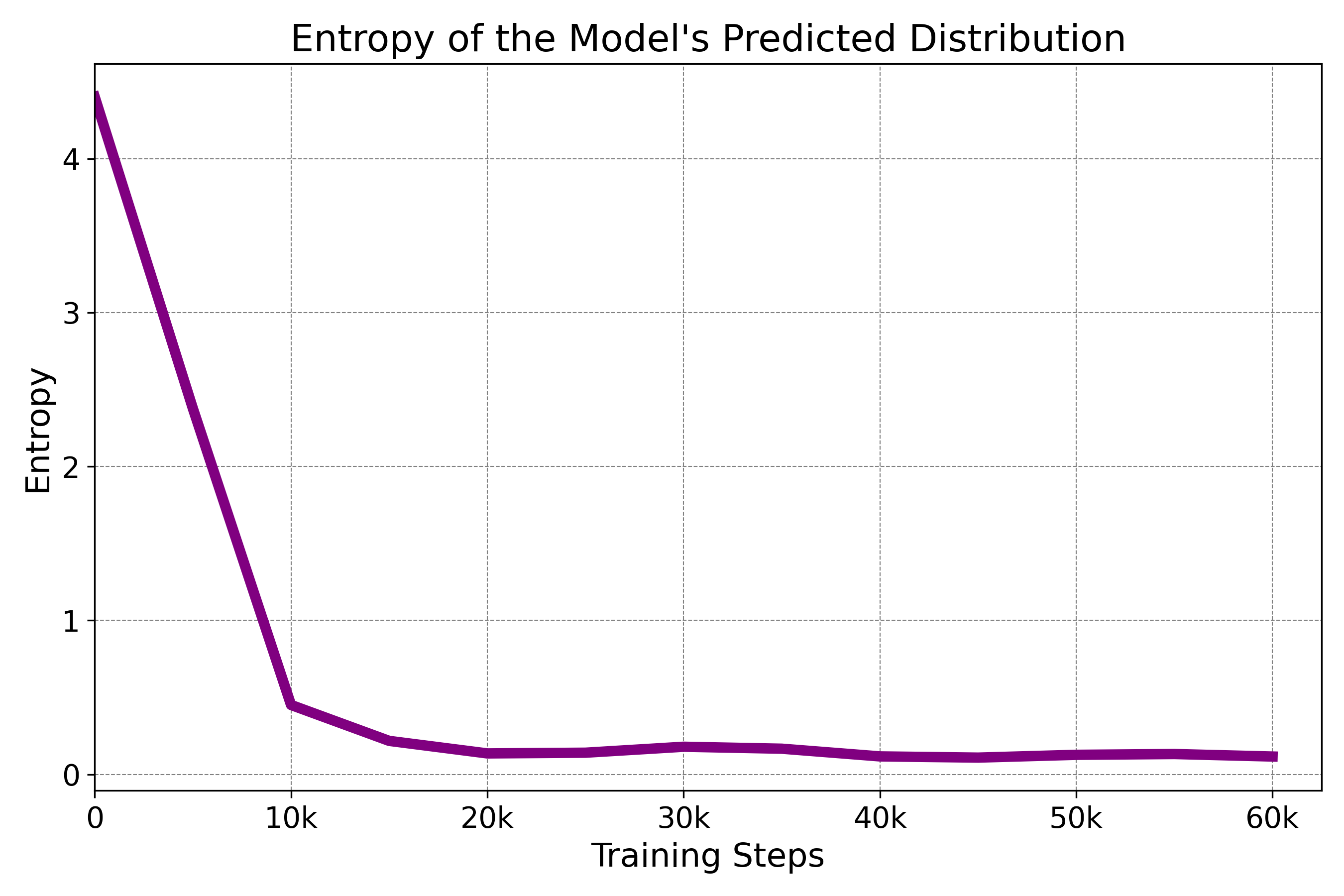}
            \label{fig:sub3}
        \end{subfigure}
        &
       
        \begin{subfigure}{0.33\textwidth}
            \centering
            \includegraphics[width=\linewidth]{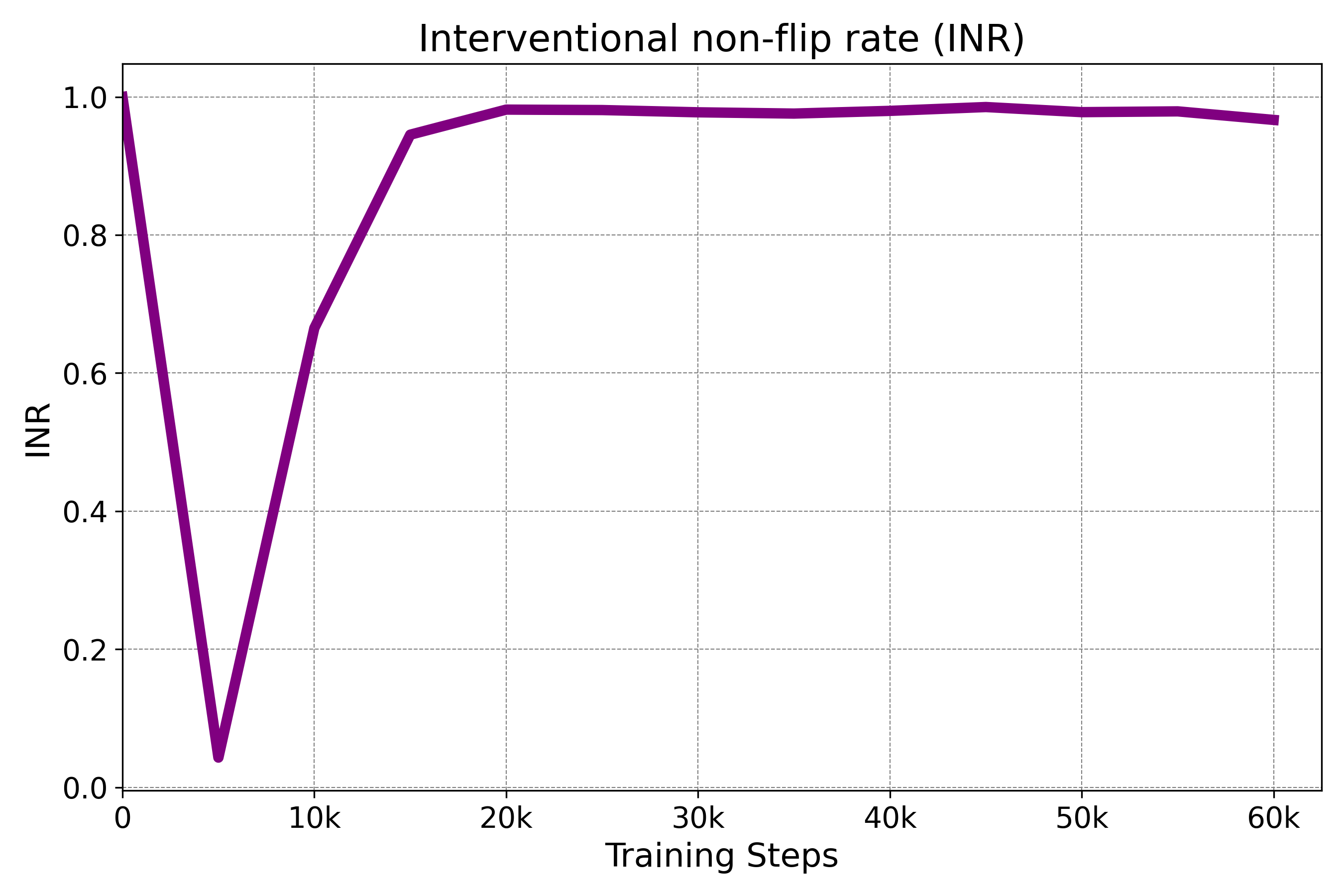}
            \label{fig:sub4}
        \end{subfigure}
    \end{tabular}

    \caption{\textbf{Training dynamics of INR and entropy for 5-layer and 3-layer transformers.} \textbf{The first row:} INR and the entropy of the model’s predictive distribution over the course of training for the 3-layer transformer. \textbf{The second row:} The 5-layer transformer, owing to its larger capacity, rapidly acquires skip-step reasoning and passes through Phases 1 and 2 without a non-monotonic rise in entropy. This is reflected by the sharp transient fluctuations in INR early in training.}
    \label{fig:5layers}
\end{figure}

\paragraph{Larger model size.}We first fix the data setup (recall Eq.~\ref{eq:format}) and increase the model depth from 3 layers to 5 layers while keeping all other hyperparameters unchanged. Figure~\ref{fig:5layers} compares the training dynamics of the 5-layer Transformer at noise level $(\varepsilon_1,\varepsilon_2)=(0.01,0.1)$ with those of the 3-layer model, reporting the evolution of INR and the entropy of the model’s predictive distribution. The 5-layer model also eventually learns the skip-step reasoning strategy; however, its entropy curve becomes essentially monotone and no longer exhibits the non-monotonic ``bump'' associated with the Phase 2 self-verification regime. Consistently, the INR of the 5-layer model drops sharply and then quickly returns to $1$ between $5\text{k}$ and $10\text{k}$ training steps, suggesting that the higher-capacity model spends little time in Phases~1 and~2 and rapidly transitions into the skip-step regime that achieves the optimal solution. In contrast, the 3-layer model lingers substantially longer in the stepwise and self-verification regimes, effectively getting trapped near these suboptimal local minima.

As capacity increases, the transition can become so rapid that intermediate phases are less visible. To better assess whether our qualitative phenomena persist at larger scales---i.e., whether they are \emph{rescalable} rather than artifacts of a particular model or problem size---we further increase the task complexity by enlarging the modulus. Concretely, we increase the dataset modulus to $N=113$ while keeping all other hyperparameters fixed, and train the same 5-layer Transformer under an identical experimental protocol. If the characteristic multi-phase dynamics re-emerge and remain qualitatively consistent in this larger-modulus regime, it provides evidence that our observations are robust to scaling in both model capacity and problem size.

\begin{figure}[H]
    \centering
    \hspace*{-1cm}
\includegraphics[width=0.9\textwidth]{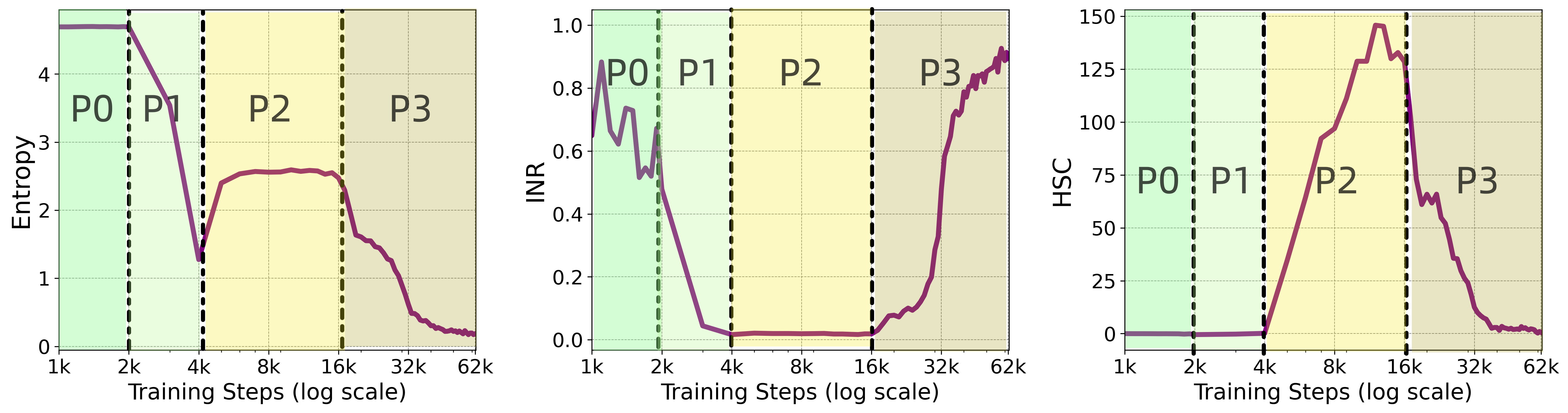}
    \caption{\textbf{Training dynamics of the 5-layer Transformer at noise levels $(\varepsilon_1, \varepsilon_2) = (0.01, 0.1)$ with modulus $N=113$.} \textbf{Left}: In the 5-layer Transformer setting, the entropy of the model’s predictive distribution still clearly delineates the training process into four distinct phases.  \textbf{Middle}: The INR metric computed for the 5-layer model captures the model’s transition across three reasoning modes, from stepwise to mixed and ultimately to skip-step reasoning.
    \textbf{Right}: During training, the 5-layer model also develops internal indicators that support self-verification, as evidenced by the corresponding evolution of the HSC.}
    \label{fig:mod113_metrics}
\end{figure}

Figure~\ref{fig:mod113_metrics} summarizes the training trajectories of a 5-layer Transformer on the $N=113$ setting with noise $(\varepsilon_1,\varepsilon_2)=(0.01,0.1)$. We monitor the entropy of the model’s predictive distribution, an intervention-based unfaithfulness measure, and the HSC metric that quantifies self-verification behavior. Notably, even under this larger-modulus task and with increased depth, the learning process follows essentially the same qualitative pattern as in the 3-layer, $N=97$ experiments. The model transitions from an initial stepwise regime, through an intermediate mixed regime, and ultimately converges to the skip-step strategy. The fact that these regime changes persist under simultaneous scaling of both model capacity and task size supports the view that the phenomena we observe reflect a robust, rescalable training dynamic rather than a peculiarity of a specific configuration.

\begin{figure}[H]
    \centering
    \hspace*{-0.5cm}
\includegraphics[width=0.9\textwidth]{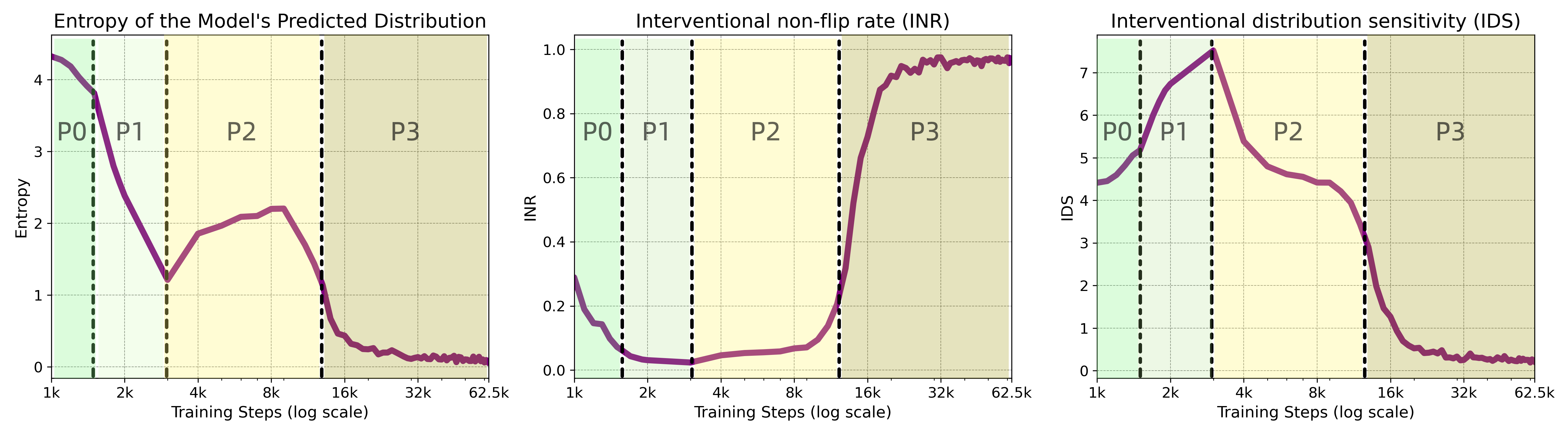}
    \caption{\textbf{Training dynamics at noise levels $(\varepsilon_1, \varepsilon_2) = (0.01, 0.1)$ with modulus $N=83$.} \textbf{Left:}
    The entropy of the model’s output predictions. \textbf{Middle:} During  training steps,  INR first drops to zero, indicating that predictions of \(e_3\) rely entirely on \(e_2\), and later rises toward one, showing that \(e_3\) is eventually predicted solely from \(e_1\). \textbf{Right:} IDS first increases and then decreases, mirroring the change in the model’s reliance on $e_2$. }
    \label{fig:entropy_mod83}
\end{figure}

\begin{figure}[H]
    \centering
    \hspace*{-1cm}
\includegraphics[width=0.9\textwidth]{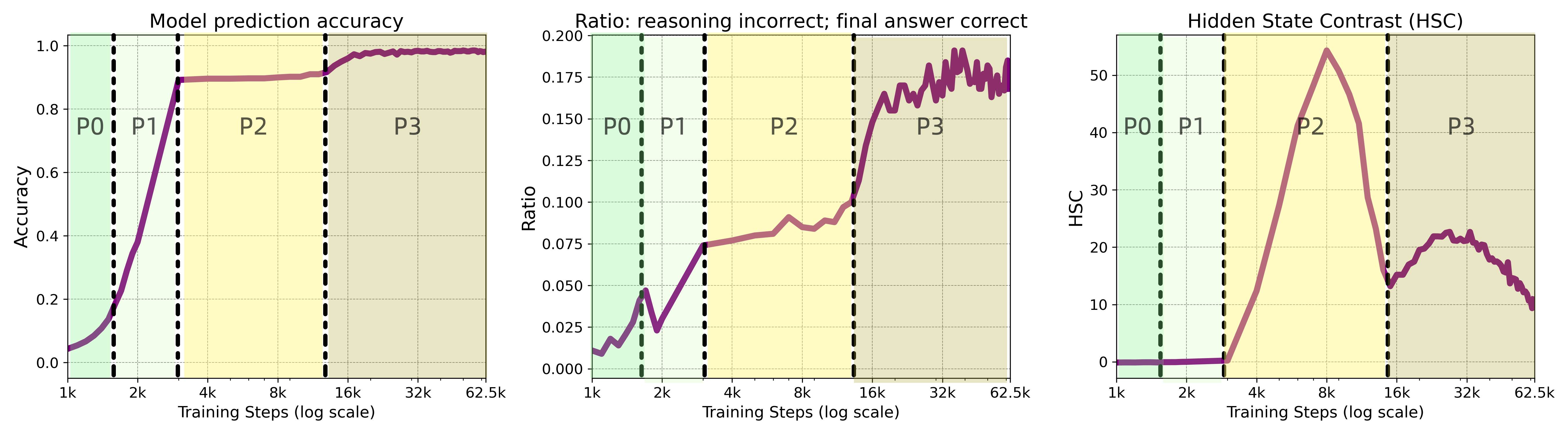}
    \caption{\textbf{Training dynamics at noise levels $(\varepsilon_1, \varepsilon_2) = (0.01, 0.1)$ with modulus $N=83$.} \textbf{Left}: The model’s prediction accuracy reaches 90\% in Phase 1, remains essentially unchanged throughout Phase 2, and then increases to around 99\% in Phase 3, consistent with the evolution of the model’s internal reasoning mechanism. \textbf{Middle}: Unfaithfulness ratio: $e'_2 \neq e_2,\; e'_3 = e_3$. (Reasoning incorrect; final answer correct.) \textbf{Right}: Hidden state contrast (HSC) increases rapidly during Phase 2, but then drops sharply during the transition from Phase 2 to Phase 3, suggesting that a self-verification mechanism is forming within the model.}
    \label{fig:accuracy_mod83}
\end{figure}

\paragraph{Different modulus.} We changed $N$ from $97$ to $83$.  Figure~\ref{fig:entropy_mod83} and \ref{fig:accuracy_mod83} show that for the modulus-83 setting, the training dynamics of the consistency-based and intervention-based unfaithfulness measures, as well as the model’s output entropy, prediction accuracy, and hidden state contrast (HSC). These metrics exhibit trends that closely mirror those observed under modulus 97, and the training process again shows clear phase transitions, with the model’s behavior in each phase remaining effectively the same as in the modulus-97 case.

\hspace{-10cm}

\begin{figure}[H]
    \centering
    \hspace*{-0.5cm}
\includegraphics[width=0.8\textwidth]{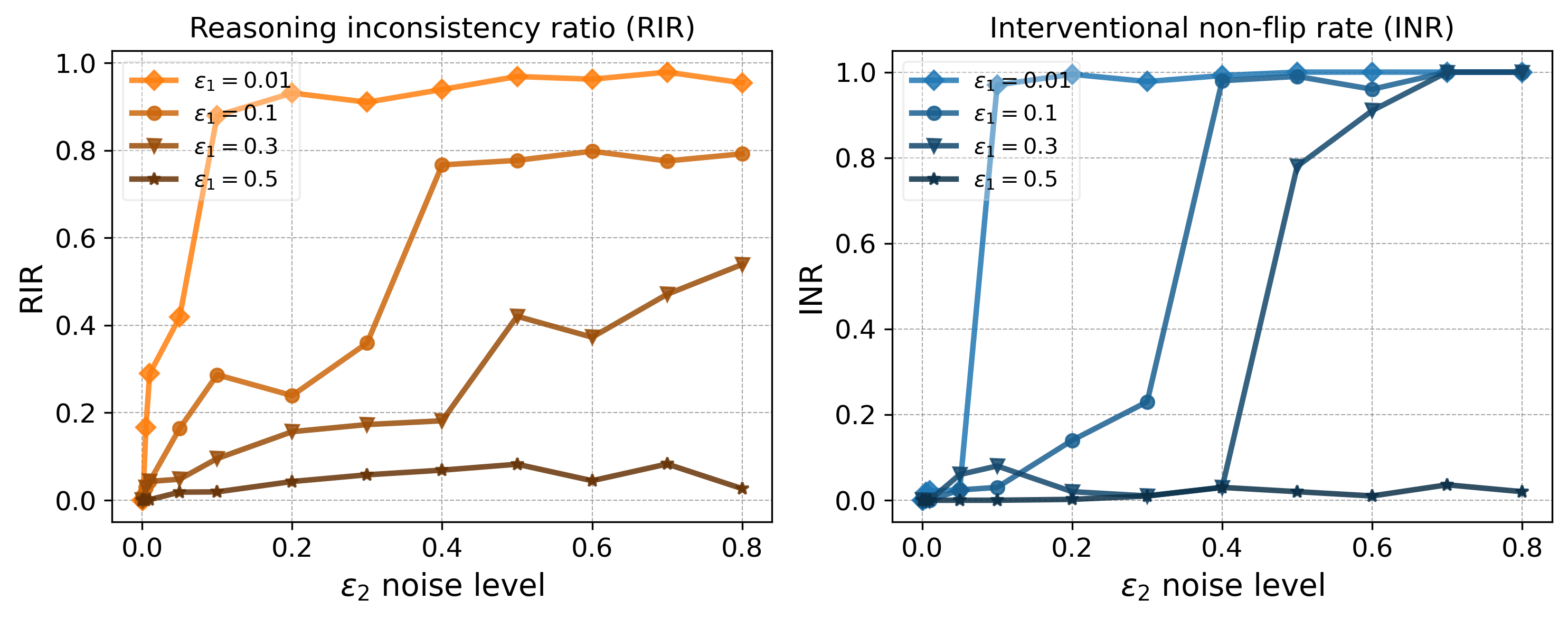}
    \caption{\textbf{Transformer models trained on modulus $N=83$ datasets under varying noise levels.} For modulus 83, we evaluate CoT faithfulness under two definitions by training transformers on datasets with varying noise levels ($\varepsilon_1$: prompt noise, $\varepsilon_2$: reasoning noise). In the low-noise regime both unfaithfulness measures remain small, but beyond a critical $\varepsilon_2$ they increase sharply—{\color{orange}orange} curves indicate inconsistent reasoning chains, while {\color{blue}blue} curves show that interventions on the reasoning steps hardly affect the final answers.
    }
    \label{fig:RIR_INR_mod83}
\end{figure}

Figure~\ref{fig:RIR_INR_mod83} shows the trends of the model’s unfaithfulness metrics under four different noise levels: $\varepsilon_1=(0.01,0.1,0.3,0.5)$. As in the modulus-97 setting, for any fixed $\varepsilon_1$ the model retains a certain degree of stepwise reasoning below a critical threshold $\tau_{c}(\varepsilon_1)$; once $\varepsilon_2$ exceeds this threshold, the unfaithfulness measures increase sharply.

\paragraph{Different data format.} We consider an alternative training-example format in which parentheses are added to the prompt expression, while keeping the modulus consistent with the main text (i.e., $N=97$). The modified data format is as follows:
\begin{equation*}
    \underbrace{(a \times b) - c}_{E_1 :\text{prompt}} \ \to \ \underbrace{d - c}_{E_2: \text{reasoning}} \ \to \ \underbrace{e}_{E_3: \text{solution}}.
\end{equation*}

\begin{figure}[H]
    \centering
    \hspace*{-0.5cm}
\includegraphics[width=0.9\textwidth]{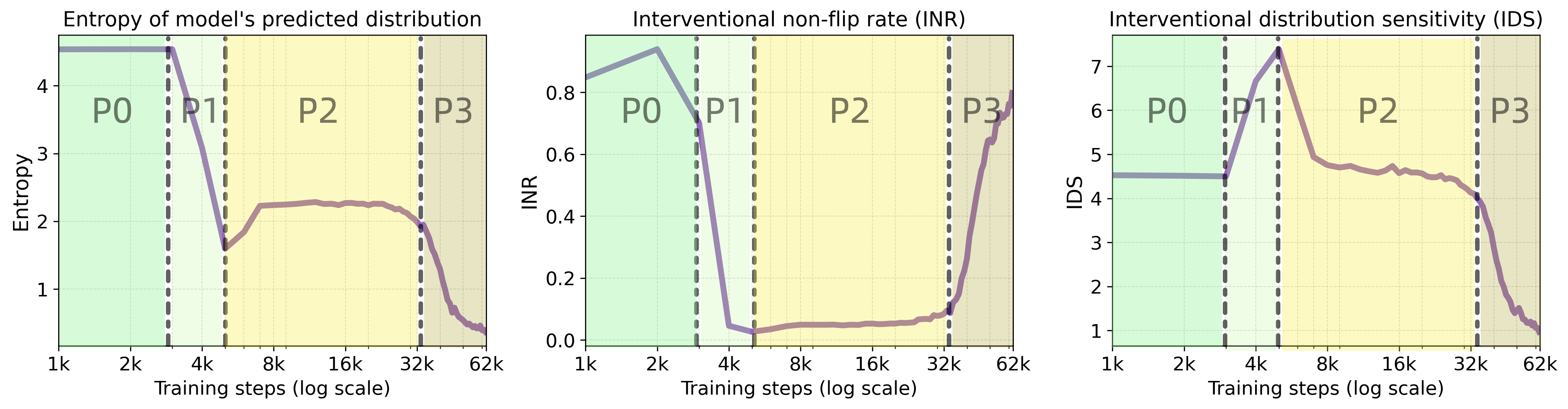}
    \caption{\textbf{Training dynamics on the parenthesized data variant.} \textbf{Left:}
    The entropy of the model’s output predictions. \textbf{Middle:} During  training steps,  INR first drops to zero, indicating that predictions of \(e_3\) rely entirely on \(e_2\), and later rises toward one, showing that \(e_3\) is eventually predicted solely from \(e_1\). \textbf{Right:} IDS first increases and then decreases, mirroring the change in the model’s reliance on $e_2$. }
    \label{fig:entropy_different_format}
\end{figure}

Figure~\ref{fig:entropy_different_format} shows the training dynamics on the parenthesized data variant. Overall, the model exhibits nearly identical behavior and phase transitions to those observed in the non-parenthesized setting. This is expected because parentheses in our construction merely standardize the surface form of the expressions and do not alter the underlying computation. Therefore, it is reasonable to focus on the simplest format—i.e., the version without parentheses (see Eq.~\ref{eq:format})—for training and analysis.
% ==== end inlined latex_files/append_additional.tex ====
\end{document}